\documentclass{article} %
\usepackage{iclr2027_arxiv,times}

\usepackage{amsmath,amsfonts,bm}

\def\eqref#1{equation~\ref{#1}}

\def\1{\bm{1}}

\DeclareMathAlphabet{\mathsfit}{\encodingdefault}{\sfdefault}{m}{sl}
\SetMathAlphabet{\mathsfit}{bold}{\encodingdefault}{\sfdefault}{bx}{n}

\usepackage{amsfonts}
\usepackage{fancyhdr}
\usepackage[hidelinks]{hyperref}
\usepackage[a4paper, top=2.5cm, bottom=2.5cm, left=2.2cm, right=2.2cm]%
{geometry}
\usepackage{times}
\usepackage{amsmath}
\usepackage{amsthm}
\usepackage{changepage}
\usepackage{amssymb}
\usepackage{graphicx}%
\usepackage{hyperref}
\usepackage{booktabs}
\usepackage{multirow}
\usepackage{url}
\usepackage{xcolor}
\usepackage{subcaption}

\title{Data Scarcity and Model Sparsity: \newline Mixtures-of-Experts Overfit More to Repeated Data}
\headertitlestring{Data Scarcity and Model Sparsity}

\usepackage{authblk}

\author[1]{Atindra Jha\thanks{equal contribution; correspondence to \texttt{atindra@cs.stanford.edu}, \texttt{margsli@cs.washington.edu}} }
\author[2]{Margaret Li$^{*}$}
\author[1]{Jure Leskovec}
\author[1]{Percy Liang}
\author[2]{Luke Zettlemoyer}
\affil[1]{Stanford University}
\affil[2]{Paul G. Allen School of Computer Science, University of Washington}

\arxivcopy %
\begin{document}

\maketitle

\newcommand{\pl}[1]{\textcolor{blue}{[PL: #1]}}
\begin{abstract}
As the supply of human-written text is 
exhausted, it has become standard practice to repeat language model training data. Prior work has studied data repetition for \emph{densely activated} Transformers, but the effects of data repetition remains largely unexplored for recently dominant sparse architectures such as Mixture-of-Experts (MoE), despite their increased compute efficiency.
We vary data repetition rates across single- and multi-domain data mixes, and across MoE settings, including expert count and granularity. We consistently find, for models ranging from 80M to 1B active (8.5B total) parameters, that MoEs degrade more rapidly under data repetition. This effect increases with sparsity, dictated by \emph{total} rather than active parameters. While 80M dense models can repeat data over $8\times$ with minimal degradation, 
MoEs (active-parameter-matched) instead begin to suffer at 4 repetitions, and deteriorate rapidly, ceding their performance benefits in all-unique data settings to dramatically underperform dense models after 32 repetitions.
We experiment with existing regularization methods as a potential remedy. We find that some methods, such as dropout, can mitigate overfitting. In particular, with strong masking-based regularization,
MoEs are able to outperform dense models even when data is repeated more than 64 times. However, no method fully matches the performance of all-unique training data with matched total data and compute. 
Finally, we analyze internal mechanisms correlated with MoE overfitting in high repetition regimes, 
and find that MoE routing universally stabilizes early in training, and that expert specialization correlates with overfitting to repeated data.
In sum, our work addresses the adverse interactions between sparsity and data repetition: we present evidence for the core mechanisms of overfitting and its potential remediation, and suggest promising avenues for future methods to reduce over-specialization in model parameters by disrupting memorization patterns.

\end{abstract}

\providecommand{\figph}[3]{%
  \begin{figure}[t]
  \centering
  \fbox{\parbox{0.92\linewidth}{%
    \vspace{0.7em}%
    \centering\textbf{[FIGURE PLACEHOLDER]}\\[0.5em]%
    \raggedright\small #3%
    \vspace{0.7em}}}
  \caption{#2}
  \label{#1}
  \end{figure}}

\section{Introduction}
\label{sec:intro}

As language model training begins to exhaust even massive-scale web crawls, 
the amount of unique training data has become a constraining factor for language model training, in addition to compute. 
It is now common practice to repeat some or all training data, despite the known tendency of language models to overfit to data under high repetition rates \citep{muennighoff2023scaling}. Simultaneously, the compute costs of large-scale LLM training have driven the adoption of relatively compute-efficient Mixture-of-Experts models \citep{shazeer2017outrageously,switch,olmoe}. These models achieve compute efficiency via \emph{sparsity}, the ratio of \emph{total} to \emph{active} parameters. 
When training only with unique data, increased sparsity is known to consistently improve training efficiency.
However, the interaction between sparsity and data repetition remains relatively unexplored.

Sparsity decouples \emph{total} parameters from \emph{active} parameters, and data repetition decouples \emph{total} data tokens from \emph{unique} data tokens. 
Thus, both sparsity and data repetition introduce axes of variation 
to modern scaling laws, which prescribe simple token-to-parameter ratios without disambiguating between \emph{unique} or \emph{total} training tokens over \emph{active} or \emph{total} parameters.
MoE performance also depends on architectural details such as expert size and count. Further, not all data tokens are equal, with significant research devoted to filtering, deduplication, and data mix domain makeup \citep{li2024datacomplm}. Each axis of variation is entangled with all others, but prior work primarily investigates these axes in isolation: the data-constrained scaling laws of \citet{muennighoff2023scaling} are fit to dense models on a single corpus (C4), and \citet{xue2023repeat} consider a single MoE configuration to posit that multi-epoch degradation increases with total parameters.

To investigate these interactions more fully, 
we conduct an in-depth grid sweep over axes of variation. Across three active-parameter scales (80M, 200M, 1B), we compute-match by training on a fixed \emph{total} data budget 
for each active-parameter scale. We vary the \emph{unique} tokens (equivalently, the \emph{repetition rate $R$}), comparing dense transformers to MoE architectures with varied expert count and granularity. We consider a variety of domains (web crawl, code, scientific, and encyclopedic text), both as single-domain corpora and as components of data mixes with different per-domain repetition rates. Our results indicate that MoEs overfit more to repeated data in comparison to dense models; 80M MoEs clearly degrade at $4\times$ data repetition (compared to $8\times$ for dense models), and cede their performance advantage at $32\times$ data repetition. Further, overfitting becomes more severe with higher MoE sparsity, and its patterns depend on \emph{total} rather than active parameters. Surprisingly, this phenomenon follows similar patterns across data-token-per-parameter ratios and across our varied data domains, only slightly decreased by quality filtering. However, when repeated data is mixed into a non-repeated dataset, the non-repeated dataset 
may have a regularizing effect. 
We also find that some existing regularization methods (dropout, output masking) reduce the impact of high data repetition rates, especially in MoEs. At sufficiently high dropout probability, MoEs can outperform dense models even at $64\times$ data repetition. 
Finally, we perform mechanistic analyses and show that MoE routers stabilize their decisions early in training, which suggests that expert parameters update on a small and stationary subset of the total tokens, and we measure the resulting impact of data repetition on expert specialization.

Overall, our contributions are as follows: 
\begin{itemize}
\item We show that the benefit of sparsity is conditional on the \emph{unique} data budget: 
Across data domains and mixes, MoEs outperform dense models on unique data, but underperform at high data repetition rates. 
Data quality filtering has only minor impact on repetition effects: 
models overfit slightly more to unfiltered data (\S\ref{sec:expts_olmo_mix}-\ref{sec:expts_filtering}).
\item We study domain-specific repetition rates in data mixes. Performance degradation is confined to repeated domains, 
and unique tokens from another domain may mitigate overfitting of repeated domains (\S\ref{sec:expts_mixes}). 

\item We apply regularization techniques and demonstrate that 
dropout and output masking can mitigate overfitting from data repetition (\S\ref{sec:reg}).

\item We analyse MoE expert activations and outputs. Our evidence shows 
that MoE router decisions stabilize early, which exacerbates overfitting as expert parameters over-specialize, updating on a small and near-stationary subset of tokens (\S\ref{sec:analysis}).
\end{itemize}

\begin{figure*}[!ht]
    \centering
    \begin{subfigure}[t]{\textwidth}
    \includegraphics[width=\linewidth]{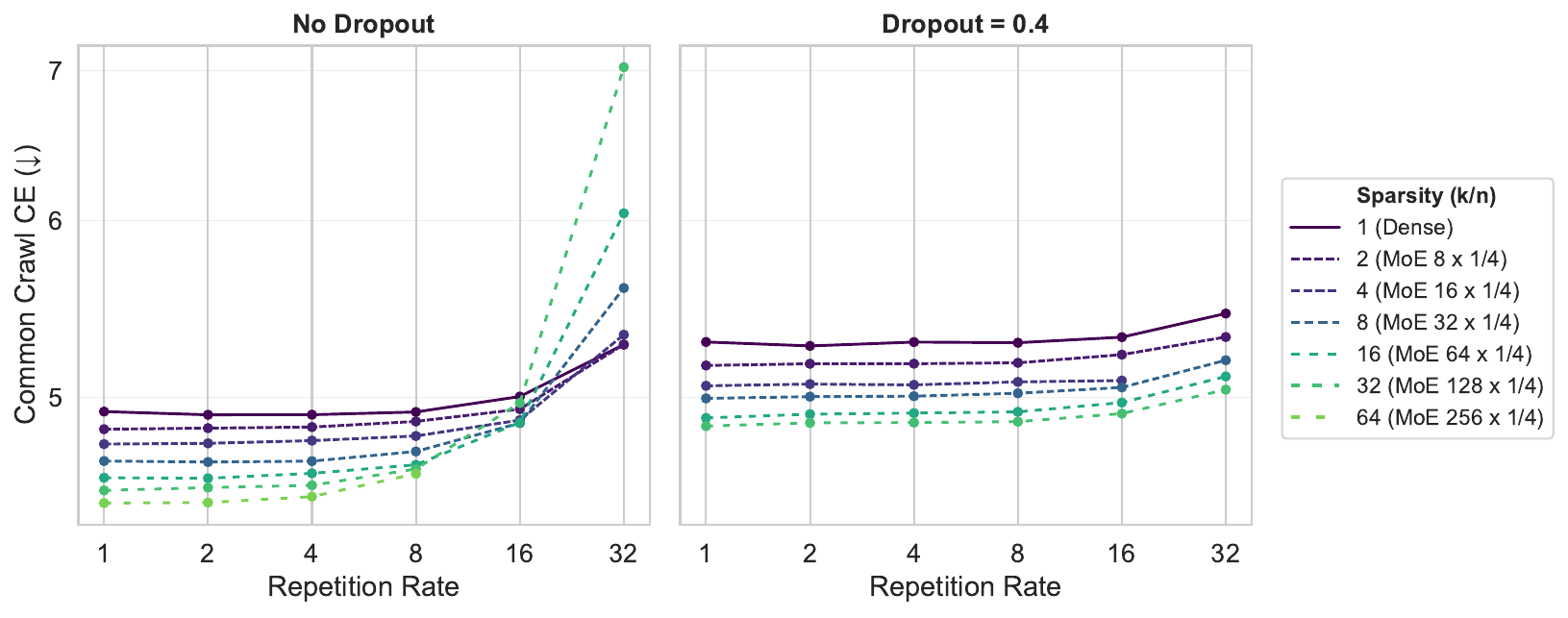}
    \end{subfigure}
    \caption{\textbf{Sparser MoEs degrade more with data repetition (left), but regularization may ameliorate the effect (right).}
    Matching compute via total data and active parameters, we find that higher repetition of fewer unique tokens results in performance degradation. This phenomenon increases in severity with MoE sparsity, and thus depends on \emph{total} rather than active parameters (\S\ref{sec:expts_olmo_mix}). Patterns are consistent across data mixes, total tokens, and other configuration details (\S\ref{sec:expts_olmo_mix}-\ref{sec:expts_mixes}). However, sufficiently strong regularization may substantially decrease overfitting (\S\ref{sec:reg}).
    }
    \label{fig:fig1}
\end{figure*}

\section{Background}
\label{sec:background}
\subsection{Mixture of Experts Language Models}
\label{sec:bg_moe}

In Mixture-of-Experts Transformer LMs, the Feed-Forward (FFN) of each layer is replaced by 
$n$ parallel FFN experts $E_1, \ldots, E_n$ and a router that selects a subset of the experts to apply to each token. 
For each hidden token representation $h$, the router
produces a score $s_i(h)$ for each expert $i$ and returns a weighted sum of the top-$k$ experts:
\[
  y \;=\; \sum_{i \in \mathrm{TopK}(s(h))} g_i(h)\, E_i(h),
\]
where $g_i$ are the router's softmax scores of the selected experts. 
Because not all parameters are active for each token, MoEs decouple \emph{total} from \emph{active} parameters. 

Recent MoEs employ \emph{fine-grained} experts \citep{deepseek}, where the \emph{granularity} is defined as the ratio of expert-FFN to dense-FFN dimensions. For example, an MoE with granularity $g=\frac{1}{2}$ has expert FFNs with $\frac{1}{2}$ the intermediate dimension of a dense FFN.
MoE architectures may vary the expert granularity $g$, the total count of experts $n$, and the active count $k$. These design choice axes complicate the study of MoEs, as active parameter count and FLOPS-per-token change with each configuration.

\subsection{Data Repetition}
\label{sec:bg_repetition}

In a \emph{data-constrained} scenario, it is common to repeat some or all of the training data. A simple approach might iterate over the entire training data corpus of $U$ unique tokens over multiple passes, or \emph{epochs}, until the desired total token budget $T$ is reached. This requires a repetition rate of $R = T/U$. In other cases, training data may be defined as a mix of various data domains combined in fixed percentages. If only a subset of the domains are data-constrained, it is common to set domain-specific repetition rates $R_i$ for each domain $i$ \citep{soldaini2024dolma,olmoe}.

\subsection{Regularization Methods}
\label{sec:bg_reg}

Known remedies for overfitting trade off training fit to recover generalization:
Dropout \citep{srivastava2014dropout} randomly drops units during training, which prevents them from co-adapting to form complex memorization patterns.
Weight decay, in the decoupled form used by AdamW \citep{loshchilov2019decoupled}, shrinks parameters towards zero independently of the gradient. Gradient norm clipping \citep{pascanu2013difficulty} rescales gradients whose norm exceeds a threshold, stabilizing the effect of any single datapoint. 
Some regularization methods target MoE components: Router jitter injects multiplicative noise into routing, and expert dropout applies a separate, often larger dropout rate inside experts \citep{switch,stmoe}. We evaluate regularization methods under data repetition in \S\ref{sec:reg}.

\providecommand{\figph}[3]{%
  \begin{figure}[!t]
  \centering
  \fbox{\parbox{0.92\linewidth}{%
    \vspace{0.7em}%
    \centering\textbf{[FIGURE PLACEHOLDER]}\\[0.5em]%
    \raggedright\small #3%
    \vspace{0.7em}}}
  \caption{#2}
  \label{#1}
  \end{figure}}

\section{Experiments}
\label{sec:expts}

\paragraph{Models.} We train compute-matched densely- and sparsely-activated (MoE) Transformer LMs in the style of \citet{olmoe}. Specifically, we train models with 80 million, 200 million, and 1 billion active parameters, denoted as 80M, 200M, and 1B, respectively. We vary our MoE model configurations to study the effect of \emph{total expert count} and \emph{expert granularity}. Our MoE models have $n\in\{8, 16, 32, 64, 128\}$ total experts with granularity $g\in\{\frac{1}{2},\frac{1}{4}, \frac{1}{8}, \frac{1}{16}, \frac{1}{32}\}$. To match active parameters, we always set the top-$k$ activation to $k=\frac{1}{g}$, resulting in sparsity $s = \frac{n}{k} = n\cdot g$.
We denote our models \emph{MoE (n x g)}, e.g., \emph{MoE (64 x 1/4)} 
represents 4 active experts out of 64 total with granularity 1/4.
Additional model architecture details are in Appendix~\ref{app:model_arch}.

\paragraph{Data Repetition.} Following common practice from \citet{chinchilla}, our models are trained with total training data tokens $T \approx 20 \cdot N_a$, where $N_a$ denotes the number of \emph{active} parameters.
We only vary the tokens-per-active-parameter $T/N_a$ ratio in \S\ref{sec:expts_olmo_mix} to study its interaction with sparsity and data repetition. Under a fixed total data $T$, we vary the number of \emph{unique} tokens $U$. Each model is trained by iterating through its allotted $U$ unique tokens $R = T/U$ times. We consider subsets of $R\in\{1, 2, 4, 8, 16, 32, 64, 128, 256, 512, 1024\}$.

For any unique-token budget $U$, we construct the training set $\mathcal{D}_U$ by taking the first $U$ tokens from a fixed random permutation of the full dataset $\mathcal{D}$. We use the same permutation across all experiments, ensuring that training sets are nested: for any $U_1 \leq U_2$, $\mathcal{D}_{U_1} \subseteq \mathcal{D}_{U_2}$
Additional data repetition details are in Appendix~\ref{app:data_rep}.

\paragraph{Training Data.} 
We train on constituent domains from the OLMoE \citep{olmoe} data mix, both individually and in custom mixes: web crawl data (DCLM; \citet{li2024datacomplm}), code (Starcoder; \citet{li2023starcoder}), scientific text (peS2o; \citet{peS2o}), and encyclopedic text (Wikipedia; \citet{soldaini2024dolma}). See Appendix~\ref{app:train_data_sources}.

\paragraph{Evaluation.} We report Cross-Entropy Loss (CE Loss) on various held-out data, including web crawl, code, scientific text, and encyclopedic text. We also measure CE Loss and accuracy on downstream tasks. Results on downstream tasks are in Appendix~\ref{app:downstream_tasks}. We show random seed sensitivity in Appendix~\ref{app:random_seed}. Additional details in Appendix~\ref{app:eval_data_sources}.

\subsection{Data Repetition Effects across MoE Architectures}
\label{sec:expts_olmo_mix}

We train a variety of dense and MoE Transformer LMs on the OLMoE data mix to investigate the effects of data repetition. Specifically, we fix the \emph{total} train token budget $T = 20 \cdot N_{a}$, but vary the repetition rate $R$, so that \emph{unique} tokens $U$ decreases as $R$ increases to satisfy $R\cdot U = T$ for fixed $T$. 

\paragraph{MoEs degrade more than dense LMs under data repetition.} Figure~\ref{fig:olmoe_mix} shows the effect of data repetition on validation loss. Across model scales, dense Transformer performance degrades at high data repetition rates, with validation loss rising slightly at $R=8$ and sharply at $R > 64$. MoEs respond more dramatically to data repetition, with a noticeable performance impact at $R=4$ and a sharper increase at higher $R$. Though MoEs outperform dense models at $R\leq 16$, the catastrophic effect of data repetition leads to a reversal at $R = 32$, where dense models outperform MoEs. Other held-out LM tasks (Appendix~\ref{app:extra_lm}) and downstream tasks (Appendix~\ref{app:downstream_tasks}) exhibit similar trends.

\paragraph{At extremely high repetition rates, validation loss lowers again.} Results in Figure~\ref{fig:olmoe_mix} demonstrate a least optimal data repetition rate; validation loss rises with $R$ until this point, then falls.
We also consider much higher $R\in\{2^{15}, 2^{20}\}$, and find that validation loss plateaus, then may rise yet again (Appendix Figure~\ref{fig:extended_app}).

\paragraph{Training loss falls to zero at high repetition rate.} As shown in Figure~\ref{fig:train_v_valid}, training loss decreases with increased repetition rate, falling under 1E-2 at $R=512$ for 80M dense, and $R=160$ for 200M dense models. Training loss decrease is a reflection of validation loss increase, which agrees with other indications of overfitting via training data memorization. Additional model scales in Appendix Figure~\ref{fig:train_valid_app}.

\paragraph{Sparsity increases overfitting; data repetition effects depend on \emph{total} parameters.} We vary the number of total experts and the expert granularity in Figure~\ref{fig:expert_count_granularity}. In both cases, increased \emph{total} parameters results in a sharper response to data repetition. As the \emph{active} parameters remain fixed throughout these experiments, we hypothesize that this phenomenon is related to the unique tokens to \emph{total} parameters ratio. Thus, we compare the 200M dense models to the 80M MoE (32 x 1/4) and MoE (64 x 1/4) models, which have 158M and 244M \emph{total} parameters, 
respectively. In Figure~\ref{fig:olmoe_mix}, the data repetition effects of the 200M model appear to lie between those of the MoE (32 x 1/4) and MoE (64 x 1/4) models at 80M active parameters. 

\paragraph{Data repetition effects do not depend on \emph{total} data budget.} 

We train 80M active parameter models with $4\times$ more data, resulting in a total-tokens-to-active-parameter ratio $T/N_{a} = 80$. The resulting trends (Figure~\ref{fig:data_per_params}) are very similar to those with $T/N_{a} = 20$ (Figure~\ref{fig:olmoe_mix}). Thus, the data repetition overfitting response may change only slowly with the total token budget, but instead depend most directly on total parameters and repetition count.

\ifarxiv
\begin{figure*}[!t]
    \centering
    \begin{subfigure}[t]{\textwidth}
        \begin{subfigure}[t]{\textwidth}
            \centering
            \includegraphics[width=\linewidth]{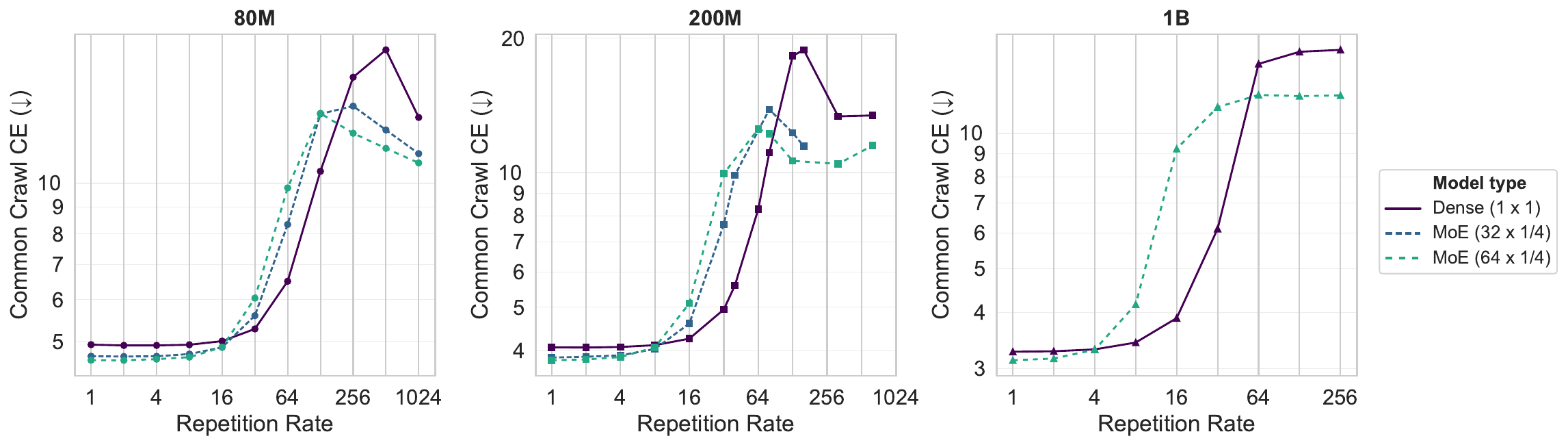}
        \end{subfigure}
    \end{subfigure}

    \caption{\textbf{Across active parameter scales, data repetition rates over $8\times$ result in increasingly severe overfitting. Sparser models overfit more (\S\ref{sec:expts_olmo_mix}).} At 80M, 200M, and 1B active parameters, we fix the total data budget $T = 20 \cdot N_{a}$, and vary the data repetition rate $R$ via different sized unique token sets. As $R$ increases, models increasingly overfit. Sparsity exacerbates overfitting behavior. 
    Larger and sparser models overfit more at lower $R$.}
    \label{fig:olmoe_mix}
\end{figure*}

\begin{figure*}[!ht]
    \centering
    \begin{subfigure}[t]{\textwidth}
        \begin{subfigure}[t]{\textwidth}
            \centering
            \includegraphics[width=\linewidth]{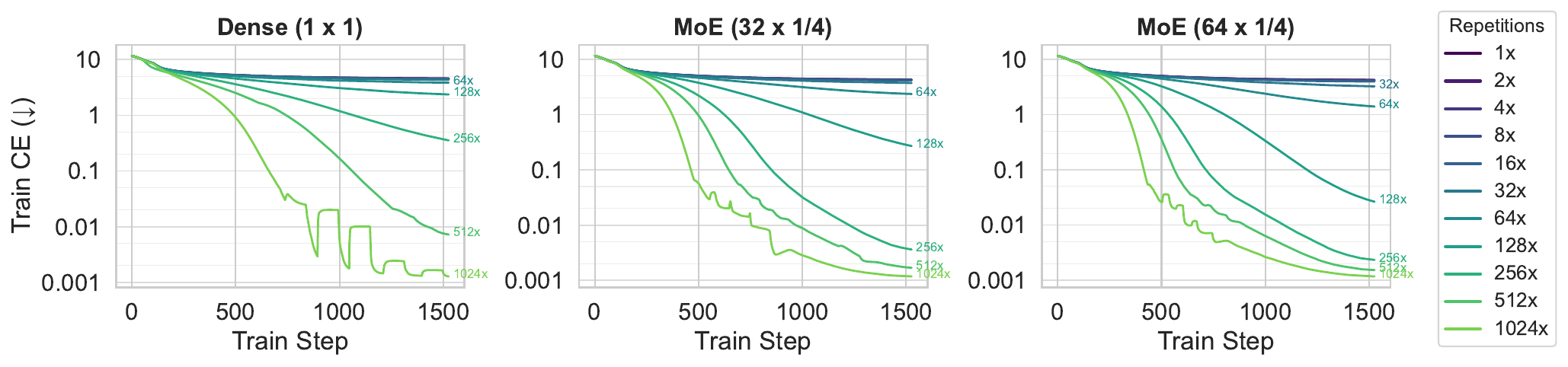}
        \end{subfigure}
        \par\vspace{1em}
        \begin{subfigure}[t]{\textwidth}
            \centering
            \includegraphics[width=\linewidth]{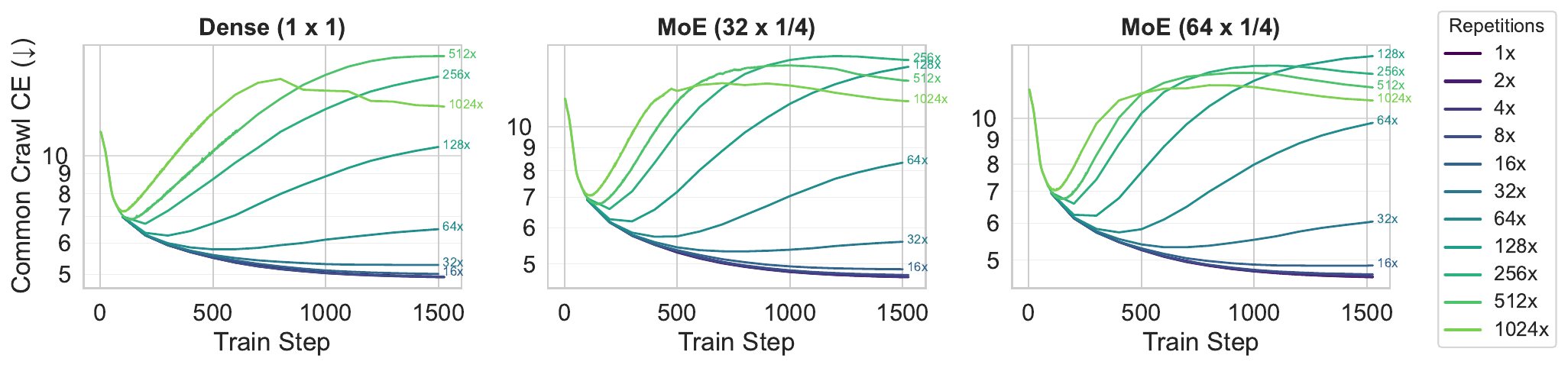}
        \end{subfigure}
    \end{subfigure}

    \caption{\textbf{At higher repetition rates, training loss falls to 0 as models overfit to the repeated data (\S\ref{sec:expts_olmo_mix}).}
    Early in training, train (above) and validation (below) 
    loss fall together, but, at high repetition rate $R$, train loss falls rapidly to 0, indicating memorization of training data, while validation loss rises. Additional model scales in Appendix Figure~\ref{fig:train_valid_app}.}
    \label{fig:train_v_valid}
\end{figure*}

\begin{figure*}[!t]
    \centering
    \begin{subfigure}[t]{\textwidth}
        \begin{subfigure}[t]{\textwidth}
            \centering
            \includegraphics[width=0.99\linewidth]{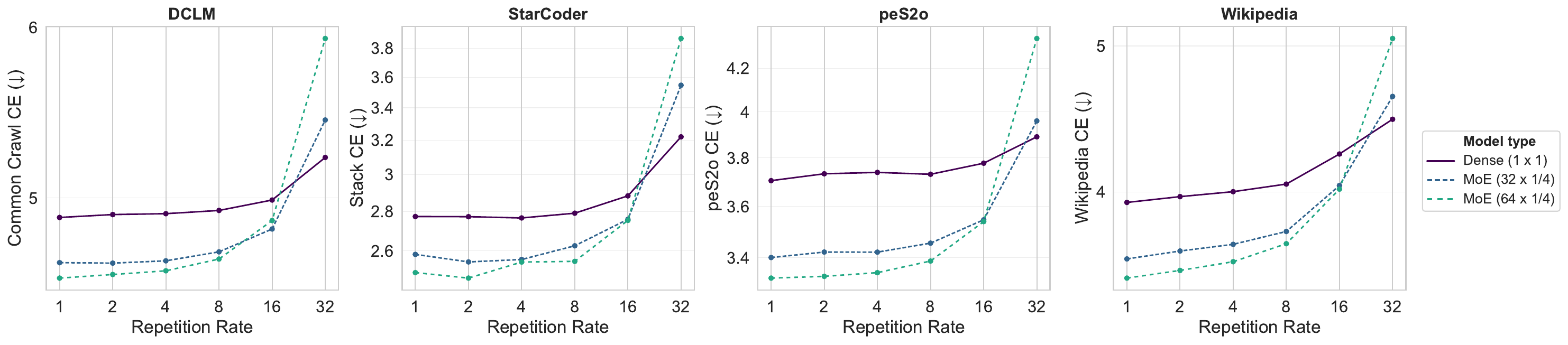}
        \end{subfigure}
    \end{subfigure}

    \caption{\textbf{Across all train domains, sparse MoE models overfit more, ceding their benefit over dense models as repetition rates rise (\S\ref{sec:expts_single_domain}).} We train dense and MoE models on a single domain. Our experiments span web crawl (DCLM), code (StarCoder), academic (peS2o), and encyclopedic (Wikipedia) text. Trends are near-identical across all data domains: models underperform as repetition rates rise, with MoEs overfitting more rapidly.
    }
     \label{fig:single_domain}
\end{figure*}

\else
\begin{figure*}[!t]
    \centering
    \begin{subfigure}[t]{\textwidth}
        \begin{subfigure}[t]{\textwidth}
            \centering
            \includegraphics[width=\linewidth]{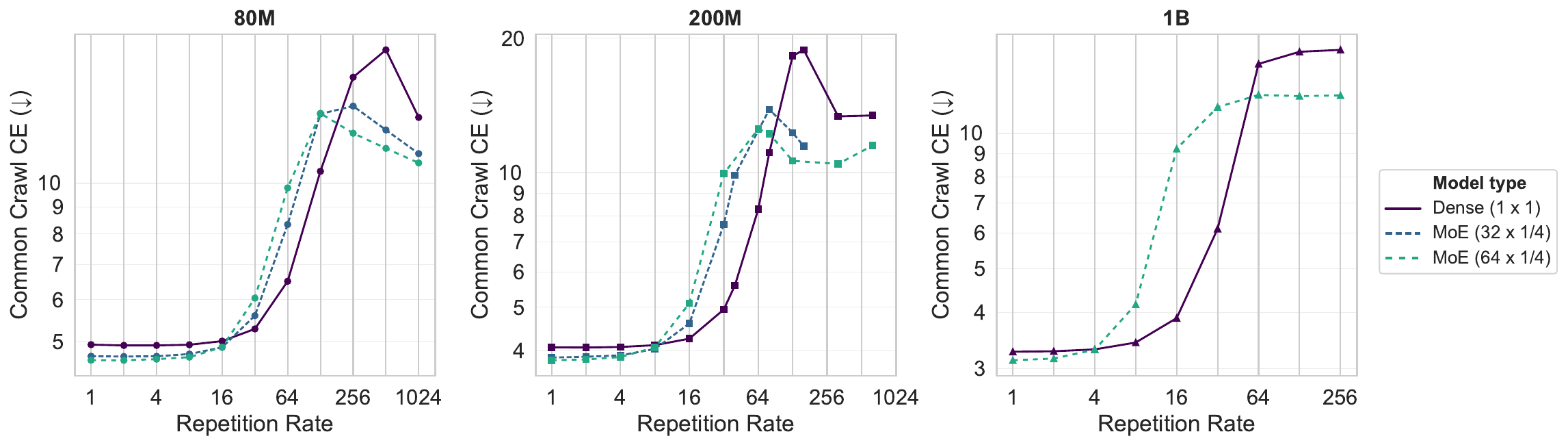}
        \end{subfigure}
    \end{subfigure}

    \caption{\textbf{Across active parameter scales, data repetition rates over $8\times$ result in increasingly severe overfitting. Sparser models overfit more (\S\ref{sec:expts_olmo_mix}).} At 80M, 200M, and 1B active parameters, we fix the total data budget $T = 20 \cdot N_{a}$, and vary the data repetition rate $R$ via different sized unique token sets. As $R$ increases, models increasingly overfit. Sparsity exacerbates overfitting behavior. 
    Larger and sparser models overfit more at lower $R$.}
    \label{fig:olmoe_mix}
\end{figure*}

\begin{figure*}[!t]
    \centering
    \begin{subfigure}[t]{\textwidth}
        \begin{subfigure}[t]{\textwidth}
            \centering
            \includegraphics[width=0.99\linewidth]{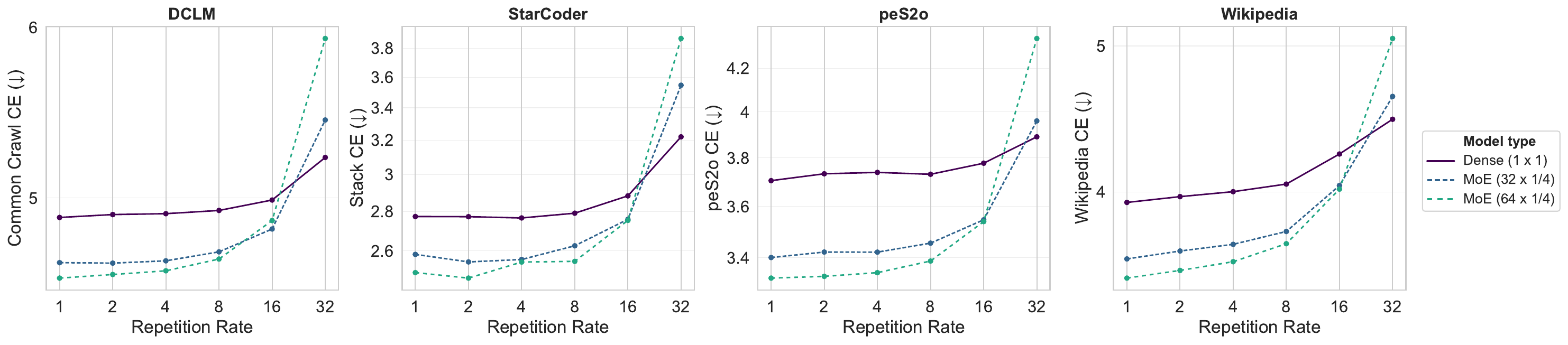}
        \end{subfigure}
    \end{subfigure}

    \caption{\textbf{Across all train domains, sparse MoE models overfit more, ceding their benefit over dense models as repetition rates rise (\S\ref{sec:expts_single_domain}).} We train dense and MoE models on a single domain. Our experiments span web crawl (DCLM), code (StarCoder), academic (peS2o), and encyclopedic (Wikipedia) text. Trends are near-identical across all data domains: models underperform as repetition rates rise, with MoEs overfitting more rapidly.
    }
     \label{fig:single_domain}
\end{figure*}

\fi

\subsection{Data Repetition Effects across Data Domains}
\label{sec:expts_single_domain}

We repeat the experimental setup in \S\ref{sec:expts_olmo_mix} with individual data domains from the OLMoE data mix: DCLM (web crawl), peS2o (academic), StarCoder (code), and Wikipedia (encyclopedic) text. Our goal is to understand variations in response to repetition across varied data domains. We focus on $R\in\{1,2,4,8,16,32\}$.

\paragraph{Patterns are largely consistent across single-domain experiments.} Results in Figure~\ref{fig:single_domain} indicate that overfitting patterns from data repetition are robust across domains, despite the diversity of data. Code, encyclopedic, web crawl, and academic text are semantically very distinct, yet all four exhibit similar patterns, for example, that MoE models begin to underperform dense models at $R\in[16,32]$.

\ifarxiv
    \begin{figure*}[!t]
    \centering
    \begin{subfigure}[t]{\textwidth}
        \begin{subfigure}[t]{0.5\textwidth}
            \centering
            \includegraphics[width=\linewidth]{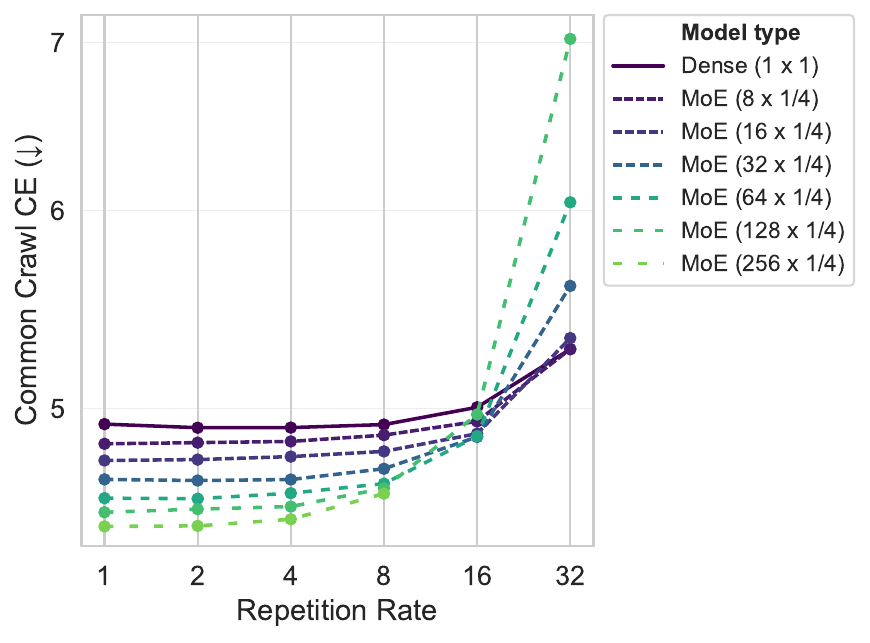}
            \subcaption{Varied Total Expert Count ($n$)}
        \end{subfigure}
        \begin{subfigure}[t]{0.5\textwidth}
            \centering
            \includegraphics[width=\linewidth]{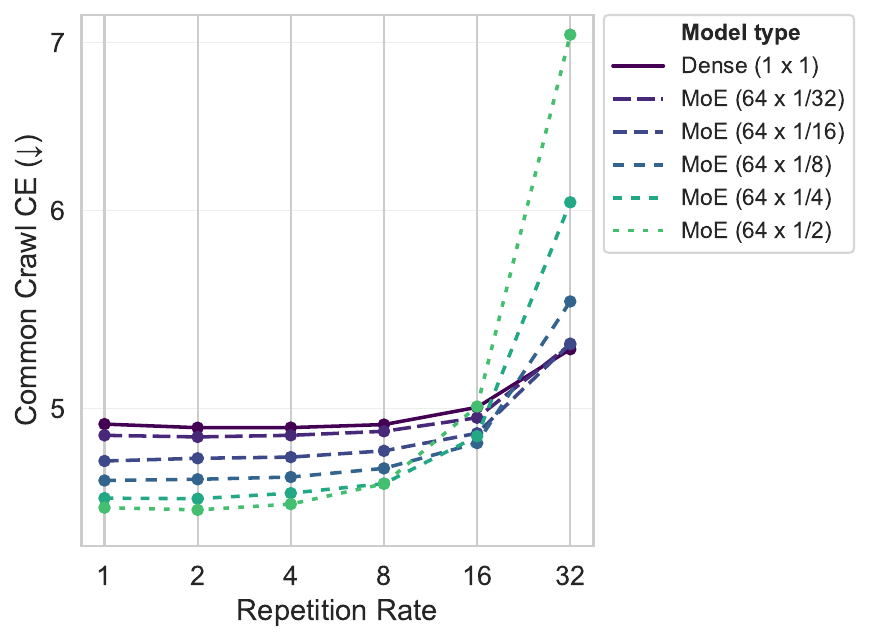}
            \subcaption{Varied Expert Size / Granularity ($g$)}
        \end{subfigure}
    \end{subfigure}
    \caption{\textbf{In MoEs, higher sparsity results in a greater reaction to data repetition (\S\ref{sec:expts_olmo_mix}).} We consider two settings: (a) fixed expert granularity, increased sparsity via greater total expert count; (b) fixed total expert count, increased sparsity via larger experts. Overfitting increases with total parameters, corresponding to (a) more total experts and (b) larger experts. Lines across (a) and (b) with the same color have identical total parameter count.
    }
    \label{fig:expert_count_granularity}
\end{figure*}

    \begin{figure*}[!t]
    \centering
    \begin{subfigure}[t]{\textwidth}
        \begin{subfigure}[t]{\textwidth}
            \centering
            \includegraphics[width=\linewidth]{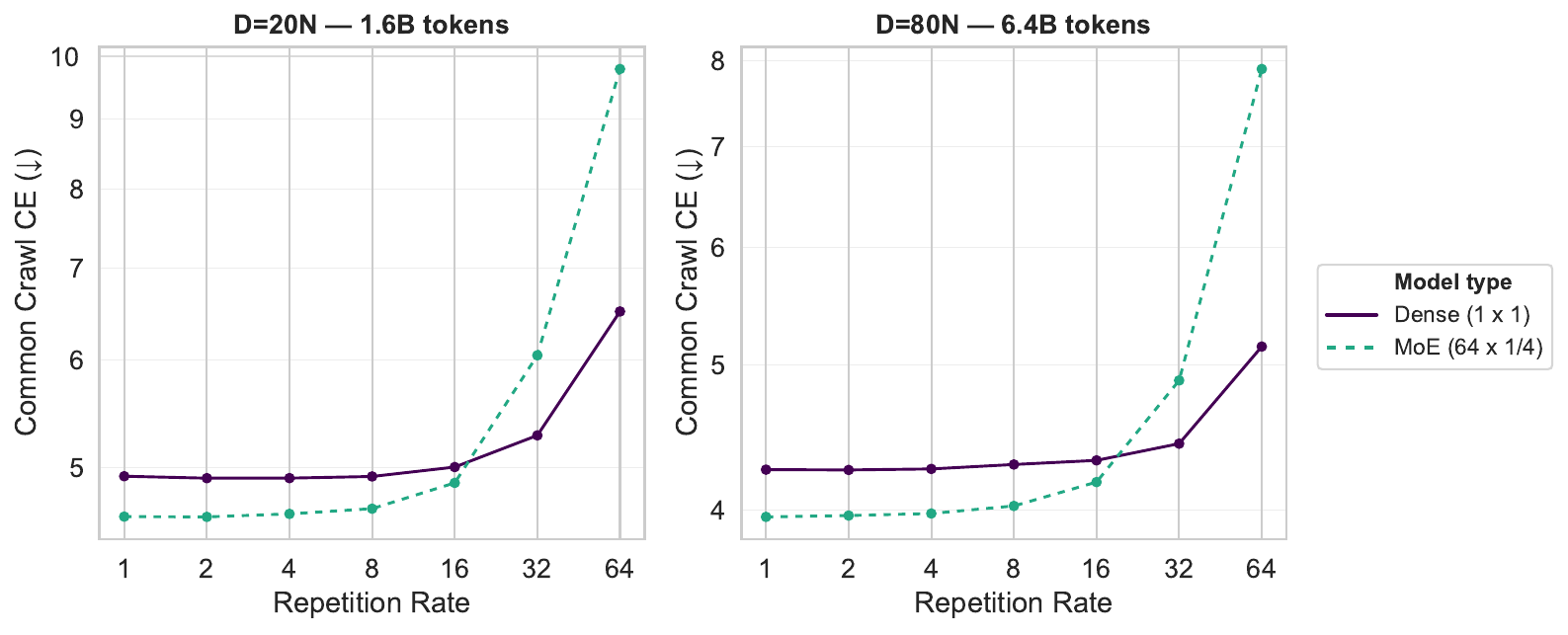}
        \end{subfigure}
    \end{subfigure}
    \caption{\textbf{Data repetition effects are consistent across data-to-parameter ratios (\S\ref{sec:expts_olmo_mix}).} We increase total tokens per active parameter from 20 to 80, and observe no difference in data repetition overfitting between the two token-to-active-parameter ratios.}
    \label{fig:data_per_params}
\end{figure*}

\else
    \begin{figure*}[!t]
    \centering
    \begin{subfigure}[t]{\textwidth}
        \begin{subfigure}[t]{0.5\textwidth}
            \centering
            \includegraphics[width=\linewidth]{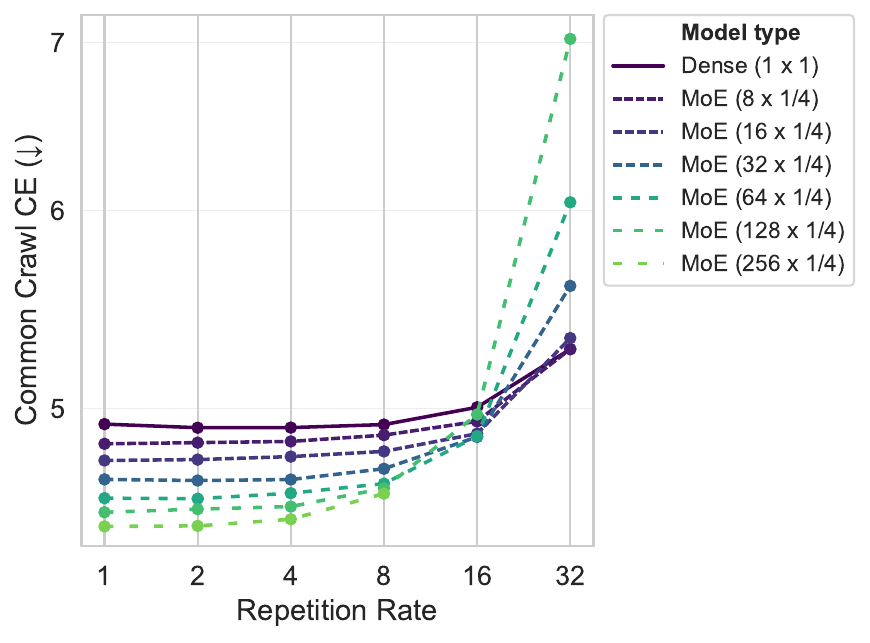}
            \subcaption{Varied Total Expert Count ($n$)}
        \end{subfigure}
        \begin{subfigure}[t]{0.5\textwidth}
            \centering
            \includegraphics[width=\linewidth]{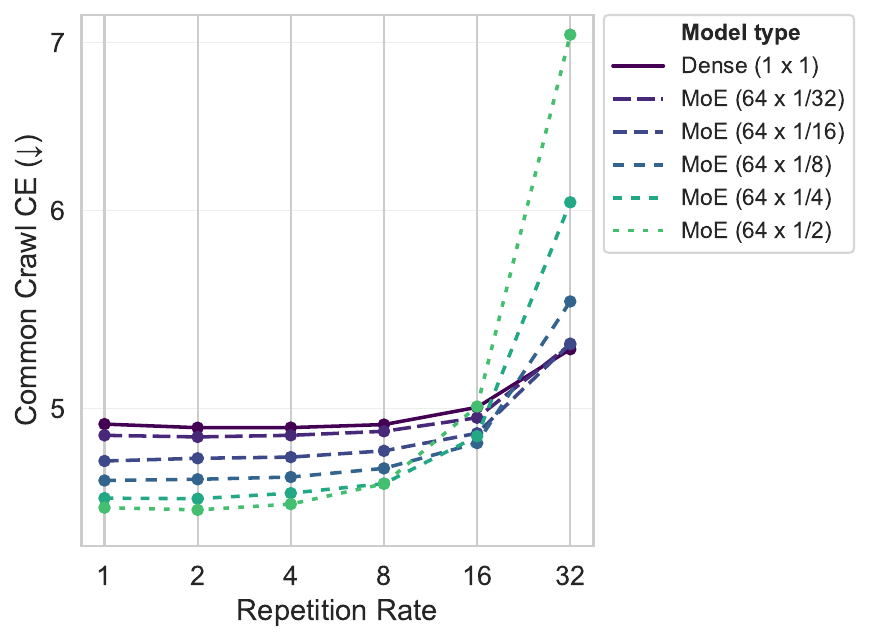}
            \subcaption{Varied Expert Size / Granularity ($g$)}
        \end{subfigure}
    \end{subfigure}
    \caption{\textbf{In MoEs, higher sparsity results in a greater reaction to data repetition (\S\ref{sec:expts_olmo_mix}).} We consider two settings: (a) fixed expert granularity, increased sparsity via greater total expert count; (b) fixed total expert count, increased sparsity via larger experts. Overfitting increases with total parameters, corresponding to (a) more total experts and (b) larger experts. Lines across (a) and (b) with the same color have identical total parameter count.
    }
    \label{fig:expert_count_granularity}
\end{figure*}

\fi

\subsection{Data Repetition Effects Under Data Filters}
\label{sec:expts_filtering}

\ifarxiv
    \begin{figure*}[!t]
    \centering
    \begin{subfigure}[t]{\textwidth}
        \centering
        \includegraphics[width=0.99\textwidth]{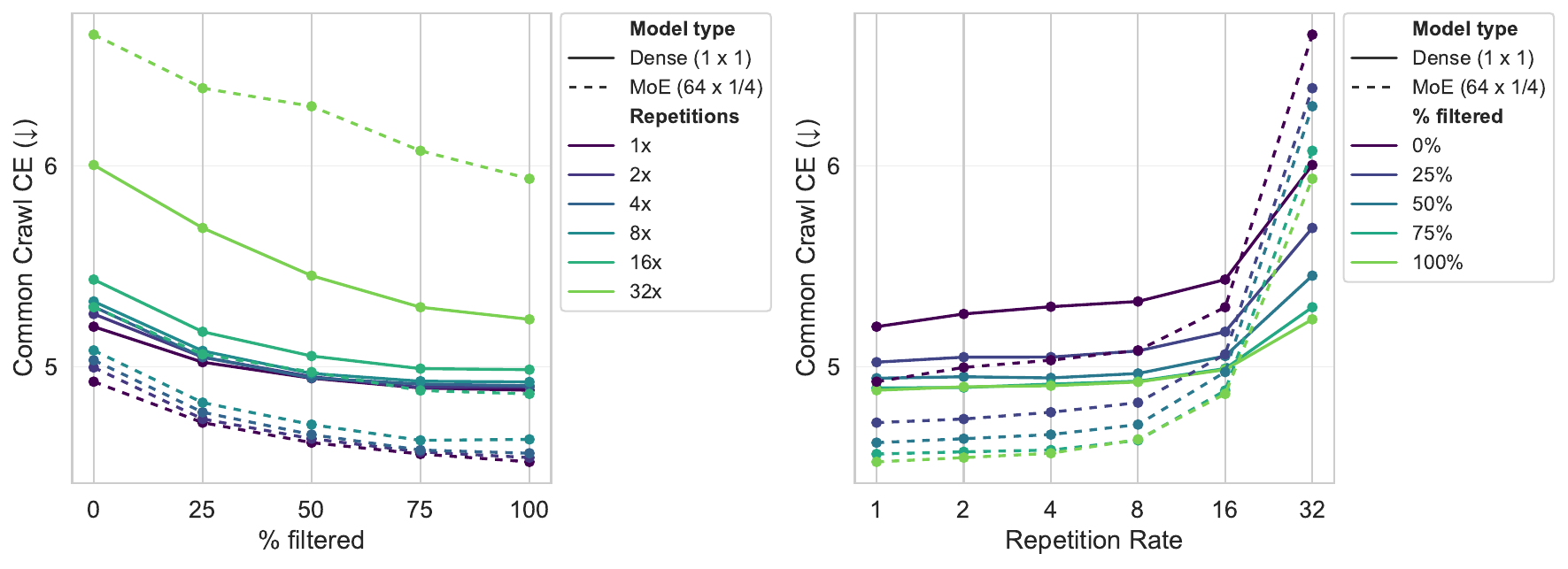}
    \end{subfigure}

    \caption{\textbf{Repetition effects are consistent across data quality (\S\ref{sec:expts_filtering}).} 
    Validation CE Loss for 80M dense and MoE (64 x 1/4) on the  \textsc{DCLM-pool} and  \textsc{-baseline}  interpolation, where \emph{\% filtered} indicates the proportion of the data mix dedicated to the heavily filtered \textsc{DCLM-baseline} data. Left: CE against the \textsc{DCLM-baseline} \%. A higher percentage of filtered data results in better performance. As in \S\ref{sec:expts_olmo_mix}-\ref{sec:expts_single_domain}, MoE performance deteriorates more rapidly than dense at higher $R$, regardless of data. Right: CE Loss against repetition rate. Trends are largely similar across data quality and model architecture, though dense models trained on higher proportions of unfiltered data (0-50\% \textsc{DCLM-baseline}) may degrade more rapidly with $R$.
    }
    \label{fig:filtering}
\end{figure*}

\else
    \begin{figure}[!t]
    \centering
    \begin{minipage}{0.58\textwidth}
    \begin{subfigure}[t]{\textwidth}
        \begin{subfigure}[t]{\textwidth}
            \centering
            \includegraphics[width=\linewidth]{fig_pdfs/paper/budget_1C_vs_4C__80M__dense_moe64__eval_dolma_cc_loss.pdf}
        \end{subfigure}
    \end{subfigure}
    \caption{\textbf{Data repetition effects are consistent across data-to-parameter ratios (\S\ref{sec:expts_olmo_mix}).} At 80M active parameter scale, we increase total tokens per active parameter from 20 to 80. We observe very similar patterns in data repetition overfitting between the two token-to-active-parameter ratios.}
    \label{fig:data_per_params}
    \end{minipage}\hfill
    \begin{minipage}{0.41\textwidth}
        \centering
    \begin{subfigure}[t]{\textwidth}
        \centering
        \includegraphics[width=0.85\textwidth]{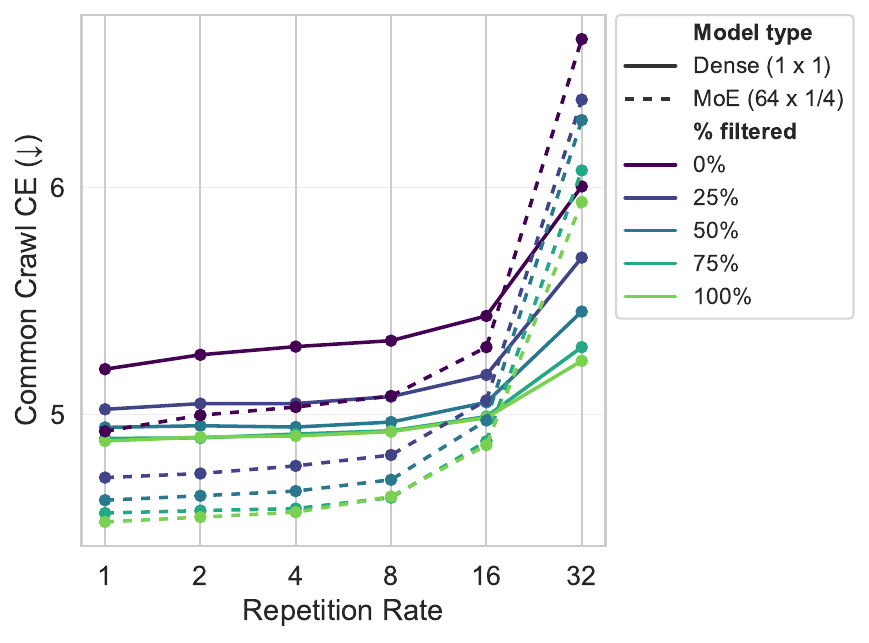}
    \end{subfigure}

    \caption{\textbf{Repetition effects are consistent across data quality (\S\ref{sec:expts_filtering}).} 
    Interpolating between raw \textsc{DCLM-pool} and  heavily-filtered \textsc{DCLM-baseline}:
    Trends are similar across data quality. As in \S\ref{sec:expts_olmo_mix}-\ref{sec:expts_single_domain}, MoEs consistently deteriorate more rapidly than dense at high $R$.
    }
    \label{fig:filtering}
    \end{minipage}
\end{figure}

\fi

We also measure the impact of quality filtering on data repetition. It is common practice to preprocess raw web crawls to deduplicate, extract text from HTML, and remove data deemed \emph{low quality} by pre-defined filters, often significantly reducing the available tokens. \textsc{DCLM-baseline} 
retains 2.4\% of the raw 280T-token \textsc{DCLM-pool} \citep{li2024datacomplm}. Under data constraints, filtering may remove lower quality tokens, but also further reduces the quantity of unique tokens. We study this tradeoff between token quality and repetition rate by considering 5 data settings, which mix \textsc{DCLM-baseline} with \textsc{DCLM-pool}.
We interpolate between all-unfiltered and all-filtered with 5 settings consisting of the following DCLM (baseline \%, pool \%): (0\%, 100\%), (25\%, 75\%), (50\%, 50\%), (75\%, 25\%), (100\%, 0\%).

\paragraph{Filtering affects data repetition in some dense models.} Training on a higher percentage of unfiltered \textsc{DCLM-pool} data results in consistently worse performance (Figure~\ref{fig:filtering}).  Despite the distribution shift that arises from stringent filtering, all MoEs trained on any mixture of \textsc{DCLM-pool} and \textsc{-baseline} exhibit similar decline with data repetition. However, dense models trained exclusively on at least 50\% unfiltered data may deteriorate more rapidly with $R$. As \textsc{DCLM-baseline} filtering reduces \textsc{DCLM-pool} by a factor of over $40\times$, we compare $R=1$ on 100\% raw \textsc{DCLM-pool} against $R=32$ on 100\% \textsc{DCLM-baseline}. In our setting, training on all unique unfiltered data is superior to repeating strictly filtered data, but an intermediate quality filter and repetition rate is likely superior to either extreme.

\subsection{Data Repetition At Mixed Rates Within Diverse Data Mixes}
\label{sec:expts_mixes}
\begin{figure*}[!t]
    \centering
    \begin{subfigure}[t]{0.75\textwidth}
        \begin{subfigure}[t]{\textwidth}
            \centering
            \includegraphics[width=\linewidth]{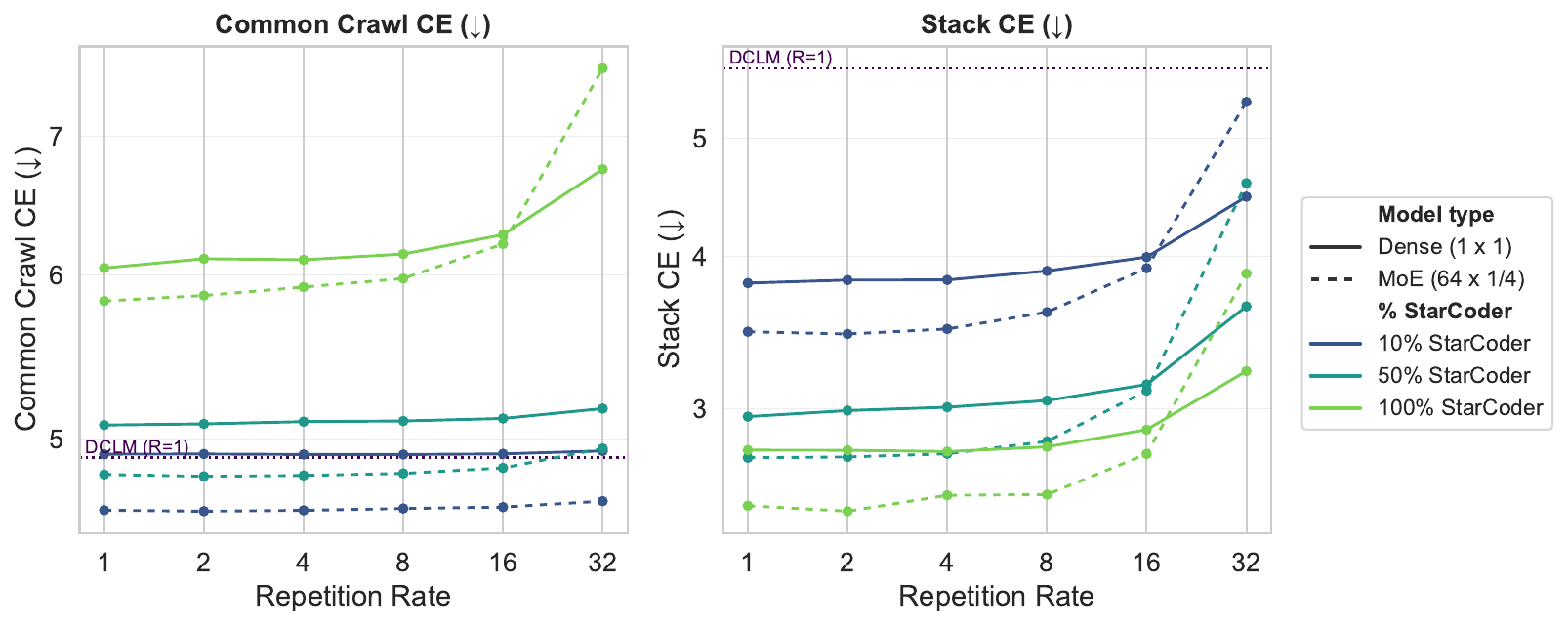}
        \end{subfigure}
        \subcaption{DCLM + StarCoder}
        \label{fig:mix_rep_rate_starcoder}
    \end{subfigure} 
    \par\vspace{1em}
    \begin{subfigure}[t]{0.75\textwidth}
        \begin{subfigure}[t]{\textwidth}
            \centering
            \includegraphics[width=\linewidth]{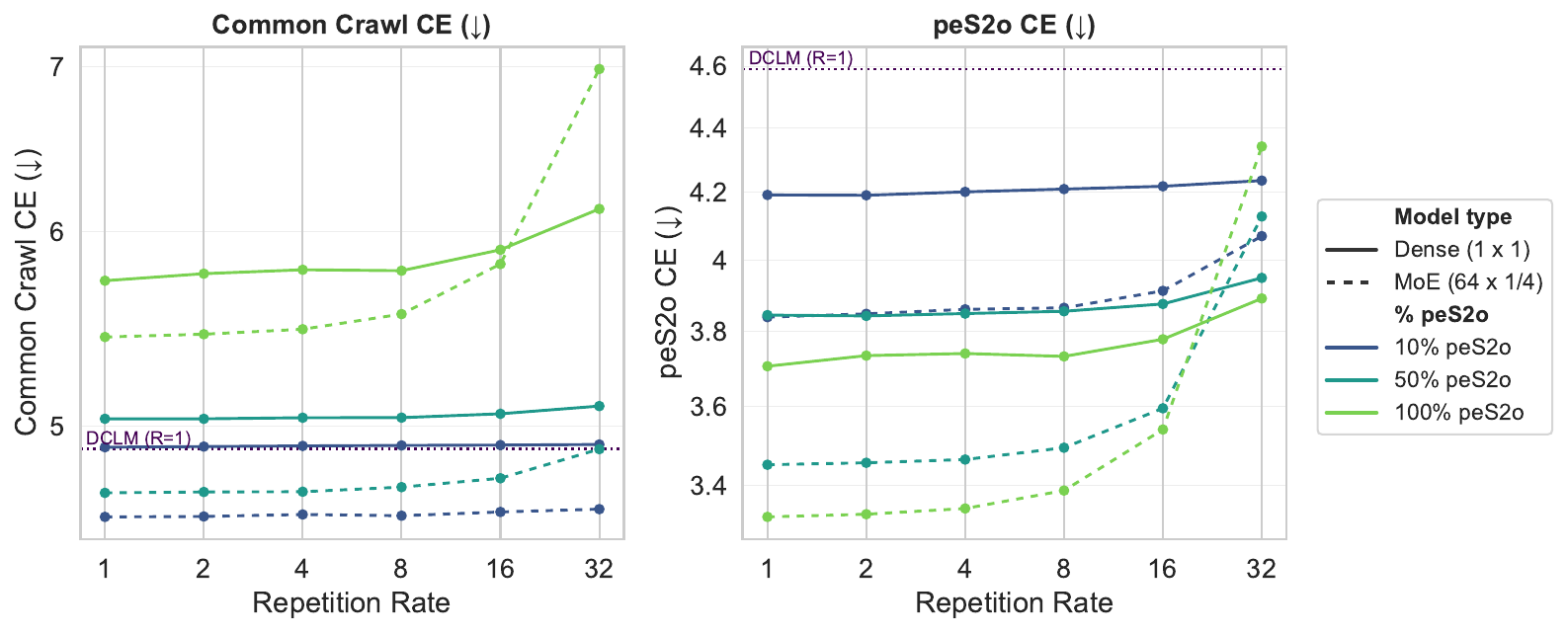}
        \end{subfigure}
        \subcaption{DCLM + peS2o}
        \label{fig:mix_rep_rate_pes2o}
    \end{subfigure}
    
    \caption{\textbf{Mixing a repeated data domain with a non-repeated domain may have a regularizing effect (\S\ref{sec:expts_mixes}).}
    We mix peS2o and StarCoder, repeated $R\in\{1, 2, 4, 8, 16, 32\}$ times, into non-repeated DCLM, with various proportions of peS2o/StarCoder. We find that DCLM + StarCoder mixes degrade similarly with repetition, regardless of the StarCoder mixing percentage. However, DCLM appears to have a regularizing effect in DCLM + peS2o mixes: as DCLM takes up a larger proportion of the mix, high repetition rates ($R=32$) degrade at a slower rate.
    }
    \label{fig:mix_rep_rate}
\end{figure*}

In practice, LMs are trained on data mixes  of diverse domains at varying levels of repetition. For example, encyclopedic text is often more constrained than web crawl data, and is therefore repeated at a higher rate. For example, GPT-3 was trained for 3.4 epochs over its Wikipedia component, but less than 1 epoch over Common Crawl \citep{brown2020language}.

To investigate the effects of varied repetition rates within a data mix, we consider a simple setting in which the total data budget $T$ is split between two domains. The first domain is DCLM, which is never repeated. The second is peS2o or StarCoder, which is repeated with $R\in\{1,2,4,8,16,32\}$. We divide the total data budget $T$ between the 2 domains (DCLM, peS2o) or (DCLM, StarCoder) at the following proportions: (100\%, 0\%), (90\%, 10\%), (50\%, 50\%), or (0\%, 100\%). Only the repeated domain's unique pool shrinks as $R$ grows. As a concrete example, the (50\%, 50\%) DCLM-StarCoder setting at $R=8$ has $T=1.6B$ total tokens, consisting of $1.6B*0.5=0.8B$ unique DCLM tokens and $1.6B*0.5/8=0.1B$ unique StarCoder tokens repeated 8 times.

\paragraph{Mixing repeated StarCoder with all-unique DCLM does not impact repetition effects.} In Figure~\ref{fig:mix_rep_rate_starcoder}, we find that, regardless of the proportion of $T$ assigned to StarCoder, code validation loss rises according to the same pattern in \S\ref{sec:expts_olmo_mix}-\ref{sec:expts_single_domain}. Web crawl validation loss only rises with the repetition rate of StarCoder if it occupies at least half of $T$. 

\paragraph{Mixing repeated peS2o with all-unique DCLM may have a regularizing effect.} In Figure~\ref{fig:mix_rep_rate_pes2o}, academic text validation loss increases with the peS2o repetition rate, but the effect is substantially dampened as peS2o occupies a smaller proportion of the total training data. As with the DCLM-StarCoder experiments above, web crawl validation loss suffers from peS2o repetition only when the peS2o total data proportion is high. This suggests that the unrepeated DCLM data component may have a regularizing effect on the peS2o data repetition.

\paragraph{
Mixing repeated data with unrepeated data of a semantically similar domain may reduce data repetition effects.} Academic text, as represented by peS2o, has higher semantic similarity to web text than code, as represented by StarCoder. Our results suggest a promising possibility: even at very high repetition rates, repetition effects might be reduced by mixing the repeated data domain with an non-repeated or less-repeated, semantically similar data domain.

\providecommand{\figph}[3]{%
  \begin{figure}[t]
  \centering
  \fbox{\parbox{0.92\linewidth}{%
    \vspace{0.7em}%
    \centering\textbf{[FIGURE PLACEHOLDER]}\\[0.5em]%
    \raggedright\small #3%
    \vspace{0.7em}}}
  \caption{#2}
  \label{#1}
  \end{figure}}

\providecommand{\propbox}[2]{%
  \begin{center}
  \fbox{\parbox{0.92\linewidth}{%
    \vspace{0.5em}%
    \centering\textbf{[PROPOSED EXPERIMENT: #1]}\\[0.5em]%
    \raggedright\small #2%
    \vspace{0.5em}}}
  \end{center}}

\section{Regularization Techniques for Data Repetition}
\label{sec:reg}

\ifarxiv 
    \begin{figure*}[!t]
    \centering
    \begin{subfigure}[t]{0.75\textwidth}
        \begin{subfigure}[t]{\textwidth}
            \centering
            \includegraphics[width=\linewidth]{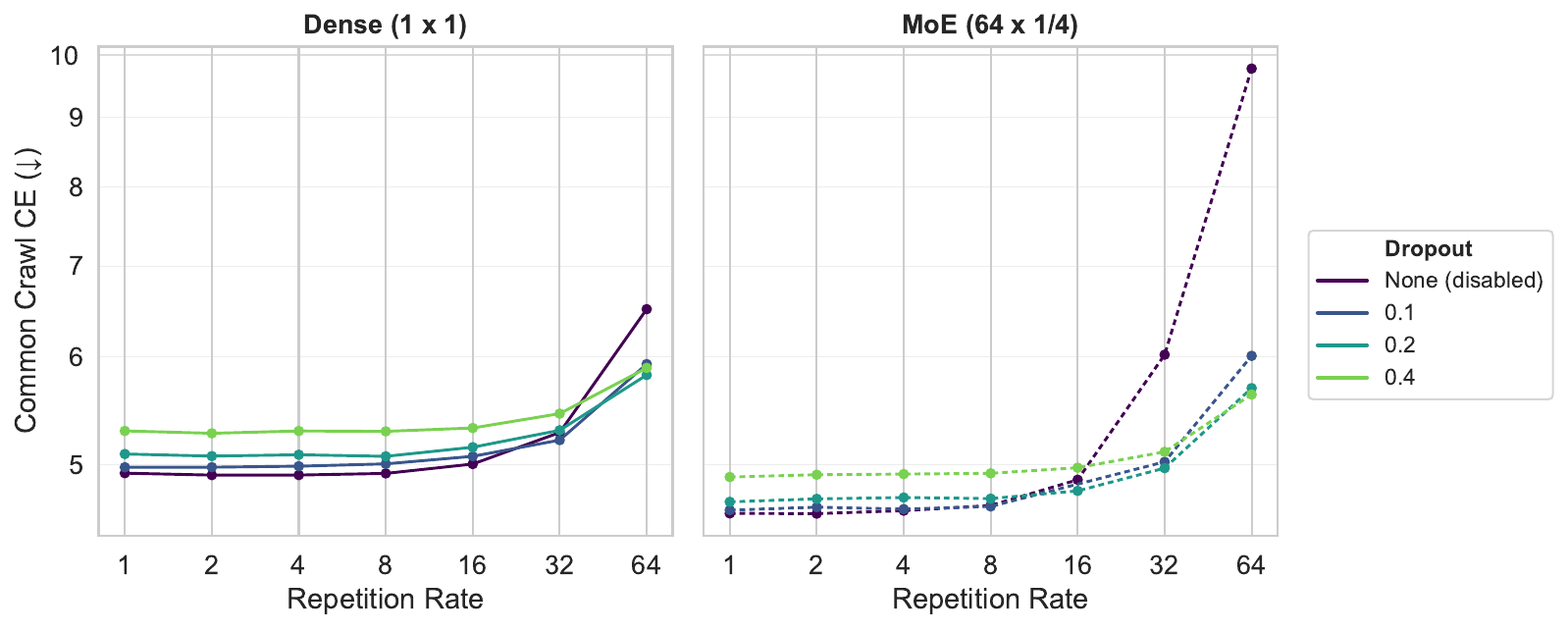}
        \end{subfigure}
        \subcaption{Dropout}
        \label{fig:regularizers:dropout}
    \end{subfigure}
        \begin{subfigure}[t]{0.75\textwidth}
            \centering
            \includegraphics[width=\linewidth]{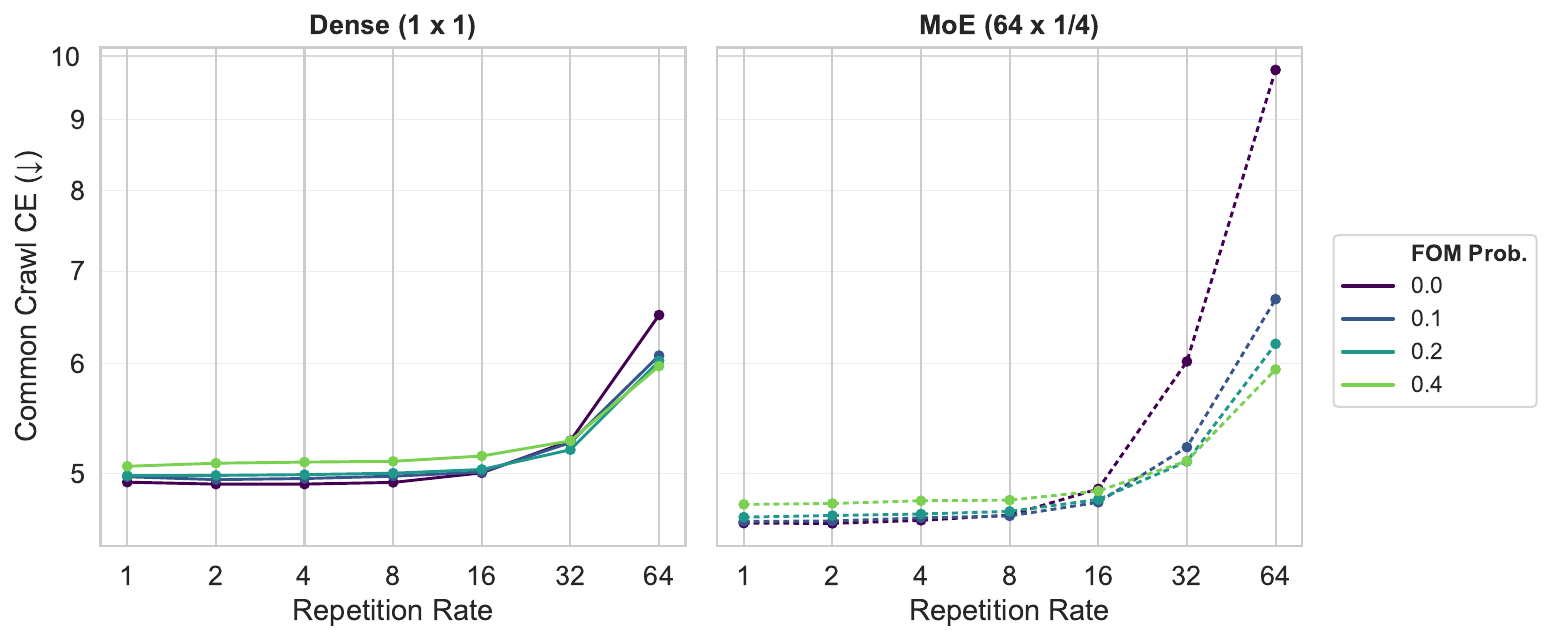}
            \subcaption{Final Output Masking}
        \label{fig:regularizers:fom}
        \end{subfigure}
    \caption{
    \textbf{Dropout (a), FFN Output Masking (b), Expert Dropout, and Expert Output Masking each reduces overfitting from data repetition (\S\ref{sec:reg}).}
    Of the regularization methods studied in \S\ref{sec:reg}, these four dramatically decrease the response to data repetition. However, weight decay and gradient norm clipping, as well as MoE router jitter, have minimal effect. See additional figures in Appendix~\ref{app:reg}.
    }
    \label{fig:regularizers}
\end{figure*}

\else
    \begin{figure*}[!t]
    \centering
    \begin{subfigure}[t]{0.49\textwidth}
        \begin{subfigure}[t]{\textwidth}
            \centering
            \includegraphics[width=\linewidth]{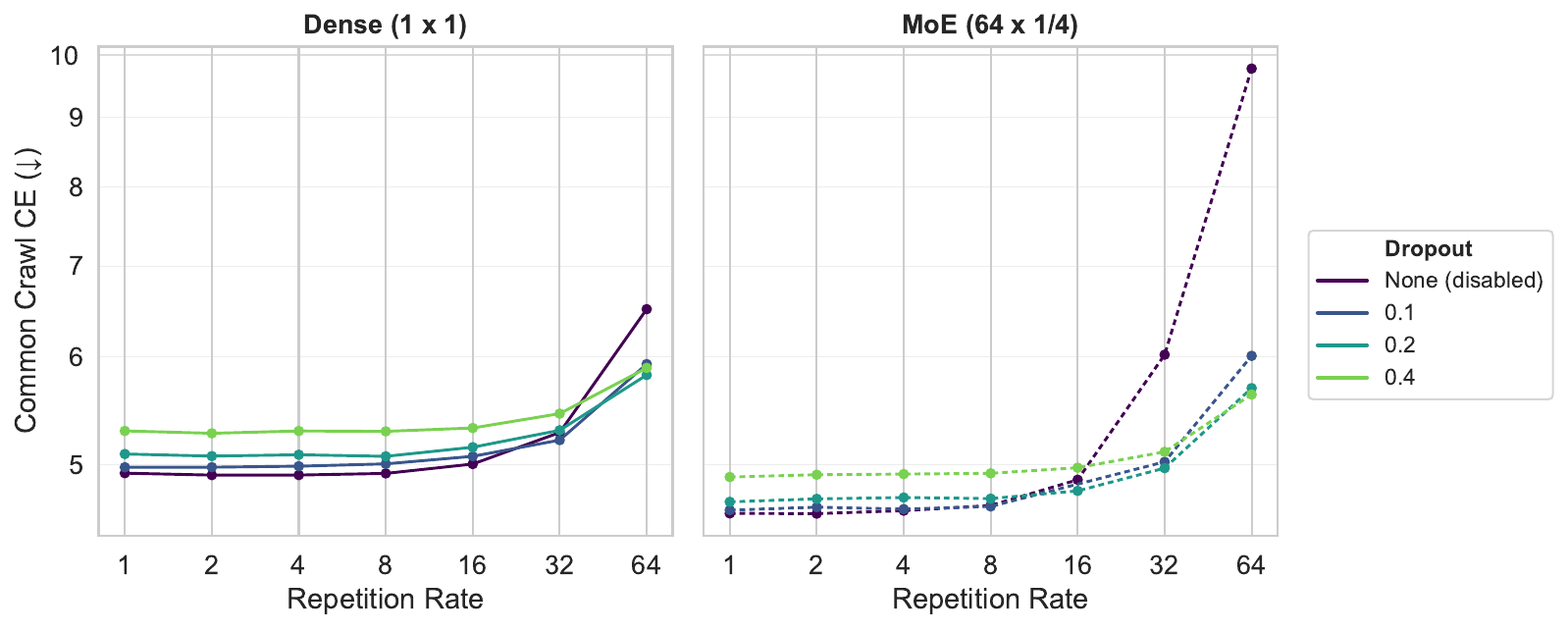}
        \end{subfigure}
        \subcaption{Dropout}
        \label{fig:regularizers:dropout}
    \end{subfigure}
        \begin{subfigure}[t]{0.49\textwidth}
            \centering
            \includegraphics[width=\linewidth]{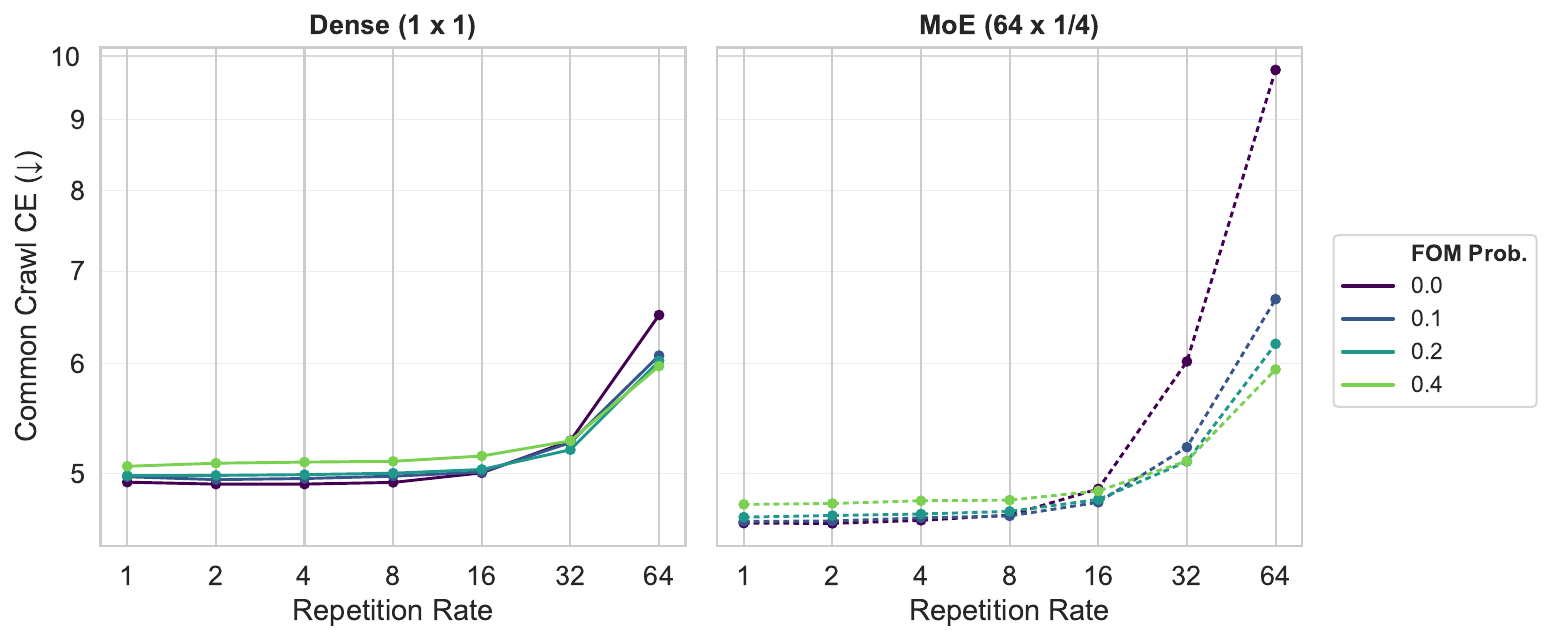}
            \subcaption{Final Output Masking}
        \label{fig:regularizers:fom}
        \end{subfigure}
    \caption{
    \textbf{Dropout (a), FFN Output Masking (b), Expert Dropout, and Expert Output Masking each reduces overfitting from data repetition (\S\ref{sec:reg}).}
    Of the regularization methods studied in \S\ref{sec:reg}, these four dramatically decrease the response to data repetition. However, weight decay and gradient norm clipping, as well as MoE router jitter, have minimal effect. See additional figures in Appendix~\ref{app:reg}.
    }
    \label{fig:regularizers}
\end{figure*}

\fi

We now apply known regularization methods with two motivations: to further probe the mechanisms responsible for performance decline under data repetition, and to take initial steps towards solutions.

\paragraph{Dropout} We use element-wise \emph{residual dropout} \citep{srivastava2014dropout} on the output of each sub-layer with probability $p\in\{0.0, 0.1, 0.2, 0.4\}$. High dropout probability hurts performance at low data repetition, but dramatically reduces the impact of extreme data repetition for all architectures (Figure~\ref{fig:regularizers:dropout}).

\paragraph{Gradient Norm Clipping} The gradient clipping threshold upper bounds the magnitude of each batch update, and is often used to decrease instability \citep{pascanu2013difficulty}. We sweep this threshold over $\{0.2, 1.0, 2.0, \textrm{None}\}$, where \emph{None} indicates no clipping.  All four settings result in similar performance, within variance (Appendix Figure~\ref{fig:regularizers_app}).

\paragraph{Weight Decay} We sweep decoupled AdamW weight decay over $\lambda \in \{0.05, 0.1, 0.2, 0.4\}$. Performance differences, though well-ordered, are small and remain within noise thresholds (Appendix Figure~\ref{fig:regularizers_app}).

\paragraph{FFN Output Masking} FFN output masking (FOM) zeroes, with probability $p$, the entire dense FFN or MoE output for a token without rescaling. FOM is similar to a coarse-grained dropout applied only to FFNs. We consider $p\in\{0.0, 0.1, 0.2, 0.4\}$ and observe that, at low repetition rate, it incurs a smaller penalty to performance than residual dropout. At high repetition rate, it has a similar regularizing effect to dropout (Figure~\ref{fig:regularizers:fom}).

\paragraph{Expert Dropout}
Expert dropout (\citet{switch}; \citet{stmoe}) applies dropout only to the expert hidden activation. We consider expert dropout rate in \{0.0, 0.1, 0.2, 0.4\}, and compare to the dense analogue, which applies dropout to the FFN. Expert Dropout, like FOM, acts only on the FFN, and indeed behaves similarly to FOM (Figure~\ref{fig:regularizers_app}).

\paragraph{Expert Output Masking} Expert output masking (EOM) independently zeroes each (token, selected expert) output with probability $p$ during training. No rescaling is applied. EOM thus operates only on the FFNs, like FOM and Expert Dropout, with an intermediate granularity. We consider $p\in\{0.0, 0.1, 0.2, 0.4\}$ and observe that EOM has a similar effect to both FOM and Expert Dropout (Figure~\ref{fig:regularizers_app}).

\paragraph{Router Jitter}
Router jitter perturbs routing, multiplying the router's input by elementwise uniform noise on
$[1-\epsilon, 1+\epsilon]$ during training.
Over $\epsilon \in \{0.0, 0.1, 0.2, 0.4\}$, we observe no clear impact on performance (Appendix Figure~\ref{fig:regularizers_app}).

\section{Mechanistic Investigation of Data Repetition}
\label{sec:analysis}

In \S\ref{sec:expts}, 
we hypothesize that MoEs may suffer more from data repetition if each expert's routed token set is fixed early in training, thus exposing each expert FFN to a significantly reduced set of unique tokens, when compared to dense FFNs. We consider two aspects of this hypothesis: firstly, the stability of the router, which would result in a fixed data partition over experts; secondly, the resulting specialization of each expert.

\subsection{MoE Router Ossification}
\label{sec:analysis_router_oss}

We study routing patterns for each saved checkpoint: we run inference on a fixed batch of 16{,}384 tokens from the Dolma Common Crawl validation split, and record each token's top-1 expert at every MoE layer. We define the \emph{routing stability} at checkpoint $i$ as the fraction of tokens whose top-1 expert did not change from checkpoint $i-1$, averaged over layers. Separately, we consider expert co-activation and router load balance. Further details are in Appendix~\ref{app:routing_details}.

\begin{figure*}[!t]
    \centering
    \begin{subfigure}[t]{\textwidth}
        \begin{subfigure}[t]{0.55\textwidth}
            \centering
            \includegraphics[width=\linewidth]{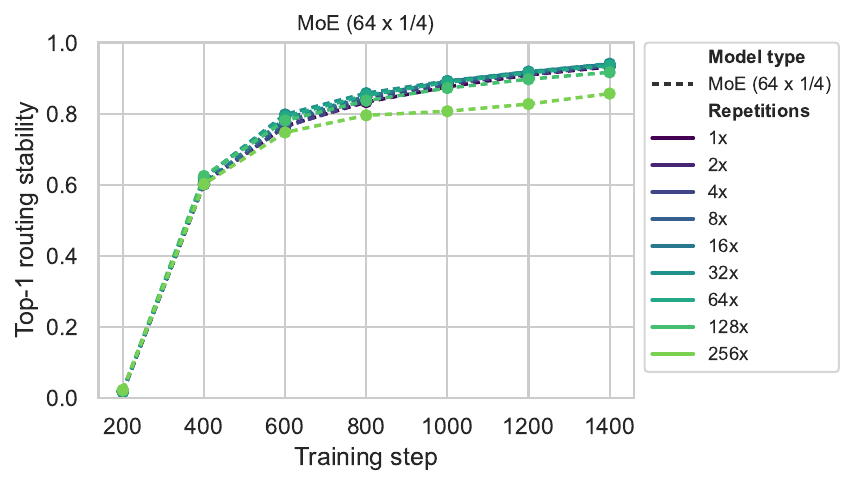}
            \subcaption{Top-1 routing stability over training (80M)}
        \end{subfigure}
        \hfill
        \begin{subfigure}[t]{0.44\textwidth}
            \centering
            \includegraphics[width=\linewidth]{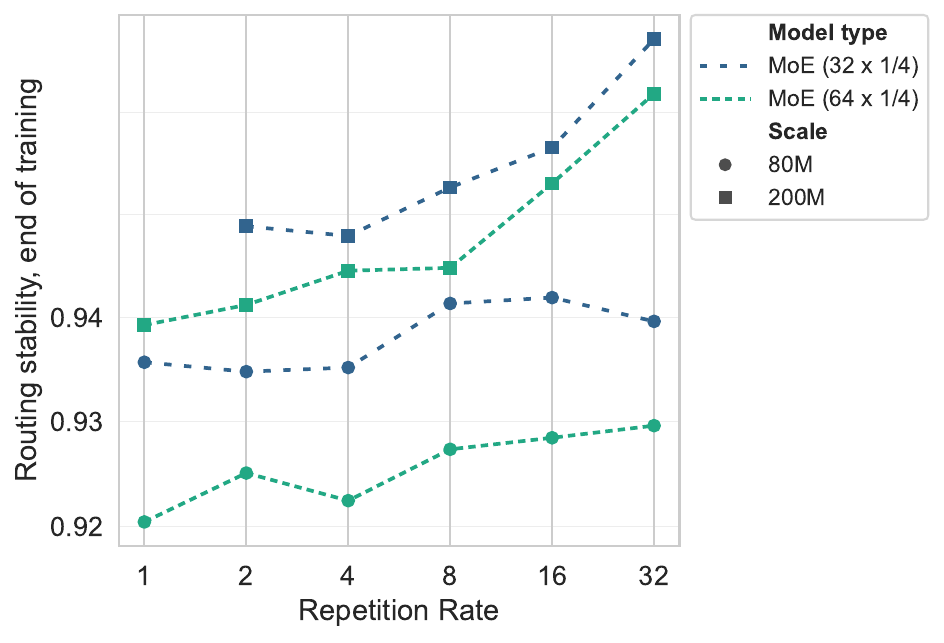}
            \subcaption{Final top-1 routing stability (80M, 200M)}
        \end{subfigure}
    \end{subfigure}
    \caption{\textbf{Routing ossifies early in training, exacerbated by repetition (\S\ref{sec:analysis_router_oss}).}
    We show Top-1 routing stability, defined as the fraction of held-out Common Crawl tokens that keep the same top-1 expert between consecutive checkpoints (200 steps apart), averaged over layers. Left: For the 80M MoE (64 x 1/4) models, consecutive checkpoint agreement is near chance ($1/64$) at the start of training, but over $0.9$ for the second half of training. Stability increases with repetition, up to $R=32$, where the router appears to destabilize.
    Right: We show the Top-1 routing stability at the end of training for MoE (32 x 1/4) and (64 x 1/4) at 80M and 200M scale. Stability rises more aggressively with $R$ at 200M active parameters. Also see Appendix Figure~\ref{fig:router_oss_app}-\ref{fig:routing_extra_200m_dropout}. }
    \label{fig:router_ossification}
\end{figure*}

\paragraph{Routing ossifies early.} We compute routing stability for MoE models in \S\ref{sec:expts}. Figure~\ref{fig:router_ossification}a shows results for 80M models. 
Routing is unstable (random chance, $0.02 \approx 1/64$) at the first checkpoint (step 200), as models begin from random initialization. By the next checkpoint (step 400, 10\% of training), routing stability rises to 60\%, and continues to rise rapidly to $>95\%$ at the end of training. 
Each MoE configuration at each scale exhibits similar patterns (Appendix~\ref{app:routing_extra}).

\paragraph{Higher data repetition exacerbates router ossification.} In Figure~\ref{fig:router_ossification}a, router stability rises uniformly across all data repetition rates until step 600, at which point higher repetition models begin to show consistently higher routing stability. In Figure~\ref{fig:router_ossification}b, end-of-training router stability rises with data repetition, across model scales and MoE sparsities (Appendix~\ref{app:routing_extra}). This supports our hypothesis that each expert's token set is nearly stationary for most of training, so under repetition an expert sees the same reduced shard of data over and over.

\paragraph{Dropout's regularizing effect does not operate through router plasticity.} We train 200M models using the dropout settings of \S\ref{sec:reg} (with checkpoints 1,000 steps apart, not directly comparable to the above results). Late-training router stability again rises slowly but consistently with repetition (Appendix Figure~\ref{fig:routing_extra_200m_dropout}).
At $R=64$, models with dropout show less routing ossification than the no-dropout setting,
even though they overfit far less (\S\ref{sec:reg}). Since dropout partially recovers performance without affecting router ossification, we hypothesize that overfitting is not primarily driven by routing, but rather the functions learned by each expert.

\paragraph{Higher repetition encourages uniformly distributed expert co-activation.} Although the top-1 expert contributes the majority of the output weight, we also conduct a simple investigation of expert co-activation. For each unordered pair of experts within each layer, we count
how often both experts in the pair are active for the same token. We normalize the counts and compute the entropy of the distribution. Under repetition this distribution trends toward uniform in every configuration (Appendix~\ref{app:routing_extra}).

\paragraph{Router output magnitude is higher with data repetition; load balancing shows no clear correlation.} Router load imbalance, or the ratio between the maximum and mean expert token loads in each batch, does not predictably shift with repetition rate, except that extremely high $R$ yields outliers 1B scale. Load balancing loss also does not change predictably with repetition. Z-loss, which measures router logit magnitude, is higher in early training with higher $R$, which is consistent with earlier and more extreme router ossification (Appendix Figure~\ref{fig:routing_imbalance_app}-\ref{fig:routing_zloss_app}).

\subsection{MoE Expert Specialization}
\label{sec:analysis_exp_spec}
Early routing ossification does not necessitate expert specialization; experts with disjoint routed token sets may still learn similar functions. We investigate the specialization of experts by measuring the 
\emph{expert knockout cost}, or the loss impact of removing an expert.
Specifically, we perform inference with the final checkpoint of each model and measure, for each expert at each layer, the total increase in CE Loss resulting from masking only that expert's outputs.

\begin{figure*}[!t]
    \centering
    \begin{subfigure}[t]{\textwidth}
        \begin{subfigure}[t]{0.49\textwidth}
            \centering
            \includegraphics[width=\linewidth]{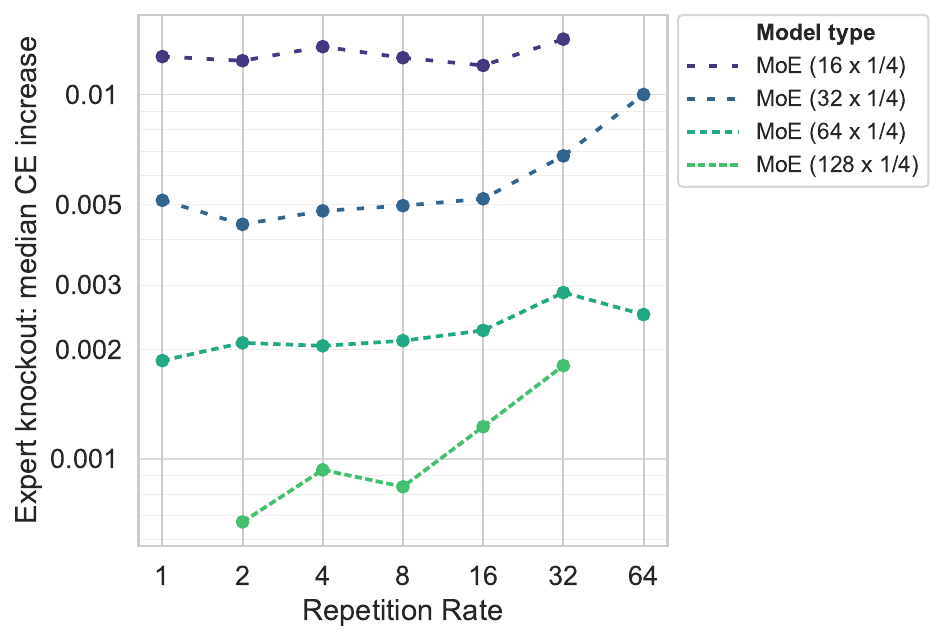}
        \end{subfigure}
        \hfill
        \begin{subfigure}[t]{0.49\textwidth}
            \centering
            \includegraphics[width=\linewidth]{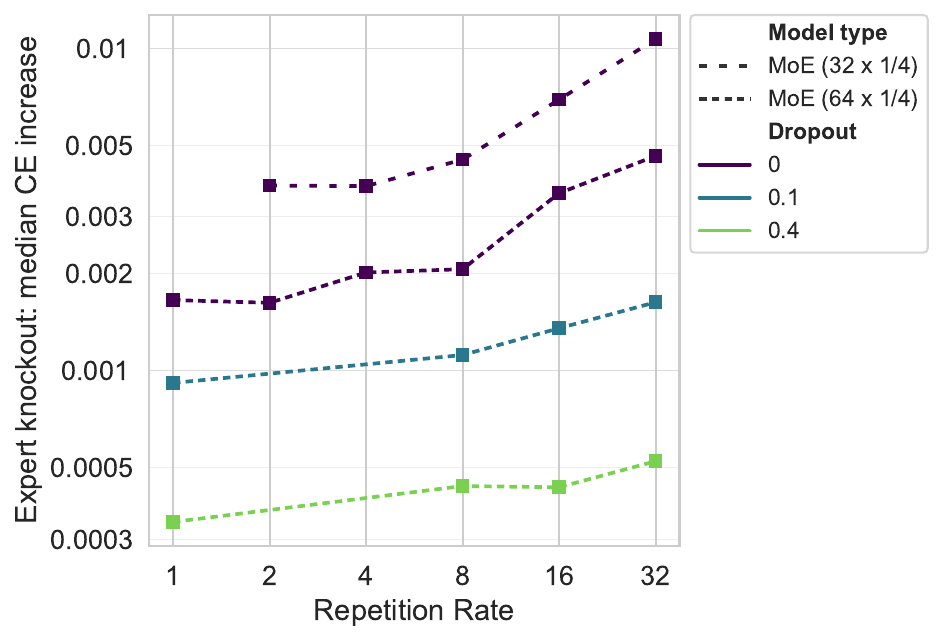}
        \end{subfigure}
    \end{subfigure}
    \caption{\textbf{Repetition increases expert specialization; dropout reduces this effect
    (\S\ref{sec:analysis_exp_spec}).}
    We compute the increase in held-out CE when one expert's output is zeroed at inference time, and show the median over all experts and MoE layers of the final checkpoint.
    Left (80M active parameters): Expert specialization grows with repetition rate, and is overall higher when there are fewer total experts, but is unaffected by expert granularity (Appendix Figure~\ref{fig:routing_extra_80m})
    Right (200M active parameters): Dropout reduces expert specialization across all $R$, and also reduces the relative magnitude of the increase in expert specialization that results from higher $R$. }
    \label{fig:expert_knockout}
\end{figure*}

\paragraph{Expert specialization rises with data repetition.} Knockout cost at $R=1$ is higher at low total expert count. As $R$ increases, knockout cost increases for all configurations, but does so more rapidly at higher expert counts. At 80M (Figure~\ref{fig:expert_knockout}a), increasing $R$ from 1 to 32 also increases expert knockout effect by 1.1$\times$ for 16 experts, and by 2.3$\times$ for 128 experts of 1/4 granularity. This pattern echos \S\ref{sec:expts}, where CE degradation under repetition also grows with expert count. Thus, repetition encourages expert specialization and reduces redundancy, especially at higher expert counts.

\paragraph{Dropout decreases expert specialization.} We compare various dropout settings at 200M (Figure~\ref{fig:expert_knockout}b). Dropout consistently reduces the median knockout cost for all values of $R$. We hypothesize that dropout combats overfitting by removing dependence on any single expert, and enforcing multiple, more varied, representations of features.

\section{Related Works}

\citet{hernandez2022repeated} show that repeating a small fraction of the training data degrades held-out loss non-monotonically. \citet{muennighoff2023scaling} studies dense models trained on C4 and finds that repetition up to roughly four epochs is comparable to all-unique data, but further repetition decays performance. More recent work on dense models has provided evidence that data repetition damage: (1) grows with model scale \citep{kazdan2026scaledependent}; 
(2) peaks at an intermediate repeat
count \citep{chudnovsky2026internal}; 
(3) is well described by a single
additive coefficient that strong weight decay can shrink \citep{lovelace2026prescriptive}; 
and (4) can be modulated by data quality and mixture weights \citep{fang2025unequal,liu2026infolaw,chen2025revisiting}. 
Very little work has considered MoEs; \citet{xue2023repeat} concludes from a single MoE configuration, a 16-expert T5, that parameter count drives multi-epoch degradation while FLOPs are close to irrelevant.

Other work has modeled the tradeoff between quality and quantity of unique data when training dense Transformers.
\citet{fang2025unequal} reports that repeating a heavily filtered set up to ten
times can beat a single pass over a superset ten times larger, while
\citet{mohri2026bitter} argues that with enough compute, larger quantities of unfiltered data is better. 
\citet{liu2026infolaw} and
\citet{chen2025revisiting} fit quality-weighted mixtures and repetition together.

Even fewer studies have considered interventions for minimizing data-repetition effects. \citet{xue2023repeat} uses dropout to reduce multi-epoch degradation, but caution that the dropout probability needs retuning as models grow. \citet{lovelace2026prescriptive}
shows that raising weight decay by an order of magnitude cuts their fitted overfitting coefficient by roughly $70\%$ at high repetition rates, at the price of a loss premium in the single-epoch regime.

\section{Conclusion}

Across all dense and MoE architectures, we find that increasing data repetition rate $R$ leads to overfitting. Sparse MoE models deteriorate earlier and more rapidly as $R$ increases. We show that the response to data repetition primarily depends on total parameters. The pattern of overfitting effects are remarkably robust to a variety of single data domains, data mixes, and different levels of data filtering. Mixing a repeated data domain into a larger or equal-sized non-repeated data domain may have a regularizing effect. 

We successfully reduce the overfitting response to data repetition through regularization methods that operate by dropping parameter outputs (dropout, expert dropout, FFN output masking, and expert output masking). However, no method fully matches performance achieved with all-unique data. Gradient Norm Clipping, Weight Decay, and Router Jitter operate through qualitatively different mechanisms to reduce update strength, reduce weight magnitude, and modify coarse-grained gradient paths, respectively, and do not have any measurable effect.

Finally, our mechanistic analyses provide evidence that MoE routing is fixed early in training, and only slightly exacerbated by data repetition. Expert specialization also rises with data repetition. Dropout does not affect router ossification, but decreases expert specialization.

In summary, our work studies the interaction between data constraints and the design of MoE architectures. We present strong evidence that data repetition causes overfitting, that the degree of performance degradation primarily depends on total parameters, and that the underlying mechanism operates partially through overly specialized parameters, broken by methods such as dropout and output masking. We recommend that future works further explore masking-based methods to minimize repetition-driven overfitting through decreased parameter specialization.

\section*{Acknowledgments}
We are grateful to Rohan Sanda for initial engineering support; to Ananya Harsh Jha and Jacqueline He for helpful discussion; and to the contributors and maintainers of the UW Hyak and Stanford Marlowe computing resources.

\bibliography{iclr2027_conference}
\bibliographystyle{iclr2027_conference}

\clearpage

\appendix
\section{Experimental Details}

\subsection{Model Architecture}
\label{app:model_arch}

\begin{table}[!ht]
    \centering
    \scriptsize
\begin{tabular}{lcccccccccccc}
\toprule
\multirow{3}{*}{\textbf{Scale}} & \multirow{3}{*}{\textbf{Layers}} &  \multirow{3}{*}{\shortstack[c]{\textbf{Model} \\ \textbf{Dim}}} & \multirow{3}{*}{\shortstack[c]{\textbf{Attention}\\ \textbf{Heads}}} & \multirow{3}{*}{\textbf{Name}} & \textbf{Activation} & \textbf{Total} & \textbf{Active} & \textbf{Expert} & \textbf{Total} & \textbf{Active} & \textbf{Total} \\ 
& & & & & \textbf{Sparsity} & \textbf{Experts} & \textbf{Experts} & \textbf{Gran.} & \textbf{Tokens} & \textbf{Param} & \textbf{Param} \\
& & & & & \textbf{(s)} & \textbf{(n)} & \textbf{(k)} & \textbf{(g)} & \textbf{($T$)} & \textbf{($N_{a}$)} & \textbf{($N$)} \\
\midrule
80M	&	8	&	336	&	7	&	dense	&	1	&		&		&		&	1.6B	&	81.8M	&	81.8M	\\
	&		&		&		&	MoE (8 x 1/4)	&	2	&	8	&	4	&	1/4	&		&		&	92.7M	\\
	&		&		&		&	MoE (64 x 1/32)	&	2	&	64	&	32	&	1/32	&		&		&	92.7M	\\
	&		&		&		&	MoE (16 x 1/4)	&	4	&	16	&	4	&	1/4	&		&		&	114.4M	\\
	&		&		&		&	MoE (64 x 1/16)	&	4	&	64	&	16	&	1/16	&		&		&	114.4M	\\
	&		&		&		&	MoE (32 x 1/4)	&	8	&	32	&	4	&	1/4	&		&		&	157.7M	\\
	&		&		&		&	MoE (64 x 1/8)	&	8	&	64	&	8	&	1/8	&		&		&	157.7M	\\
	&		&		&		&	MoE (64 x 1/4)	&	16	&	64	&	4	&	1/4	&		&		&	244.4M	\\
	&		&		&		&	MoE (64 x 1/2)	&	32	&	64	&	2	&	1/2	&		&		&	417.8M	\\
	&		&		&		&	MoE (128 x 1/4)	&	32	&	128	&	4	&	1/4	&		&		&	417.8M	\\
    &		&		&		&	MoE (256 x 1/4)	&	64	&	256	&	4	&	1/4	&		&		&	764.6M	\\
\midrule
200M	&	10	&	640	&	10	&	dense	&	1	&		&		&		&	4B	&	193.9M	&	193.9M	\\
	&		&		&		&	MoE (32 x 1/4)	&	8	&	32	&	4	&	1/4	&		&		&	538.0M	\\
	&		&		&		&	MoE (64 x 1/4)	&	16	&	64	&	4	&	1/4	&		&		&	931.2M	\\
\midrule
1B	&	15	&	1664	&	16	&	dense	&	1	&		&		&		&	20B	&	998.3M	&	998.3M	\\
	&		&		&		&	MoE (64 x 1/4)	&	16	&	64	&	4	&	1/4	&	&	&		8.5B	\\
\bottomrule
\end{tabular}
    \caption{
    \textbf{Architecture Details and Parameter Counts }
    }
\label{tab:param_counts} 
\end{table}

\subsection{Hyperparameters}
\label{app:hps}

\begin{table}[!h]
    \centering
\begin{tabular}{lc}
\toprule
 Hyperparameter & Value \\ \midrule
Vocabulary size & 50K \\
Batch Size &    512	\\
Sequence Length & 2048	\\
Learning Rate & 4e-4  \\
Encoder-Decoder Weight Sharing & No \\
Feedforward Dimension & 4 x hidden dimension \\
LR Schedule & Cosine Decay \\
LR Warmup & 2000 steps \\
End LR & 0.1 x Peak LR \\
Weight Decay & \{0.0, \textbf{0.1}, 0.2, 0.4 \} \\
Max Grad Norm & \{None, 0.2, \textbf{1}, 2.0 \} \\
Dropout & \{\textbf{0.0}, 0.1, 0.2, 0.4\} \\
Nonlinearity & SwiGLU \\
MoE Z-loss & 1e-3 \\
Load Balancing Loss Weight & 1e-2 \\
MoE Token Dropping & Dropless \\
MoE Routing Choice & Token Choice \\
\bottomrule
\end{tabular}
    \caption{
    \textbf{Hyperparameter details for models in \S\ref{sec:expts}-\ref{sec:analysis}.} Multiple values indicate that we investigated different settings in \S\ref{sec:reg}, and bold values are defaults used in \S\ref{sec:expts}.}
\label{tab:hparams} 
\end{table}

\subsection{Training Data Sources}
\label{app:train_data_sources}
We take our training data from \citet{olmoe}. We use their data mix, which we call \emph{OLMoE Mix}, consisting of documents from: 
DCLM-Baseline~\citep{li2024datacomplm}, StarCoder~\citep{li2023starcoder,kocetkov2022stack3tbpermissively}, peS2o~\citep{peS2o,soldaini2024dolma}, arXiv~\citep{together2023redpajama}, OpenWebMath~\citep{paster2023openwebmath}, Algebraic Stack~\citep{azerbayev2023llemma}, English Wikipedia \& Wikibooks~\citep{soldaini2024dolma}. 

In \S\ref{sec:expts_filtering}, we also use \textsc{DCLM-Pool} \citep{li2024datacomplm}.

\subsection{Evaluation Data}
\label{app:eval_data_sources}

Our evaluation includes held-out validation sets for language modeling, as well as downstream tasks. 

The language modeling tasks are a subset of Paloma
\citep{magnusson2024palomabenchmarkevaluatinglanguage}, which consists of:
\textsc{C4} (\citet{raffel2019exploringtl} via \citet{dodge-etal-2021-documenting}), \textsc{The Pile} \citep{Gao2020ThePA}, \textsc{WikiText-103} \citep{Merity2016PointerSM}, \textsc{Dolma} \citep{dolma}, \textsc{M2D2 S2ORC} \citep{reid-etal-2022-m2d2}, \textsc{ICE} (\citet{GREENBAUM_1996} via \citet{Liang2022HolisticEO}). \textsc{Dolma} is subdivided into six domains: books, common-crawl, pes2o, reddit\_uniform, stack\_uniform, wiki.

In \S\ref{sec:expts_single_domain}, we vary the dataset used for validation loss to match the training domain. For models trained on DCLM, we evaluate on Dolma common-crawl; for peS2o, Dolma pes2o; for Wikipedia, Dolma wiki; for StarCoder, Dolma stack\_uniform.

The downstream tasks consist of BoolQ \citep{clark2019boolq}, HellaSwag \citep{zellers2019hellaswag}, and MMLU \citep{son2024kmmlumeasuringmassivemultitask}. MMLU is subdivided into four domains: humanities, STEM, social sciences, and other.

\subsection{Data Repetition}
\label{app:data_rep}
In data-constrained regimes, the set of $U$ unique tokens used for each experiment is constructed as follows: we fix a random permutation of the sequences in each data domain $\mathcal{D}$, and take the first $U$ tokens for training. We repeat these $U$ tokens for $R$ epochs, shuffling between epochs.

In \S\ref{sec:expts_filtering}, we mix tokens from \textsc{DCLM-Baseline} and \textsc{DCLM-Pool} \citep{li2024datacomplm}. This inevitably introduces a very small amount of unmeasured data repetition because our selected subsets of \textsc{DCLM-Baseline} and \textsc{DCLM-Pool} may have a non-empty intersection. However, this intersection is likely to be of negligible size. \textsc{DCLM-Baseline} consists of 5T tokens, of which we use 1.6B at most. \textsc{DCLM-Pool} consists of 240 trillion tokens, of which we use 1.6B at most. The probability that any particular token in a particular sequence from our \textsc{DCLM-Baseline} subset also appears in our \textsc{DCLM-Pool} is less than 2e-9, yielding an expected total of fewer than 3 repeated tokens. In other words, we expect effectively no repeated tokens on average.

\subsection{Routing Analysis}
\label{app:routing_details}

In \S\ref{sec:analysis}, we use a batch size of 16{,}384 tokens (8 sequences of 2{,}048 tokens). We use a fixed subset of the Dolma Common Crawl validation split. Dropout, router jitter, and other train-time regularizers are inactive.

\paragraph{Ossification (\S\ref{sec:analysis_router_oss}).} For each checkpoint we record the top-1 expert of every token at every MoE layer, defined as the expert with the largest router score. For each pair of consecutive checkpoints we report the fraction of tokens whose top-1 expert is identical, averaged over layers. Checkpoints are 200 steps apart in the core ladders and 1{,}000 steps apart in the 200M dropout arms, so stability values are only compared between runs with matching spacing. End-of-training stability is the mean over the last two regularly spaced intervals.

\paragraph{Expert knockout (\S\ref{sec:analysis_exp_spec}).} On the final checkpoint we zero all MLP weights of one expert, so its output is exactly zero for the tokens routed to it. The router is untouched and the weights of the remaining selected experts are not renormalized. We then recompute CE on the token batch, repeat for every expert in every MoE layer, and report the median and maximum increase over the CE of the unmodified model.

\paragraph{Co-activation (\S\ref{sec:analysis_exp_spec}).} On the final checkpoint we count, per layer, how often each unordered pair of experts appears together in a token's top-$k$ set, normalize the counts to a distribution, and compute its Shannon entropy. We divide by $\log\big(n(n-1)\big)$, its maximum for $n$ experts, so values are comparable across expert counts.

\clearpage
\section{Additional Results}

\ifarxiv 
\else
    \subsection{Additional Main Paper Figures}
    \begin{figure*}[!ht]
    \centering
    \begin{subfigure}[t]{\textwidth}
        \begin{subfigure}[t]{\textwidth}
            \centering
            \includegraphics[width=\linewidth]{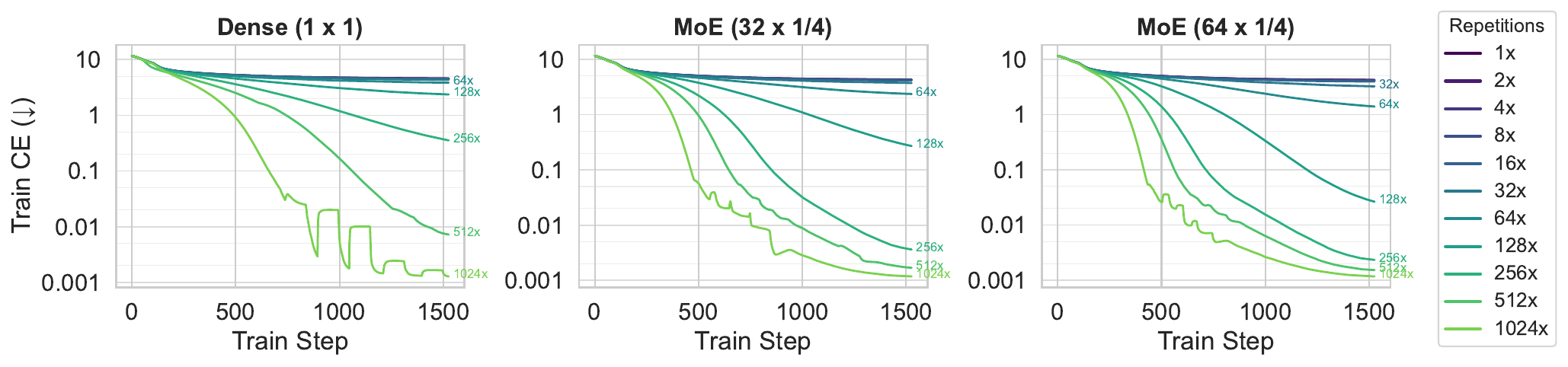}
        \end{subfigure}
        \par\vspace{1em}
        \begin{subfigure}[t]{\textwidth}
            \centering
            \includegraphics[width=\linewidth]{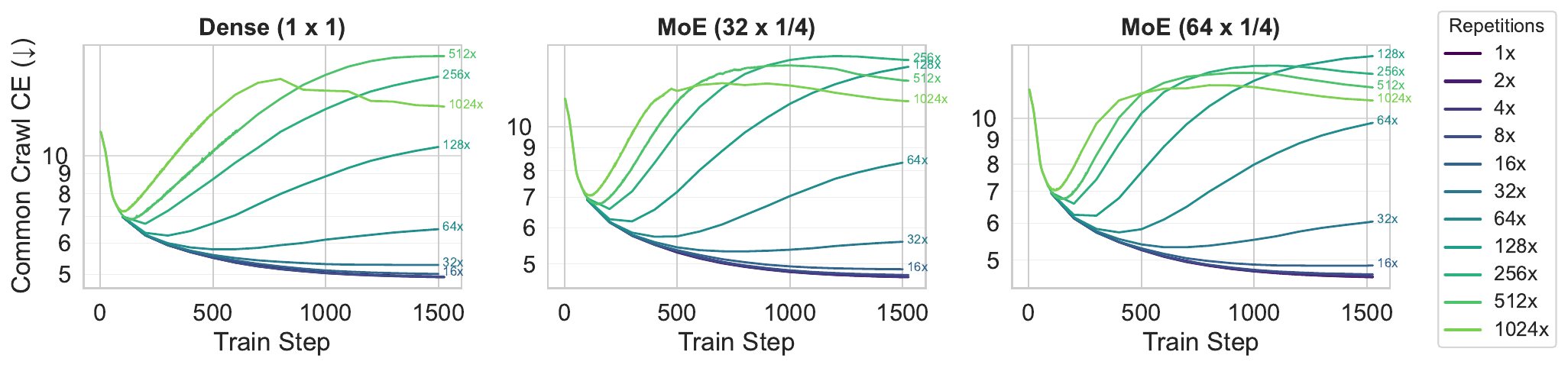}
        \end{subfigure}
    \end{subfigure}

    \caption{\textbf{At higher repetition rates, training loss falls to 0 as models overfit to the repeated data (\S\ref{sec:expts_olmo_mix}).}
    Early in training, train (above) and validation (below) 
    loss fall together, but, at high repetition rate $R$, train loss falls rapidly to 0, indicating memorization of training data, while validation loss rises. Additional model scales in Appendix Figure~\ref{fig:train_valid_app}.}
    \label{fig:train_v_valid}
\end{figure*}

    \clearpage
\fi

\subsection{Random Seed Variance}
\label{app:random_seed}
In Table~\ref{tab:random_seed}, we report the variance across 5 random seeds for Dense and MoE (64 x 1/4) models trained on the OLMoE mix at $R=1, R=32$, and evaluated on all language modeling validation datasets and downstream tasks.

\begin{table}[!ht]
    \centering
    \footnotesize
\begin{tabular}{lcc|cc|cc|cc}
\toprule
\textbf{Metric}	&	\multicolumn{4}{c}{\textbf{Dense}}	&			\multicolumn{4}{c}{\textbf{MoE (64 x 1/4)}}			\\
	&	\multicolumn{2}{c}{\textbf{R=1}}	&	\multicolumn{2}{c}{\textbf{R=32}}	&	\multicolumn{2}{c}{\textbf{R=1}}	&	\multicolumn{2}{c}{\textbf{R=32}}	\\
	&	Mean & Std. Dev.	&	Mean & Std. Dev.	&	Mean & Std. Dev.	&	Mean & Std. Dev.	\\
\midrule									
\textbf{Train Loss}	&	4.57 & 0.00 & 		4.27 & 0.00	&	4.26 & 0.01	&	3.17 & 0.02	\\
\midrule			
\textbf{Validation LM Loss}					\\
C4              &	4.86 & 0.01 & 		5.27 & 0.02	&	4.53 & 0.01	&	6.04 & 0.01	\\
Dolma Books 	&	5.04 & 0.00 & 		5.59 & 0.06	&	4.72 & 0.01	&	6.48 & 0.05	\\
Dolma Common Crawl 	&	4.92 & 0.01 & 		5.29 & 0.02	&	4.61 & 0.01	&	6.05 & 0.02	\\
Dolma peS2o 	&	4.48 & 0.01 & 		4.93 & 0.04	&	4.12 & 0.01	&	5.66 & 0.01	\\
Dolma Reddit 	&	4.74 & 0.00 & 		5.12 & 0.02	&	4.46 & 0.01	&	5.95 & 0.02	\\
Dolma Stack 	&	4.65 & 0.02 & 		6.68 & 0.10	&	4.23 & 0.02	&	7.46 & 0.04	\\
Dolma Wiki 	&	4.67 & 0.01 & 		5.16 & 0.04	&	4.31 & 0.01	&	5.87 & 0.02	\\
ICE 	&	4.97 & 0.01 & 		5.63 & 0.03	&	4.66 & 0.02	&	6.73 & 0.01	\\
M2D2 S2ORC 	&	4.84 & 0.01 & 		5.47 & 0.03	&	4.52 & 0.01	&	6.46 & 0.01	\\
Pile 	&	4.62 & 0.01 & 		5.37 & 0.05	&	4.27 & 0.01	&	6.19 & 0.01	\\
WikiText-103 	&	5.06 & 0.00 & 		5.76 & 0.08	&	4.67 & 0.02	&	6.67 & 0.03	\\
\textbf{Average}	&	4.81 & 0.01 & 		5.48 & 0.05	&	4.46 & 0.01	&	6.32 & 0.02	\\
\midrule									
\textbf{Downstream Task Loss}		\\							
BoolQ 	&	2.52 & 0.22 & 		3.14 & 0.18	&	2.33 & 0.24	&	3.95 & 0.43	\\
HellaSwag 	&	0.96 & 0.00 & 		1.05 & 0.00	&	0.89 & 0.00	&	1.23 & 0.00	\\
MMLU Humanities 	&	2.13 & 0.06 & 		3.48 & 0.11	&	1.91 & 0.13	&	4.36 & 0.92	\\
MMLU Other 	&	1.97 & 0.04 & 		2.85 & 0.17	&	1.80 & 0.06	&	2.93 & 0.20	\\
MMLU Social Sciences 	&	1.98 & 0.04 & 		2.66 & 0.14	&	1.88 & 0.10	&	3.06 & 0.31	\\
MMLU STEM 	&	2.02 & 0.03 & 		2.61 & 0.07	&	1.90 & 0.08	&	3.13 & 0.48	\\
\textbf{Average}	&	1.93 & 0.07 & 		2.63 & 0.11	&	1.78 & 0.10	&	3.11 & 0.39	\\
\midrule									
\textbf{Downstream Task Accuracy}		\\							
BoolQ 	&	0.39 & 0.01 & 		0.39 & 0.01	&	0.41 & 0.05	&	0.45 & 0.03	\\
HellaSwag 	&	0.26 & 0.00 & 		0.25 & 0.00	&	0.26 & 0.00	&	0.25 & 0.00	\\
MMLU Humanities 	&	0.24 & 0.01 & 		0.24 & 0.00	&	0.25 & 0.01	&	0.25 & 0.01	\\
MMLU Other 	&	0.27 & 0.01 & 		0.24 & 0.01	&	0.27 & 0.01	&	0.25 & 0.01	\\
MMLU Social Sciences 	&	0.23 & 0.01 & 		0.22 & 0.00	&	0.24 & 0.01	&	0.23 & 0.01	\\
MMLU STEM 	&	0.27 & 0.00 & 		0.24 & 0.01	&	0.27 & 0.01	&	0.24 & 0.01	\\
\textbf{Average}	&	0.28 & 0.01 & 		0.26 & 0.01	&	0.28 & 0.01	&	0.28 & 0.01	\\

\bottomrule
\end{tabular}
    \caption{
    \textbf{Mean and Standard Deviation across 5 random seeds.} We repeat a selection of the 80M settings from \S\ref{sec:expts_olmo_mix} using 5 random seeds for model initialization, and repeat the mean and standard deviation on each metric. Variance at $R=1$ is near-0 for validation LM datasets. Despite fixing the data used across model initialization seeds, higher data repetition rates yield slightly higher standard deviation. Downstream task loss has higher variance for MoE models, and at higher $R$. Downstream task accuracy has low variance, but remains at near-chance scores.
    }
\label{tab:random_seed} 
\end{table}

\clearpage
\subsection{Extended Settings (\S\ref{sec:expts_olmo_mix})}
\label{app:extended}
\begin{figure*}[!ht]
    \centering
    \begin{subfigure}[t]{\textwidth}
        \begin{subfigure}[t]{0.47\textwidth}
            \centering
            \includegraphics[width=\linewidth]{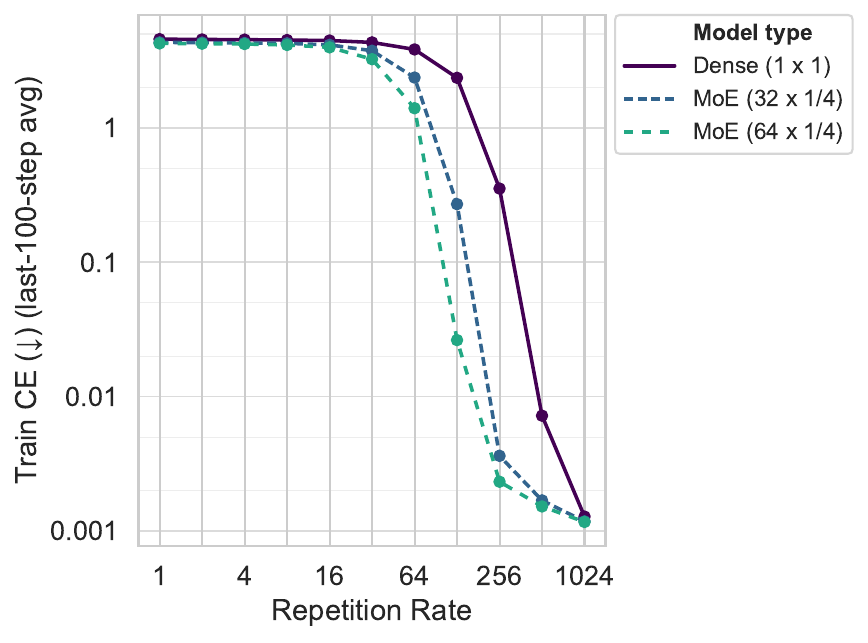}
        \end{subfigure}\hfill
        \begin{subfigure}[t]{0.47\textwidth}
            \centering
            \includegraphics[width=\linewidth]{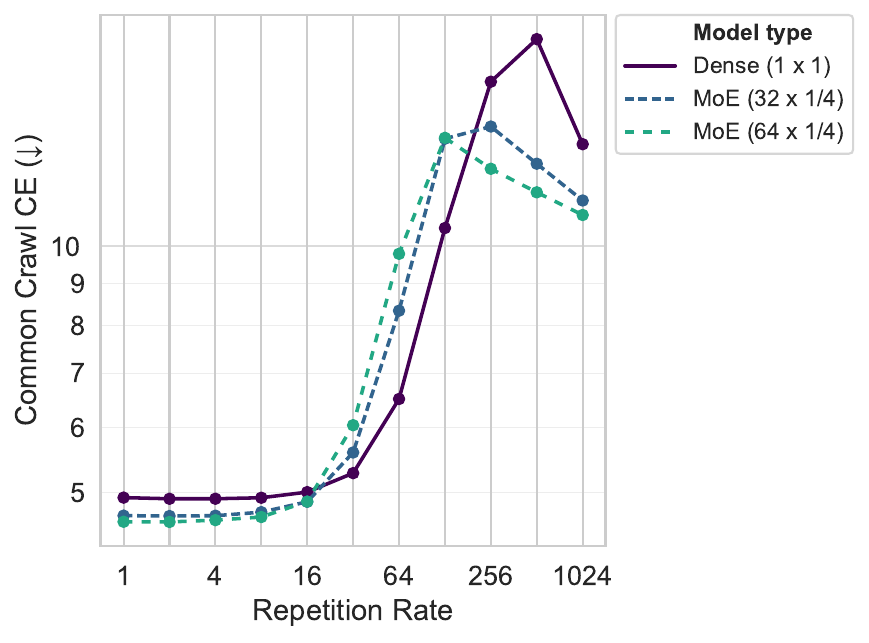}
        \end{subfigure}
        \subcaption{\textbf{80M}}
    \end{subfigure}
        \par\vspace{2em}
    \begin{subfigure}[t]{\textwidth}
        \begin{subfigure}[t]{0.47\textwidth}
            \centering
            \includegraphics[width=\linewidth]{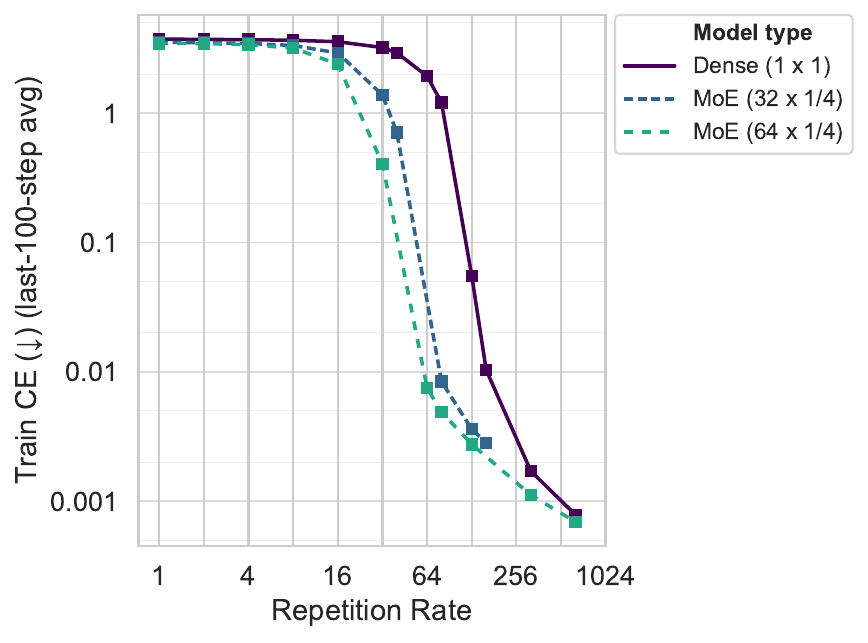}
        \end{subfigure}\hfill
        \begin{subfigure}[t]{0.47\textwidth}
            \centering
            \includegraphics[width=\linewidth]{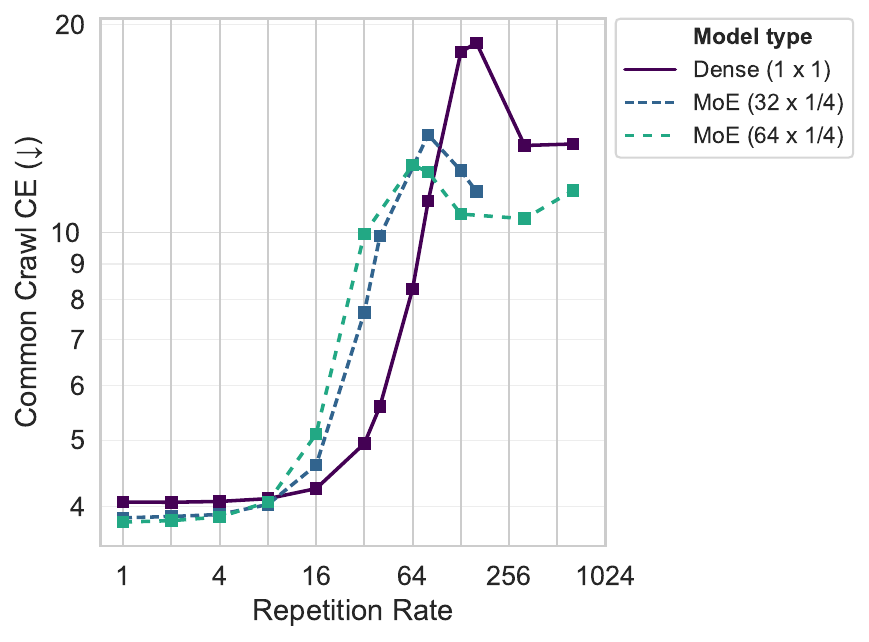}
        \end{subfigure}
        \subcaption{\textbf{200M}}
    \end{subfigure}
        \par\vspace{2em}
    \begin{subfigure}[t]{\textwidth}
        \begin{subfigure}[t]{0.47\textwidth}
            \centering
            \includegraphics[width=\linewidth]{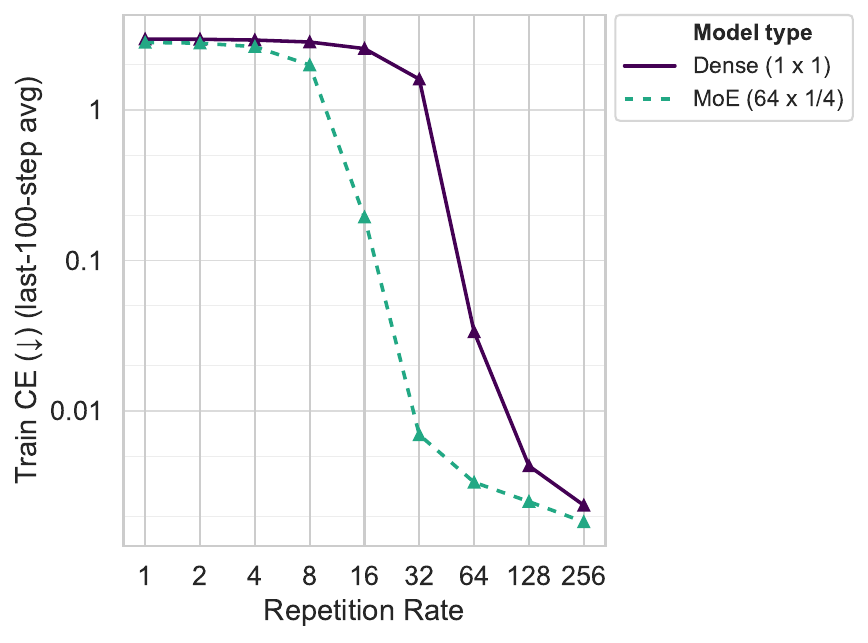}
        \end{subfigure}\hfill
        \begin{subfigure}[t]{0.47\textwidth}
            \centering
            \includegraphics[width=\linewidth]{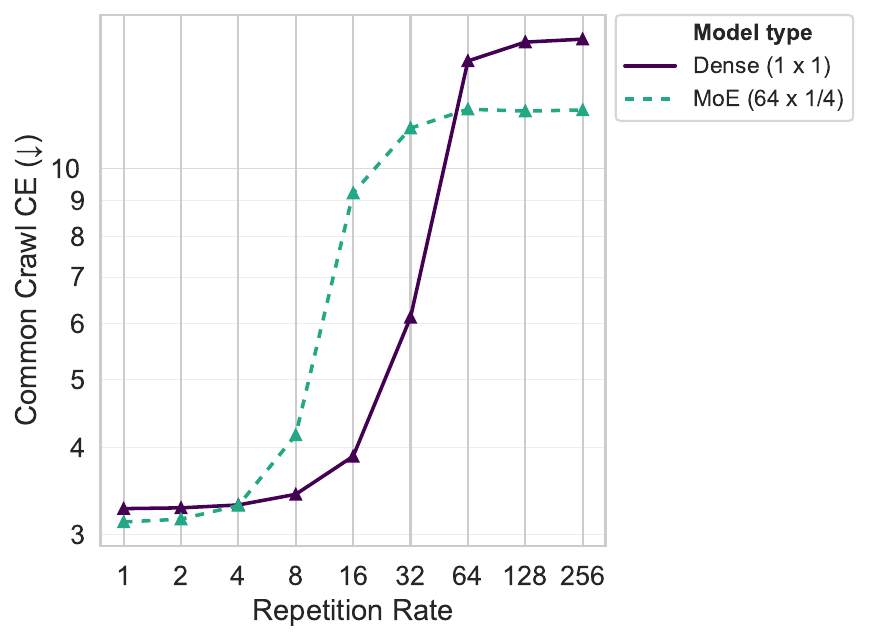}
        \end{subfigure}
        \subcaption{\textbf{1B}}
    \end{subfigure}
    \caption{
    \textbf{Across active parameter scales, data repetition rates over 8 result in increasingly severe overfitting. Sparser models overfit more (\S\ref{sec:expts_olmo_mix}).} At 80M, 200M, and 1B active parameters, we fix the total data budget $T = 20 \cdot N_{a}$, and vary the data repetition rate $R$ via different sized unique token sets. As $R$ increases, models increasingly overfit, as we observe decreasing train loss and rising validation  loss. Sparsity exacerbates overfitting behavior. Larger sparse models overfit more at lower $R$.
    We also consider two additional data repetition rates $R\approx 2^{15}, 2^{20}$ at 80M active parameters, and find that validation loss rises again.
    }
    \label{fig:extended_app}
\end{figure*}

\clearpage
\subsection{Training and Validation Loss Curves (\S\ref{sec:expts_olmo_mix})}
\label{app:training_valid_curves}
\begin{figure*}[!ht]
    \centering
        \begin{subfigure}[t]{\textwidth}
            \centering
            \includegraphics[width=\linewidth]{fig_pdfs/history/olmo_mix__80M__dense_moe32_moe64__train_loss.pdf}
            \subcaption{Train loss over training (80M)}
        \end{subfigure}
        \par\vspace{1em}
        \begin{subfigure}[t]{\textwidth}
            \centering
            \includegraphics[width=\linewidth]{fig_pdfs/history/olmo_mix__80M__dense_moe32_moe64__eval_dolma_cc_loss.pdf}
            \subcaption{Validation loss over training (80M)}
        \end{subfigure}
        \par\vspace{1em}

        \begin{subfigure}[t]{\textwidth}
            \centering
            \includegraphics[width=\linewidth]{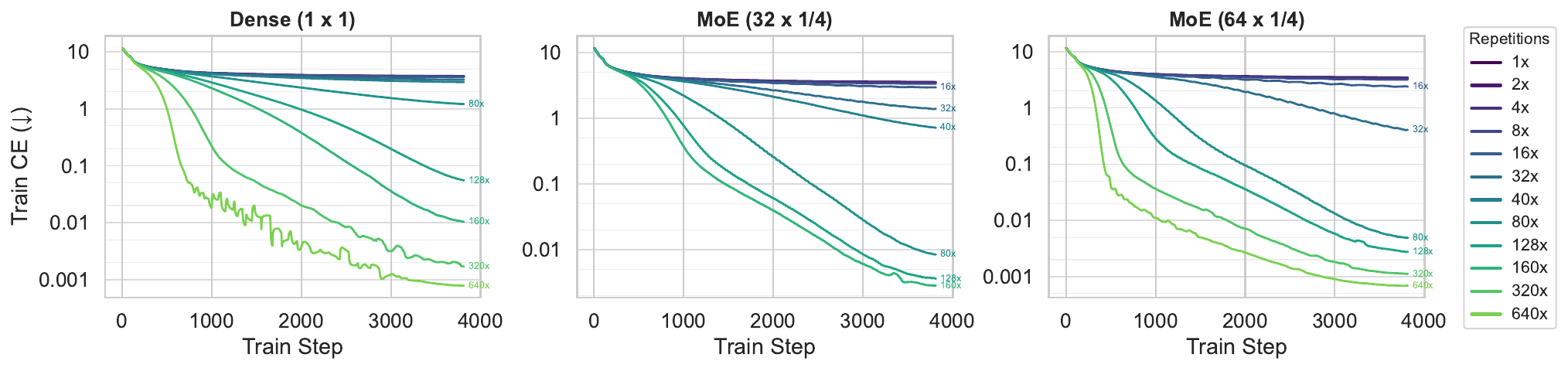}
            \subcaption{Train loss over training (200M)}
        \end{subfigure}
        \par\vspace{1em}
        \begin{subfigure}[t]{\textwidth}
            \centering
            \includegraphics[width=\linewidth]{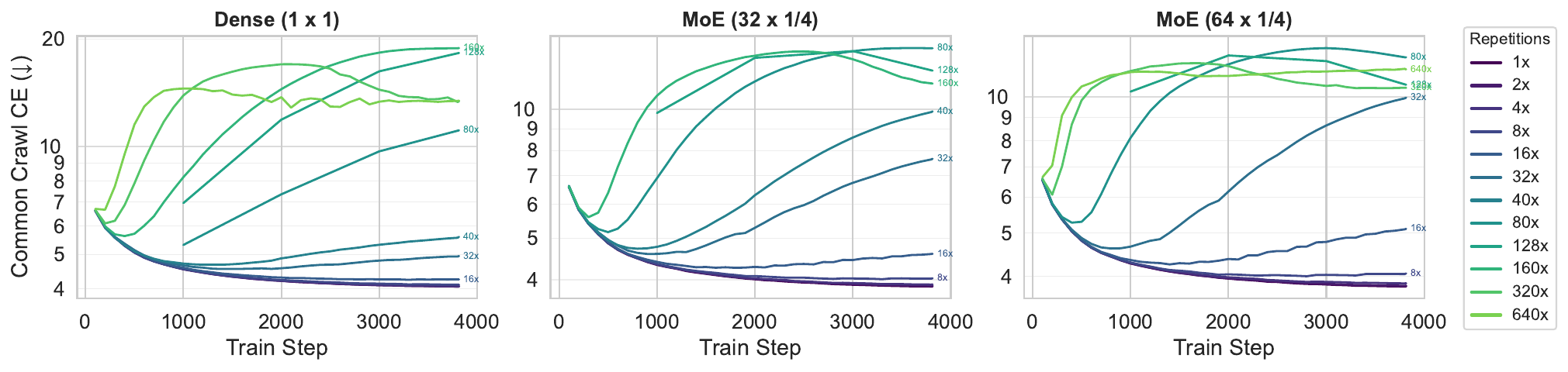}
            \subcaption{Validation loss over training (200M)}
        \end{subfigure}

\end{figure*}

\begin{figure*}[!ht]
    \centering
    \ContinuedFloat
    \begin{subfigure}[t]{\textwidth}
        \centering
        \includegraphics[width=0.7\linewidth]{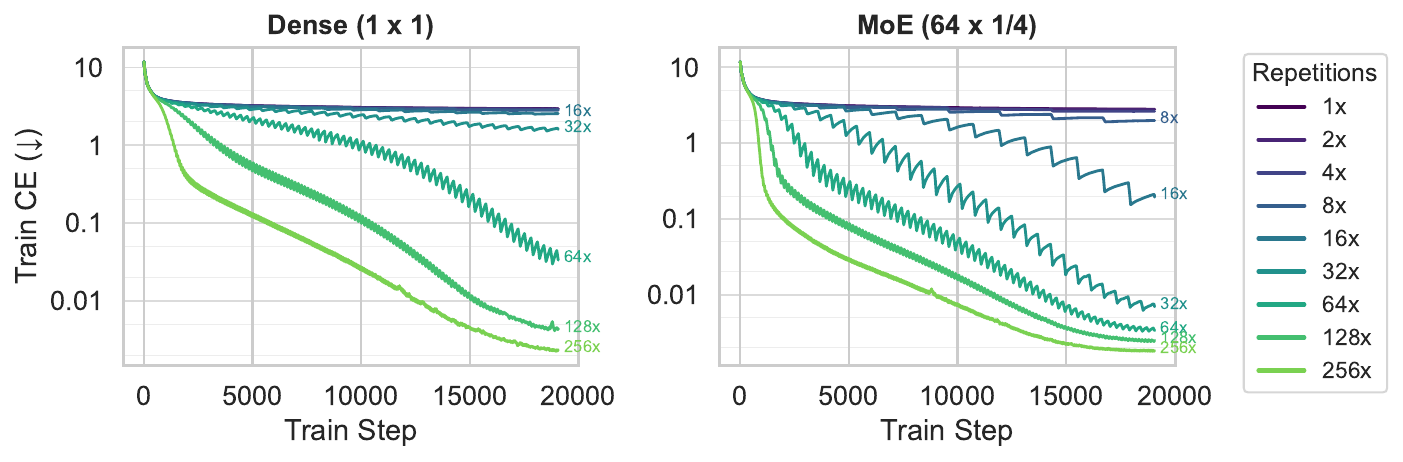}
        \subcaption{Train loss over training (1B)}
    \end{subfigure}
    \par\vspace{1em}
    \begin{subfigure}[t]{\textwidth}
        \centering
        \includegraphics[width=0.7\linewidth]{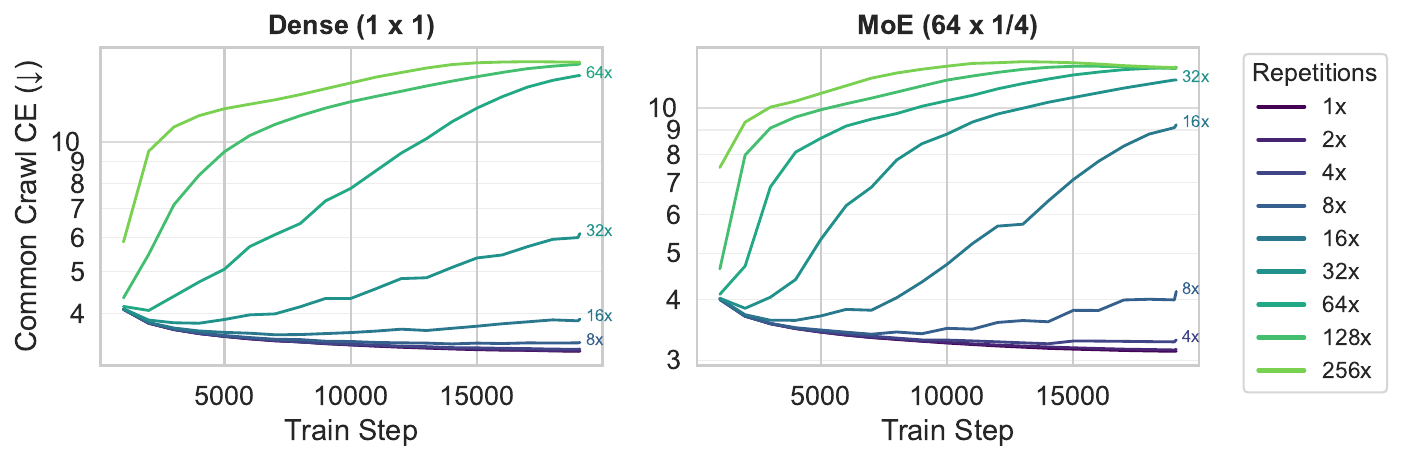}
        \subcaption{Validation loss over training (1B)}
    \end{subfigure}

    \caption{\textbf{At higher repetition rates, training loss falls to 0, which suggests overfitting to the repeated data (\S\ref{sec:expts_olmo_mix}).}
    We show the training and validation loss curves over the course of training. We consistently observe that higher repetition results in training loss curves that approach 0, mirrored by validation loss curves that rise. At sufficiently high $R$, we observe the double descent phenomenon \citet{muennighoff2023scaling}, in which validation curves peak then fall.
    }
    \label{fig:train_valid_app}
\end{figure*}

\clearpage
\subsection{Routing Load Balance and Stability}
\label{app:routing_lb}
\begin{figure*}[!ht]
    \centering
    \begin{subfigure}[t]{\textwidth}
        \begin{subfigure}[t]{\textwidth}
            \centering
            \includegraphics[width=\linewidth]{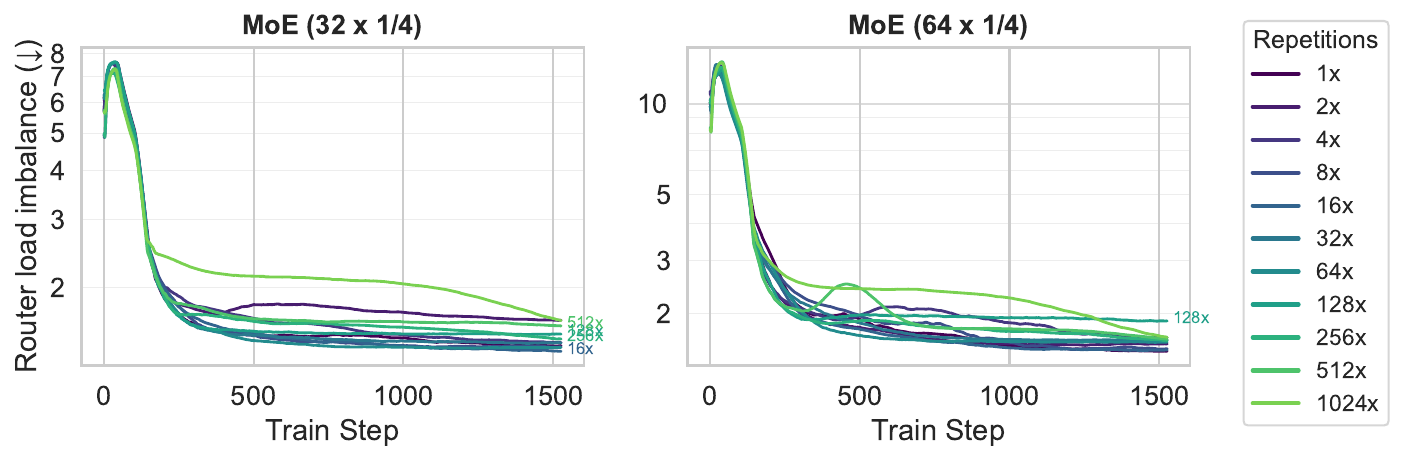}
            \subcaption{\textbf{80M active parameters}}
        \end{subfigure}
        \par\vspace{1em}
        \begin{subfigure}[t]{\textwidth}
            \centering
            \includegraphics[width=\linewidth]{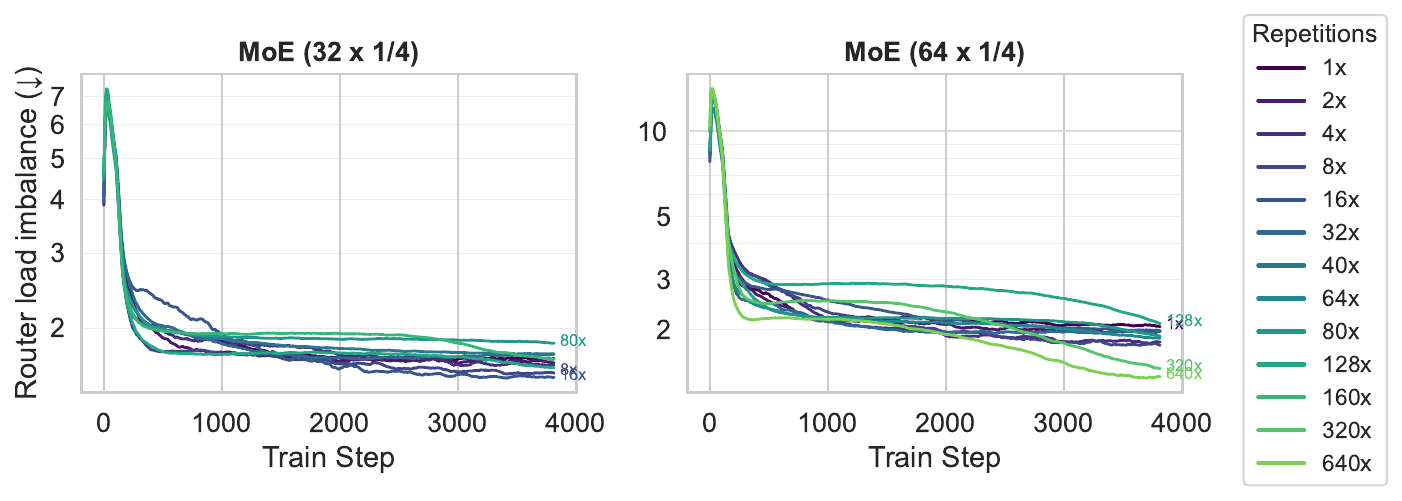}
            \subcaption{\textbf{200M active parameters}}
        \end{subfigure}
        \par\vspace{1em}
        \begin{subfigure}[t]{\textwidth}
            \centering
            \includegraphics[width=0.5\linewidth]{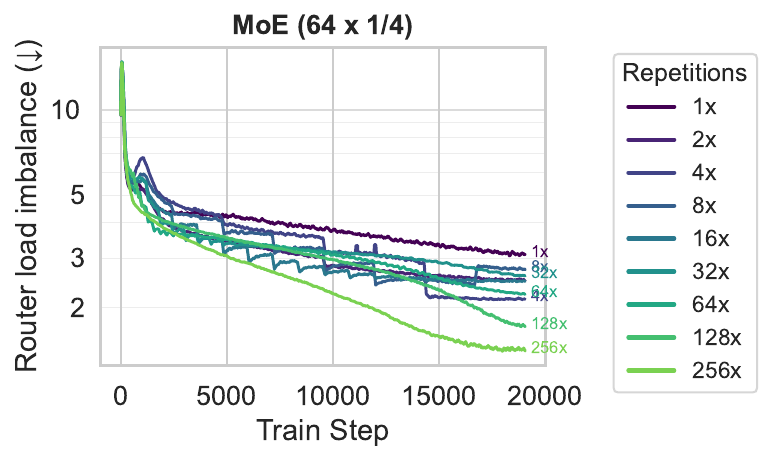}
            \subcaption{\textbf{1B active parameters}}
        \end{subfigure}
    \end{subfigure}
    \caption{\textbf{Routing imbalance training curves do not follow clear patterns at lower repetition, but are outliers at high repetition.}
    We plot the routing imbalance, defined as the ratio between the maximum and median expert load, averaged over tokens in the batch. At small scale, routing imbalance does not appear correlated with load imbalance until $R>128$, where curves become outliers. At 1B scale, load imbalance curves become disordered, and appear to partially cycle with data repetition periods.
    }
    \label{fig:routing_imbalance_app}
\end{figure*}

\begin{figure*}[!ht]
    \centering
    \begin{subfigure}[t]{\textwidth}
        \begin{subfigure}[t]{\textwidth}
            \centering
            \includegraphics[width=\linewidth]{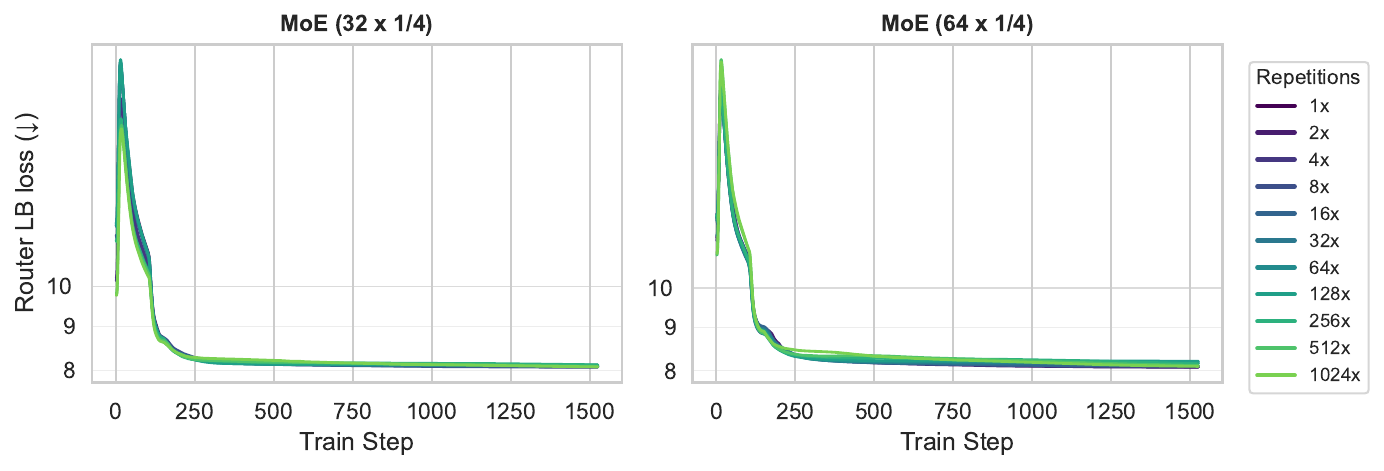}
            \subcaption{\textbf{80M active parameters}}
        \end{subfigure}
        \par\vspace{1em}
        \begin{subfigure}[t]{\textwidth}
            \centering
            \includegraphics[width=\linewidth]{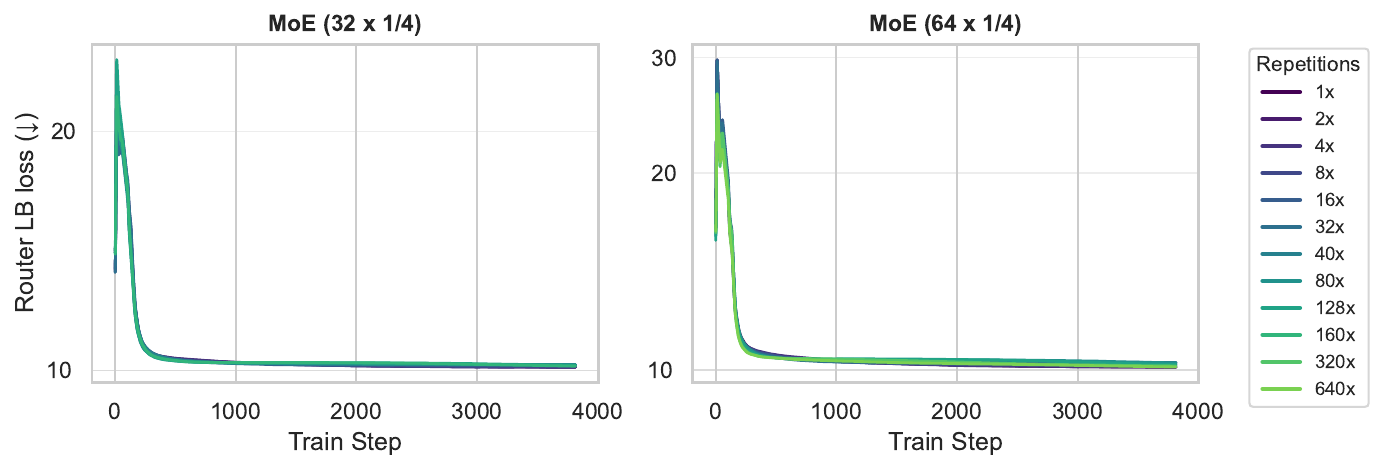}
            \subcaption{\textbf{200M active parameters}}
        \end{subfigure}
        \par\vspace{1em}
        \begin{subfigure}[t]{\textwidth}
            \centering
            \includegraphics[width=0.5\linewidth]{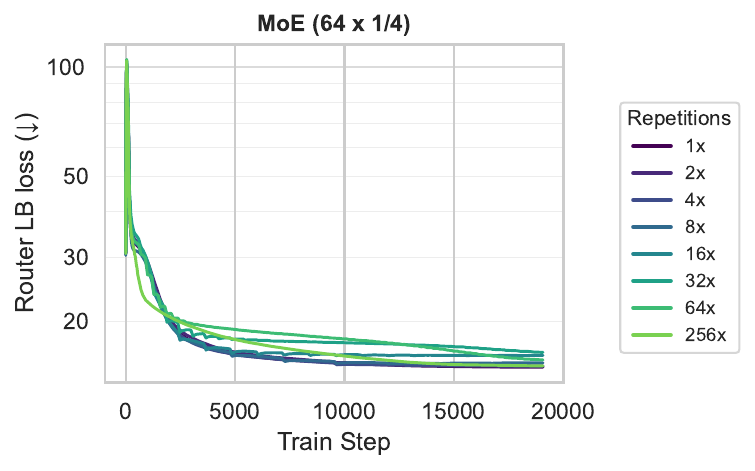}
            \subcaption{\textbf{1B active parameters}}
        \end{subfigure}
    \end{subfigure}
    \caption{\textbf{Routing load balancing loss training curves do not follow clear patterns at small model scale, but may correlate with repetition at larger scale.} We report the load balancing loss, as defined in \S\ref{sec:background}. At 80M and 200M, all settings show similar curves. At 1B, high repetition rates appear to affect load balancing loss, with intermediate values of $R=8, 16, 32$ resulting in periodicity, and outlier curves at $R=64, 128, 256$. It is possible that repetition itself increases load balancing loss, but that the extremely low training loss at high $R$ results in a relatively strong optimization signal from auxiliary losses, eventually driving load balancing loss to fall.}
    \label{fig:routing_lbloss_app}
\end{figure*}

\begin{figure*}[!ht]
    \centering
    \begin{subfigure}[t]{\textwidth}
        \begin{subfigure}[t]{\textwidth}
            \centering
            \includegraphics[width=\linewidth]{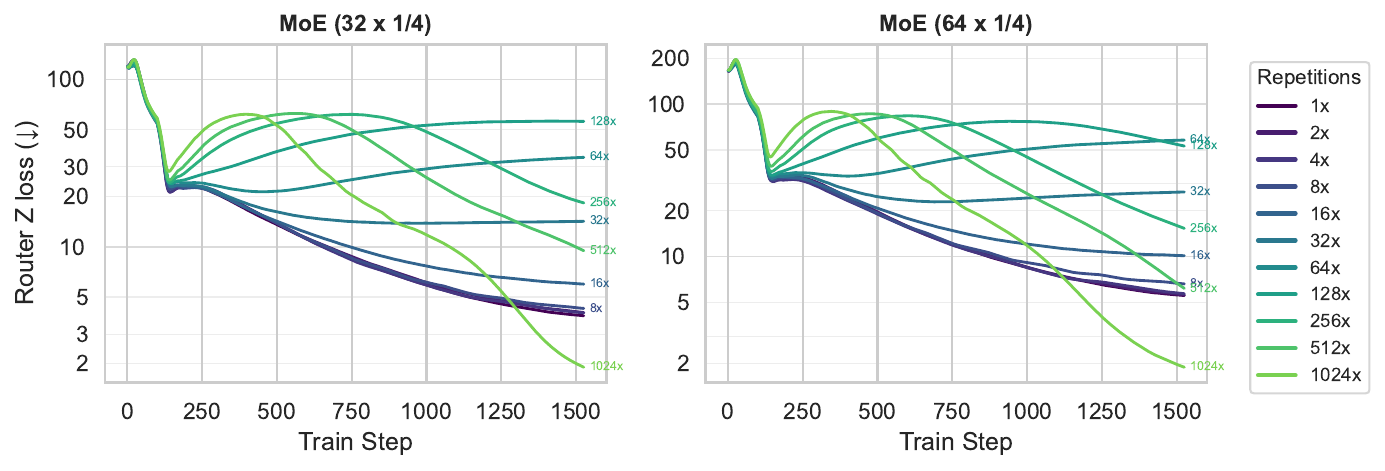}
            \subcaption{\textbf{80M active parameters}}
        \end{subfigure}
        \par\vspace{1em}
        \begin{subfigure}[t]{\textwidth}
            \centering
            \includegraphics[width=\linewidth]{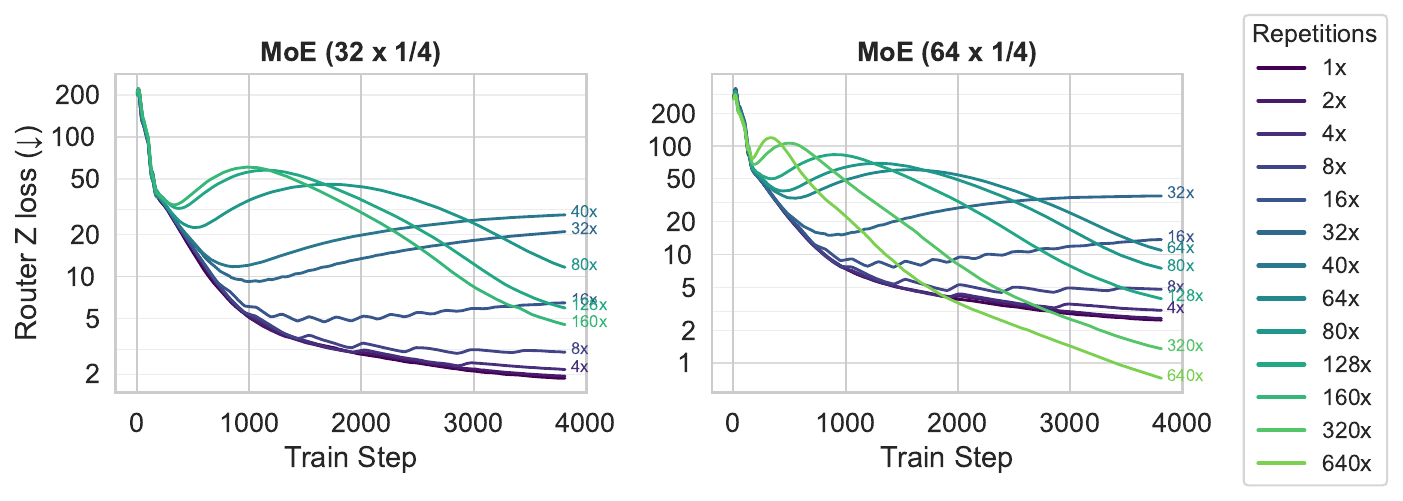}
            \subcaption{\textbf{200M active parameters}}
        \end{subfigure}
        \par\vspace{1em}
        \begin{subfigure}[t]{\textwidth}
            \centering
            \includegraphics[width=0.5\linewidth]{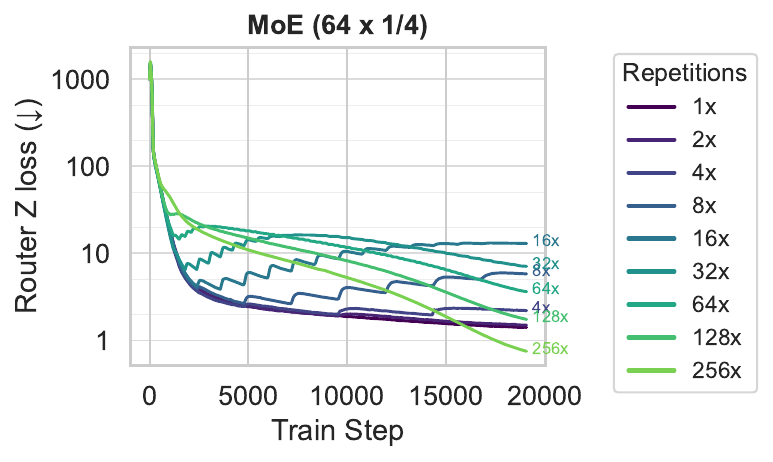}
            \subcaption{\textbf{1B active parameters}}
        \end{subfigure}
    \end{subfigure}
    \caption{\textbf{Router z-loss training curves cycle with data repetition.} We report the router z-loss \citep{zoph2022stmoe}. High repetition rates appear to affect z-loss, data-repetition driven cycles at 200M and 1B scale. Higher repetition rates result in higher z-loss with a double peak relatively early in training, resolving to a lower final z-loss. It is possible that high repetition typically results in higher z-loss, but that the extremely low training loss at high $R$ results in a relatively strong optimization signal from auxiliary losses, eventually driving z-loss to fall.}
    \label{fig:routing_zloss_app}
\end{figure*}

\clearpage
\subsection{Regularizers (\S\ref{sec:reg})}
\label{app:reg}

\begin{figure*}[!ht]
    \centering
    \begin{subfigure}[t]{0.75\textwidth}
        \begin{subfigure}[t]{\textwidth}
            \centering
            \includegraphics[width=\linewidth]{fig_pdfs/regularizers/olmo_mix__80M__dense_moe64__dropout__byarch__eval_dolma_cc_loss.pdf}
        \end{subfigure}
        \subcaption{Dropout}
        \label{fig:regularizers_app:dropout}
    \end{subfigure}
        
    \par\vspace{1em}
    \begin{subfigure}[t]{0.73\textwidth}
        \begin{subfigure}[t]{\textwidth}
            \centering
            \includegraphics[width=\linewidth]{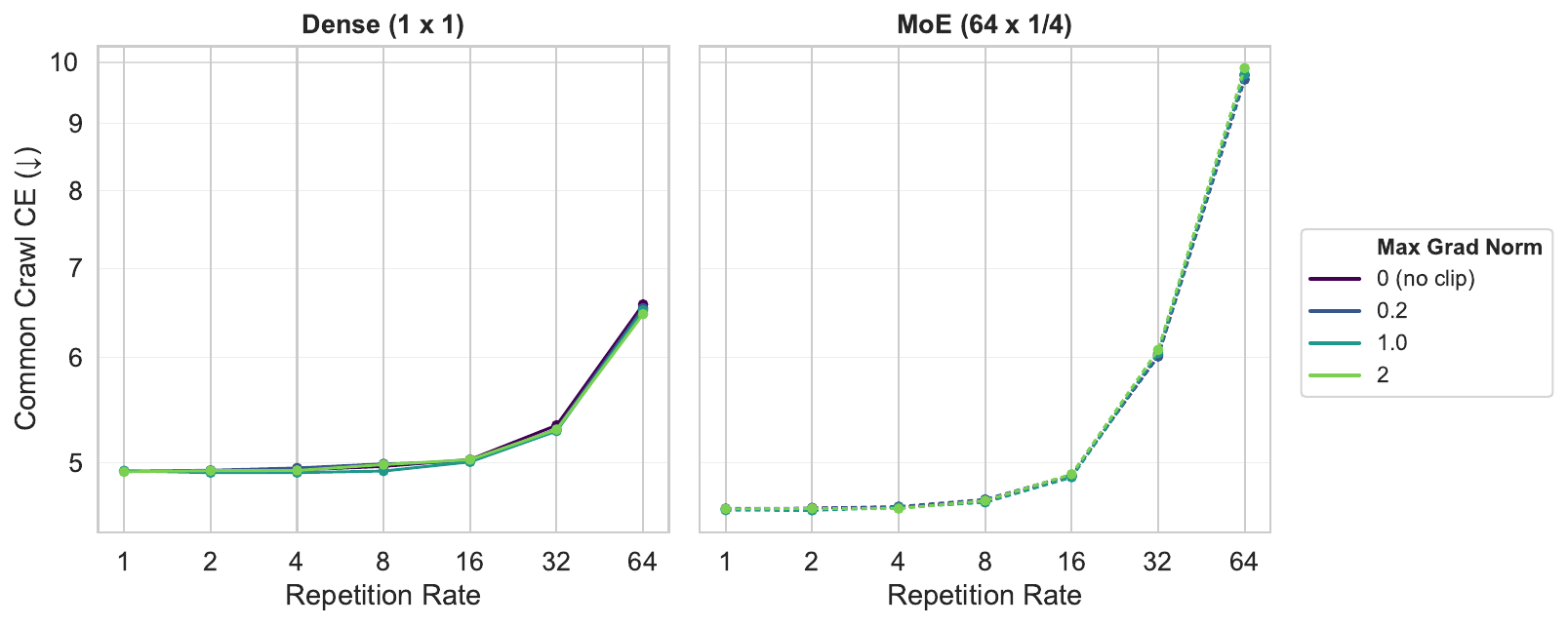}
        \end{subfigure}
        \subcaption{Gradient Norm Clipping}
        \label{fig:regularizers_app:grad_norm}
    \end{subfigure}
    
    \par\vspace{1em}
    \begin{subfigure}[t]{0.73\textwidth}
        \begin{subfigure}[t]{\textwidth}
            \centering
            \includegraphics[width=\linewidth]{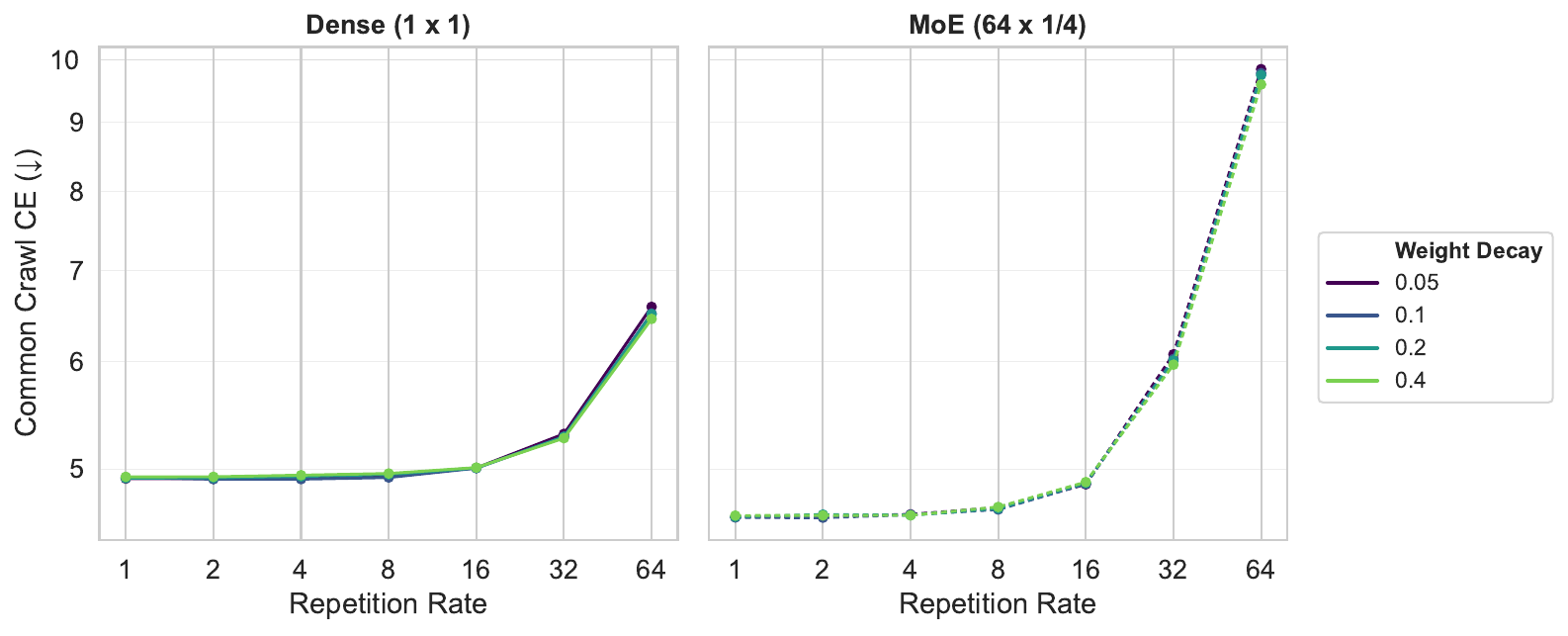}
        \end{subfigure}
        \subcaption{Weight Decay}
        \label{fig:regularizers_app:weight_decay}
    \end{subfigure}
        
    \par\vspace{1em}
    \begin{subfigure}[t]{0.73\textwidth}
        \centering
        \includegraphics[width=\linewidth]{fig_pdfs/regularizers/olmo_mix__80M__dense_moe64__fom_prob__byarch__eval_dolma_cc_loss.pdf}
        \subcaption{FFN Output Masking}
    \label{fig:regularizers_app:fom}
    \end{subfigure}
   
\end{figure*}

\begin{figure*}[!ht]
    \ContinuedFloat
    \centering
    
    \begin{subfigure}[t]{0.75\textwidth}
        \centering
        \includegraphics[width=\linewidth]{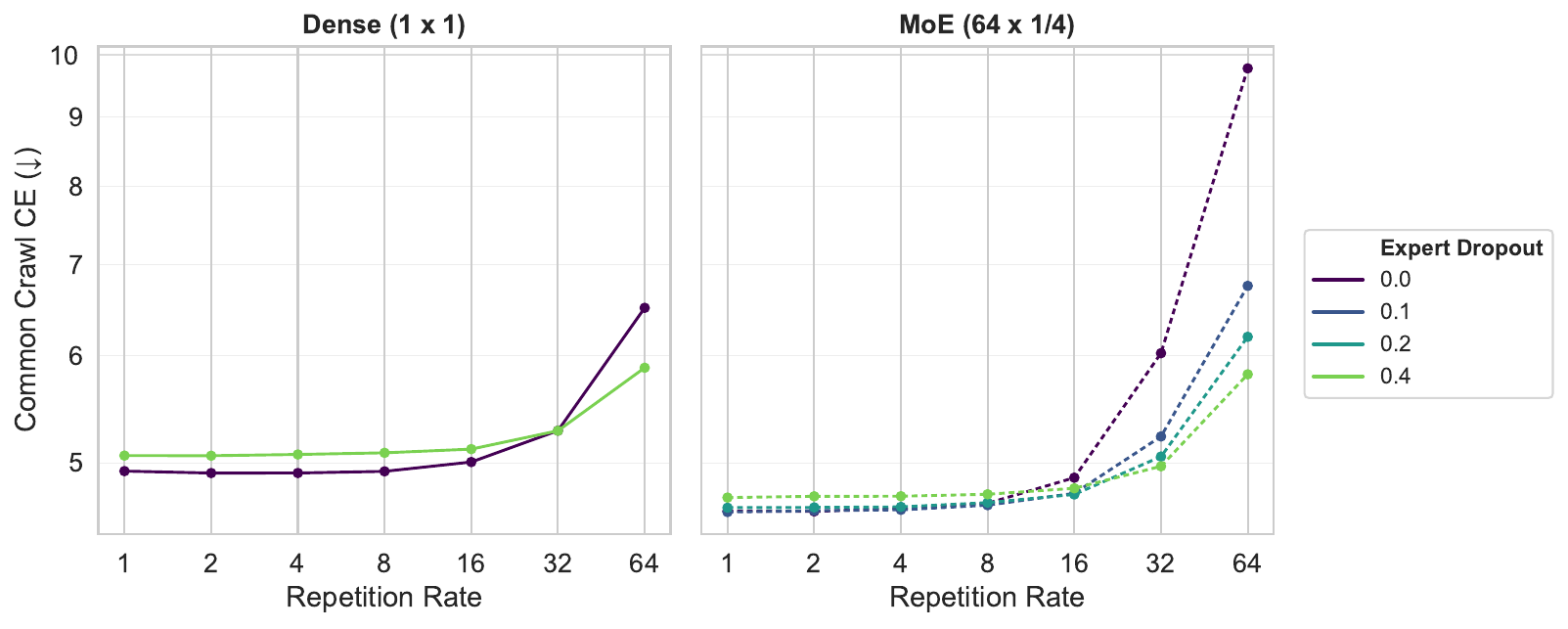}
        \subcaption{Expert Dropout}
    \label{fig:regularizers_app:expert_dropout}
    \end{subfigure}    
    \par\vspace{1em}
    \begin{subfigure}[t]{0.75\textwidth}
        \centering
        \includegraphics[width=\linewidth]{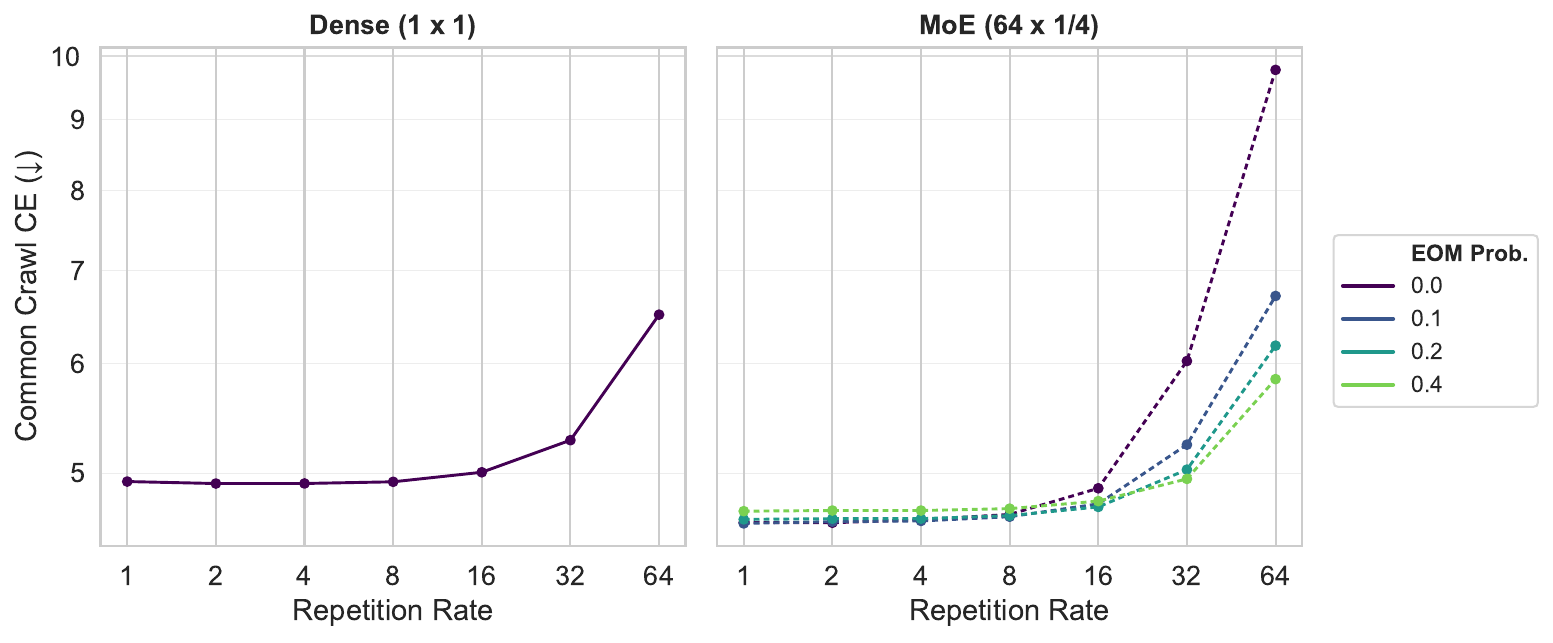}
        \subcaption{Expert Output Masking}
    \label{fig:regularizers_app:eom}
    \end{subfigure}    
    \par\vspace{1em}
    \begin{subfigure}[t]{0.75\textwidth}
        \centering
        \includegraphics[width=\linewidth]{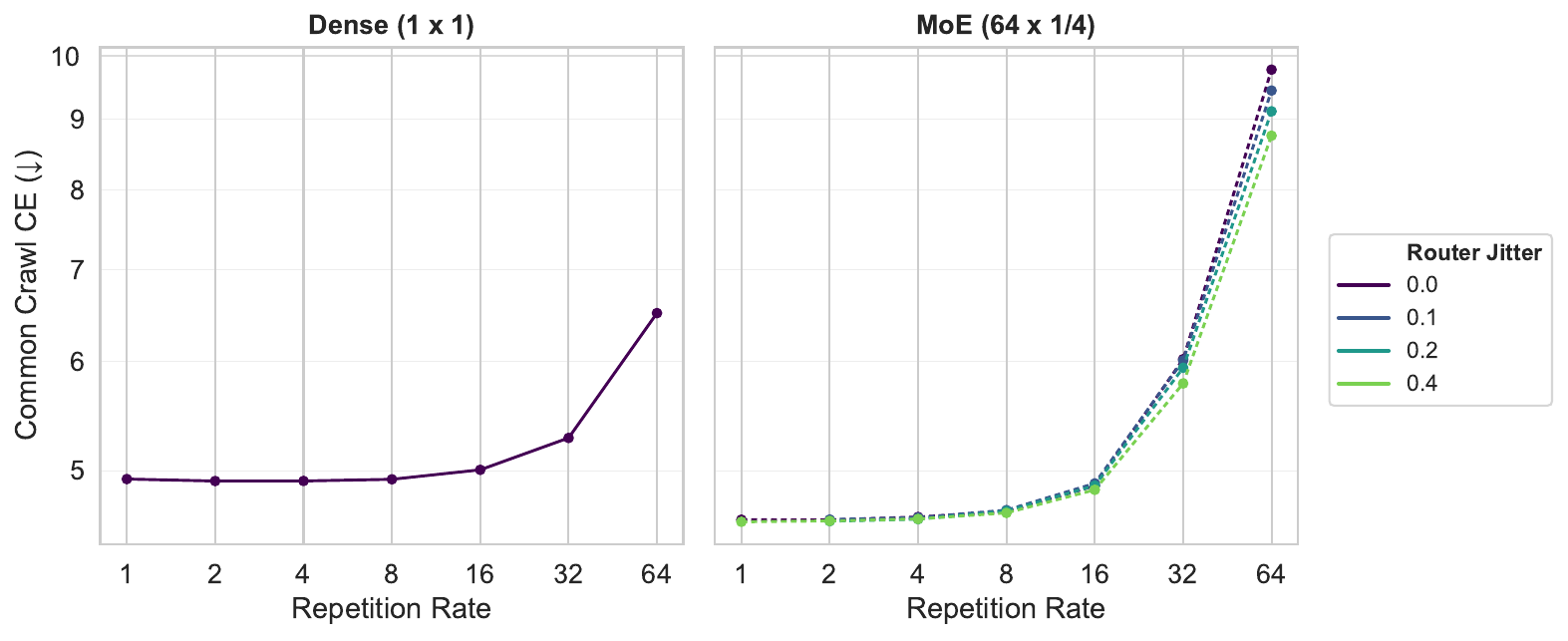}
        \subcaption{MoE Router Jitter}
    \label{fig:regularizers_app:router_jitter}
    \end{subfigure}
    \caption{
    \textbf{Dropout (a), FFN Output Masking (d), Expert Dropout (e), and Expert Output Masking (f) each reduces overfitting from data repetition (\S\ref{sec:reg}).}
    Of the regularization methods studied in \S\ref{sec:reg}, these 4 dramatically decrease the response to data repetition. However, weight decay (b) and gradient norm clipping (c), as well as MoE router jitter (f), have minimal effect. 
    }
    \label{fig:regularizers_app}
\end{figure*}

\clearpage
\subsection{Data Filtering (\S\ref{sec:expts_filtering})}
\label{app:filtering}
\begin{figure*}[!ht]
    \centering
        \begin{subfigure}[t]{\textwidth}
        \centering
            \begin{subfigure}[t]{0.49\textwidth}
                \centering
                \includegraphics[width=\linewidth]{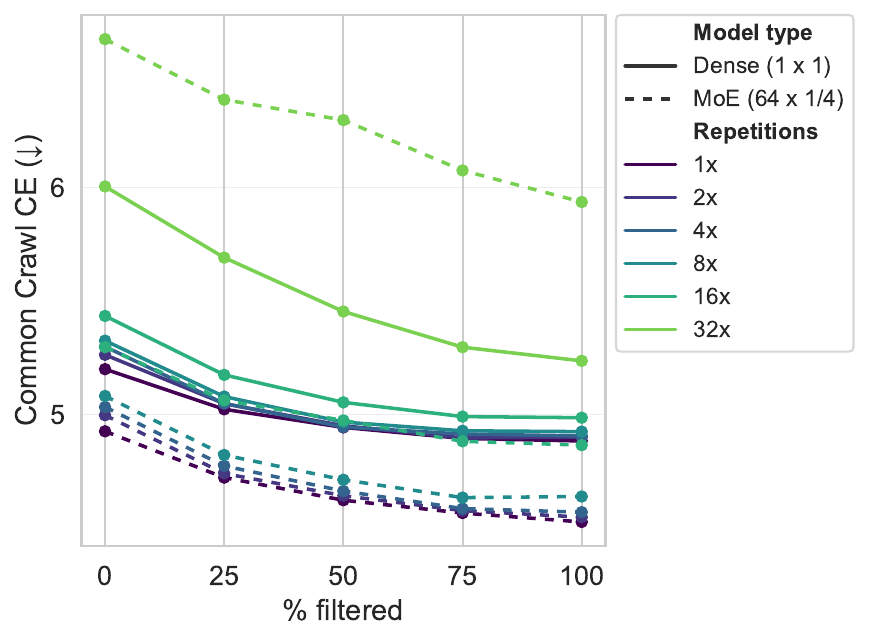}
            \end{subfigure}
            \hfill
            \begin{subfigure}[t]{0.49\textwidth}
                \centering
                \includegraphics[width=\linewidth]{fig_pdfs/filtering/famAp__80M__dense_moe64__repetition__eval_dolma_cc_loss.pdf}
            \end{subfigure}
            \subcaption{Dolma Common Crawl Validation CE Loss}
        \end{subfigure}
        \par\vspace{1em}
        \begin{subfigure}[t]{\textwidth}
        \centering
            \begin{subfigure}[t]{0.49\textwidth}
                \centering
                \includegraphics[width=\linewidth]{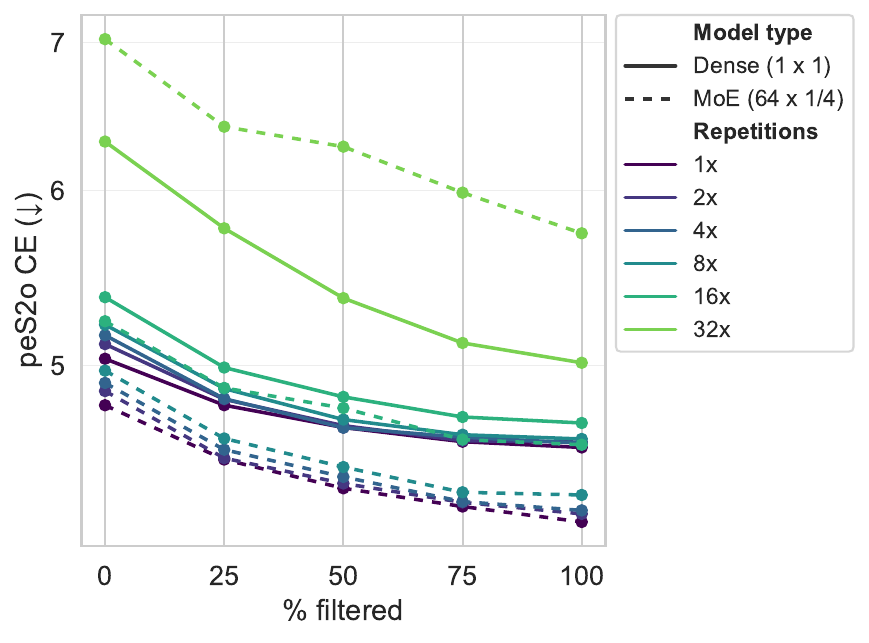}
            \end{subfigure}
            \hfill
            \begin{subfigure}[t]{0.49\textwidth}
                \centering
                \includegraphics[width=\linewidth]{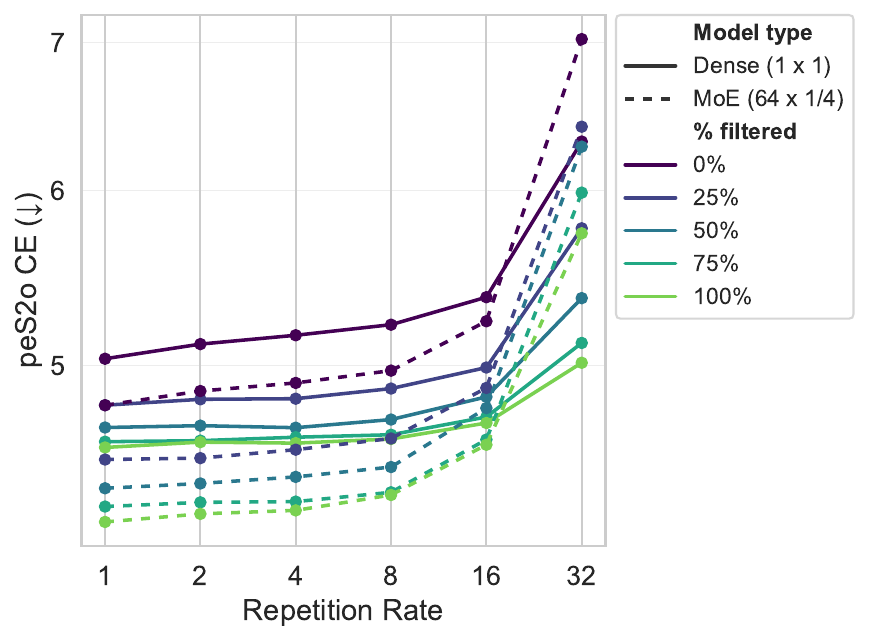}
            \end{subfigure}
            \subcaption{Dolma peS2o Validation CE Loss}
        \end{subfigure}
        \par\vspace{1em}
        \begin{subfigure}[t]{\textwidth}
        \centering
            \begin{subfigure}[t]{0.49\textwidth}
                \centering
                \includegraphics[width=\linewidth]{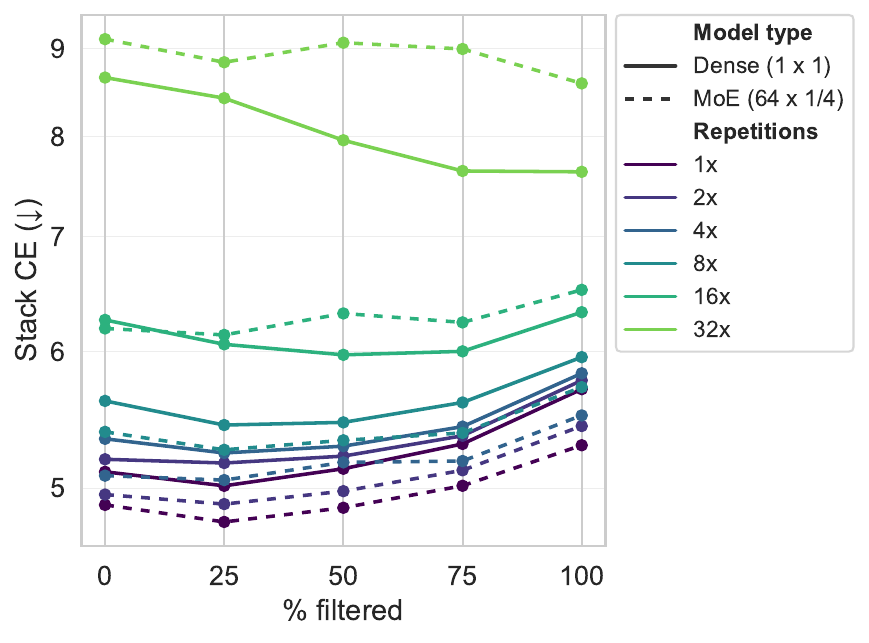}
            \end{subfigure}
            \hfill
            \begin{subfigure}[t]{0.49\textwidth}
                \centering
                \includegraphics[width=\linewidth]{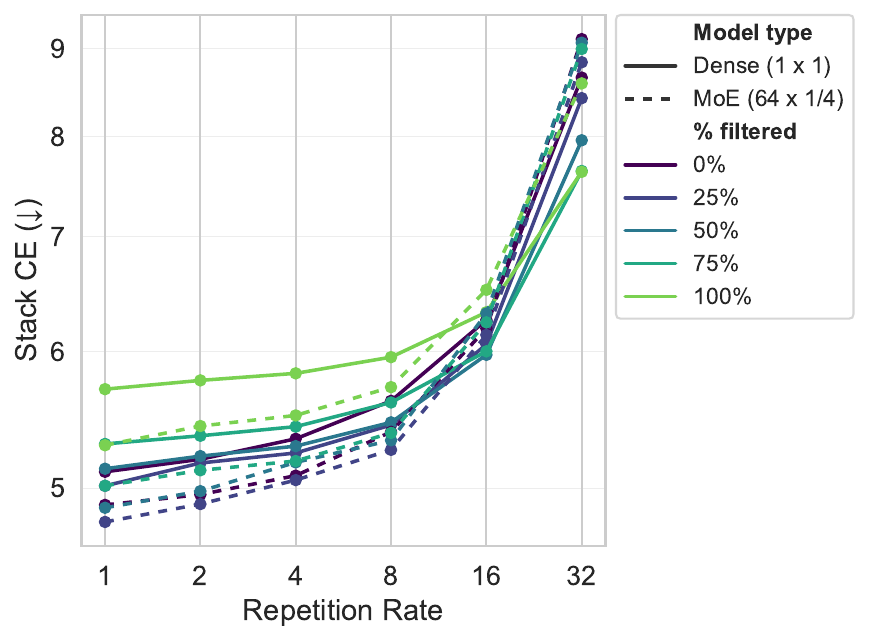}
            \end{subfigure}
            \subcaption{Dolma Stack Validation CE Loss}
        \end{subfigure}
    \caption{
    }
    \label{fig:filtering_app}
\end{figure*}

\clearpage
\subsection{Interpolating Between a Single Domain and a Mix}
\label{app:mix_to_single}
We consider an experimental setup similar to \S\ref{sec:expts_filtering}, but with interpolation between DCLM-baseline and a data mix. We use Dolma 1.7 \citep{dolma}, which is similar to the OLMoE mix (\S\ref{app:train_data_sources}; used in \S\ref{sec:expts}), but includes more sources. Dolma 1.7 and DCLM-baseline do not fall on a spectrum of quality, but rather of homogeneity. Unlike our experiments of \S\ref{sec:expts_mixes}, which are carefully controlled examinations of 2-domain mixes, the heterogeneity of Dolma 1.7 more closely resembles that of data mixes used for frontier LMs.
\begin{figure*}[!ht]
    \centering
        \begin{subfigure}[t]{\textwidth}
        \centering
            \begin{subfigure}[t]{0.49\textwidth}
                \centering
                \includegraphics[width=\linewidth]{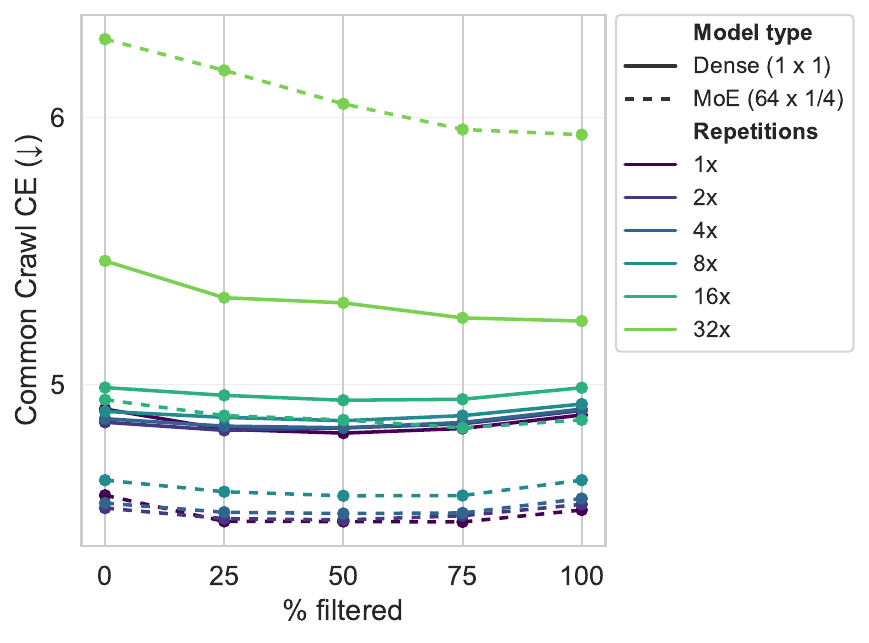}
            \end{subfigure}
            \hfill
            \begin{subfigure}[t]{0.49\textwidth}
                \centering
                \includegraphics[width=\linewidth]{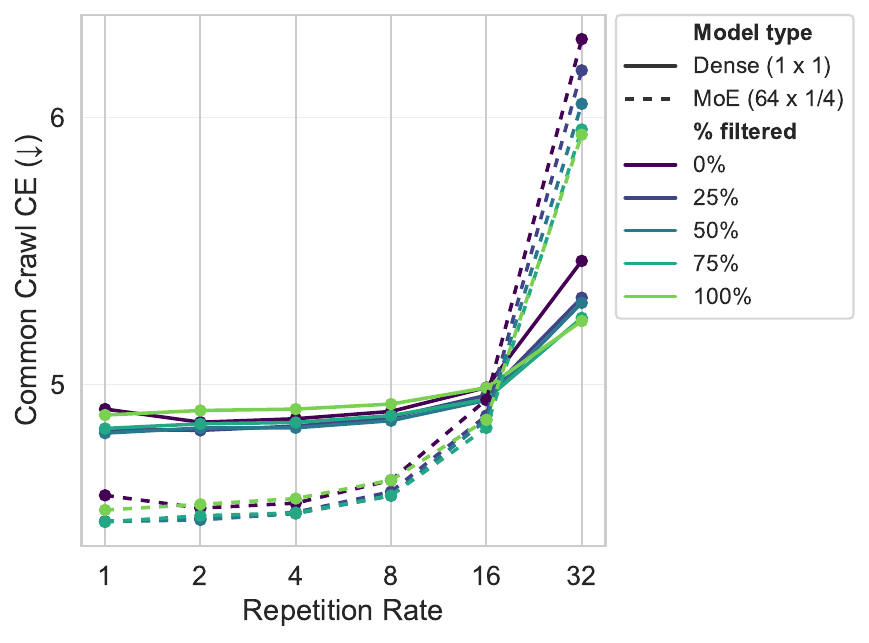}
            \end{subfigure}
            \subcaption{Dolma Common Crawl Validation CE Loss}
        \end{subfigure}
        \par\vspace{1em}
        \begin{subfigure}[t]{\textwidth}
        \centering
            \begin{subfigure}[t]{0.49\textwidth}
                \centering
                \includegraphics[width=\linewidth]{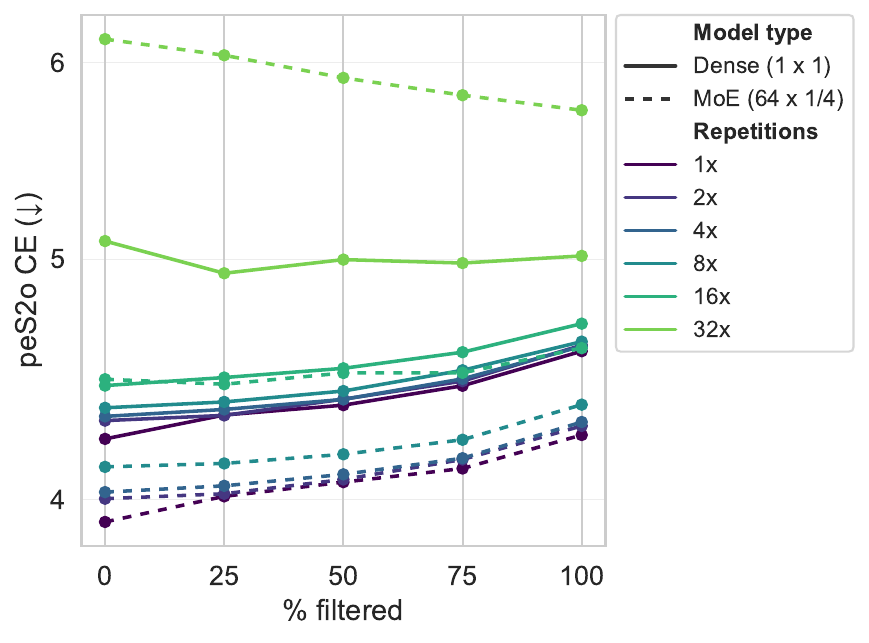}
            \end{subfigure}
            \hfill
            \begin{subfigure}[t]{0.49\textwidth}
                \centering
                \includegraphics[width=\linewidth]{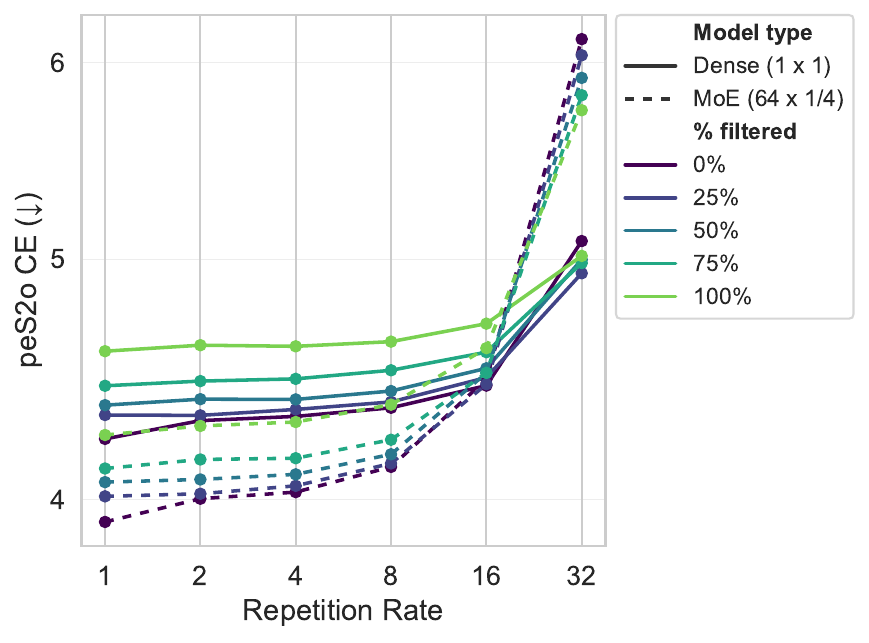}
            \end{subfigure}
            \subcaption{Dolma peS2o Validation CE Loss}
        \end{subfigure}
        \par\vspace{1em}
        \begin{subfigure}[t]{\textwidth}
        \centering
            \begin{subfigure}[t]{0.49\textwidth}
                \centering
                \includegraphics[width=\linewidth]{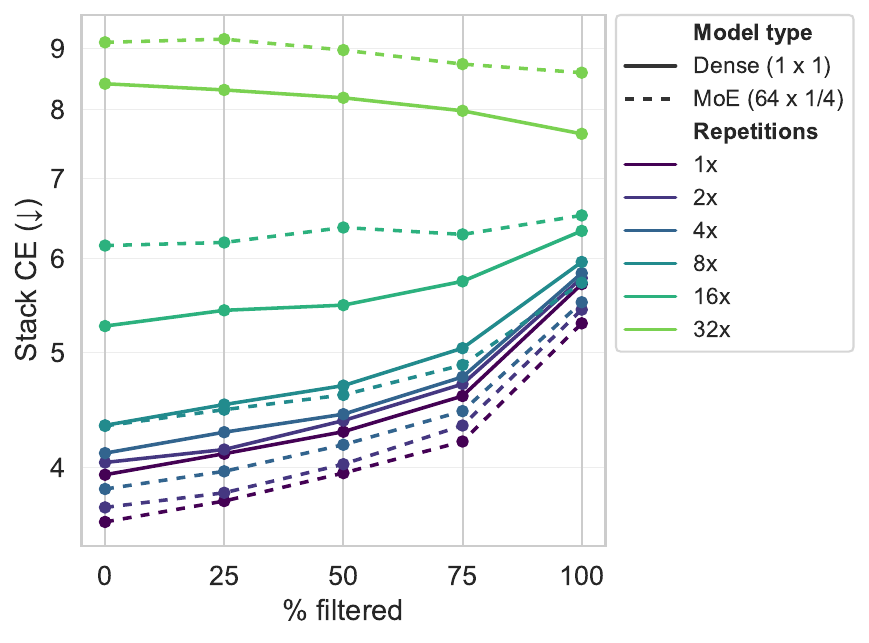}
            \end{subfigure}
            \hfill
            \begin{subfigure}[t]{0.49\textwidth}
                \centering
                \includegraphics[width=\linewidth]{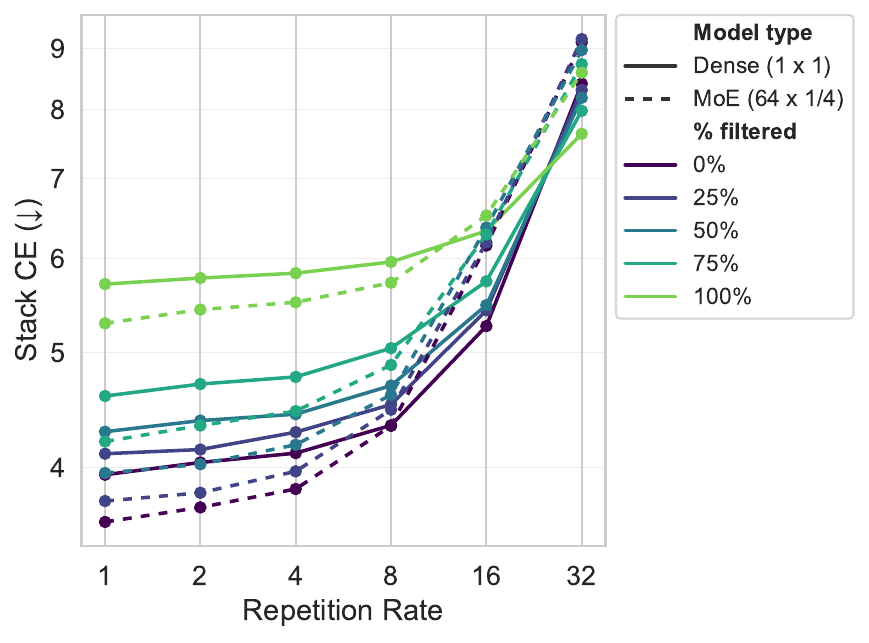}
            \end{subfigure}
            \subcaption{Dolma Stack Validation CE Loss}
        \end{subfigure}
    \caption{
    }
    \label{fig:mix_to_single}
\end{figure*}

\clearpage
\subsection{Mechanistic Analyses of Routing (\S\ref{sec:analysis})}
\label{app:routing_extra}

\begin{figure*}[!ht]
    \centering
    \begin{subfigure}[t]{\textwidth}
        \begin{subfigure}[t]{0.45\textwidth}
            \centering
            \includegraphics[width=\linewidth]{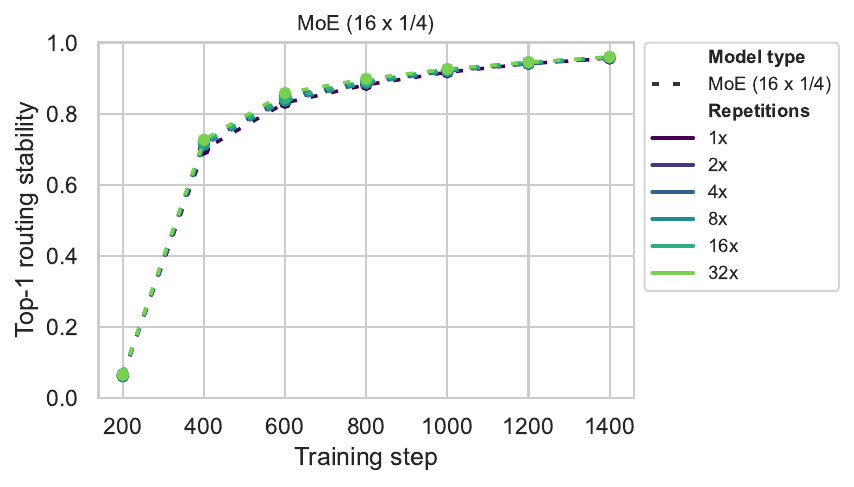}
            \subcaption{MoE (16 x 1/4)}
        \end{subfigure}
        \hfill
        \begin{subfigure}[t]{0.45\textwidth}
            \centering
            \includegraphics[width=\linewidth]{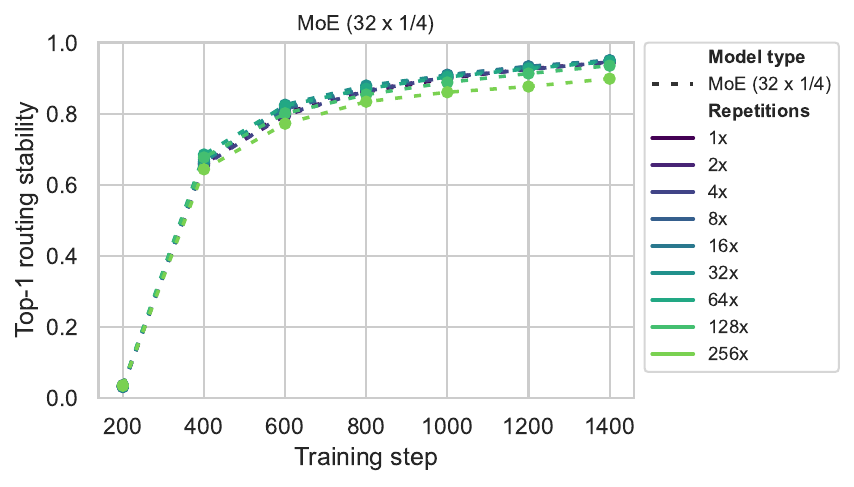}
            \subcaption{MoE (32 x 1/4)}
        \end{subfigure}
    
    \par\vspace{1em}
        \begin{subfigure}[t]{0.45\textwidth}
            \centering
            \includegraphics[width=\linewidth]{fig_pdfs/routing/routing_ossification_curves__80M__moe64__dolma_cc.pdf}
            \subcaption{MoE (64 x 1/4)}
        \end{subfigure}
        \hfill
        \begin{subfigure}[t]{0.45\textwidth}
            \centering
            \includegraphics[width=\linewidth]{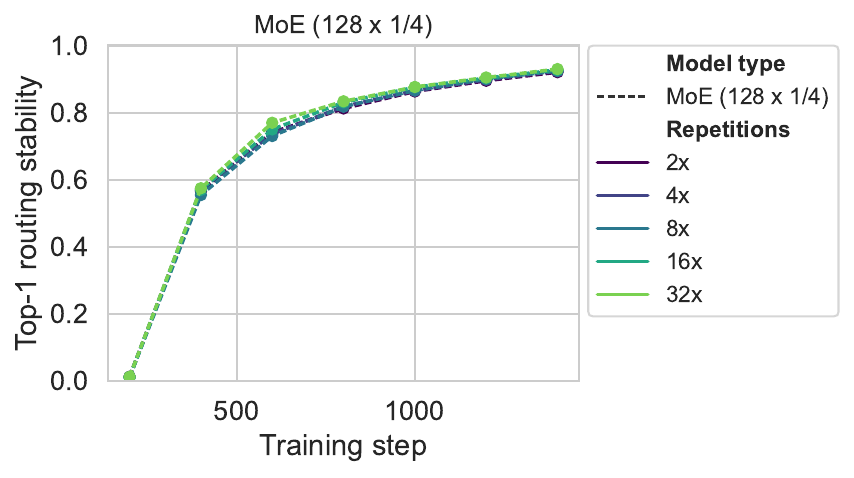}
            \subcaption{MoE (128 x 1/4)}
        \end{subfigure}
    
    \par\vspace{1em}
        \begin{subfigure}[t]{0.45\textwidth}
            \centering
            \includegraphics[width=\linewidth]{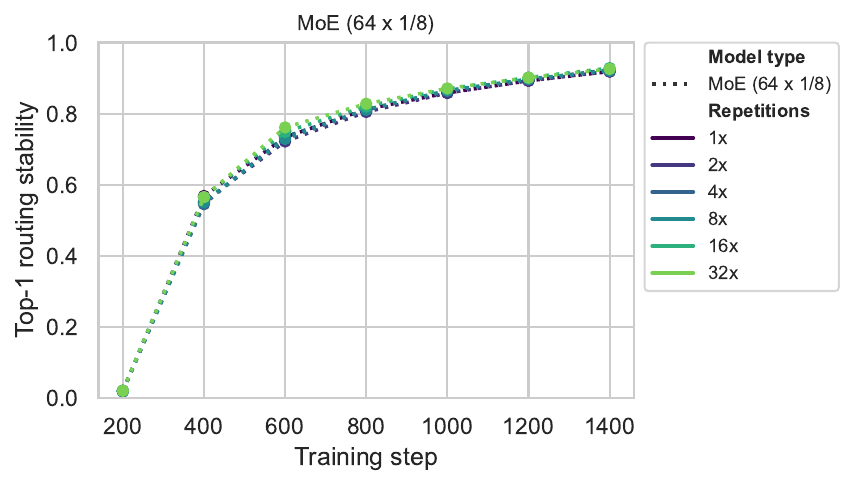}
            \subcaption{MoE (64 x 1/8)}
        \end{subfigure} 
        \hfill
        \begin{subfigure}[t]{0.45\textwidth}
            \centering
            \includegraphics[width=\linewidth]{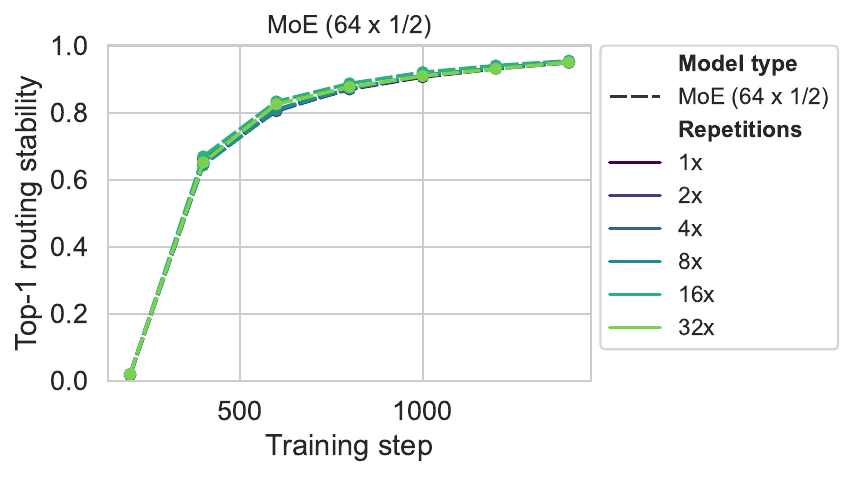}
            \subcaption{MoE (16 x 1/2)}
        \end{subfigure}
    \end{subfigure}
    \caption{\textbf{Routing ossifies early in training, exacerbated by repetition (\S\ref{sec:analysis_router_oss}).}
    We show Top-1 routing stability, which we define as the fraction of held-out Common Crawl tokens that keep the same top-1 expert between consecutive checkpoints (200 steps apart), averaged over layers, for the 80M MoE models. At the beginning of training, consecutive checkpoint agreement is near chance, at roughly $\frac{1}{n}$ for all MoE (n x g) configurations. However, routing stability rises rapidly for all settings, with fewer than 25\% of tokens routed to a different top-1 expert when comparing step 400 to 600. Router ossification is slightly higher with fewer experts (lower $n$) or with higher granularity (larger $g$). Stability at each checkpoint also increases with repetition rate $R$, up to roughly $R=16, 32$, where the router appears to destabilize.
    }
    \label{fig:router_oss_app}
\end{figure*}

\begin{figure*}[!ht]
    \centering
    \begin{subfigure}[t]{\textwidth}
        \begin{subfigure}[t]{\textwidth}
            \centering
            \includegraphics[width=\linewidth]{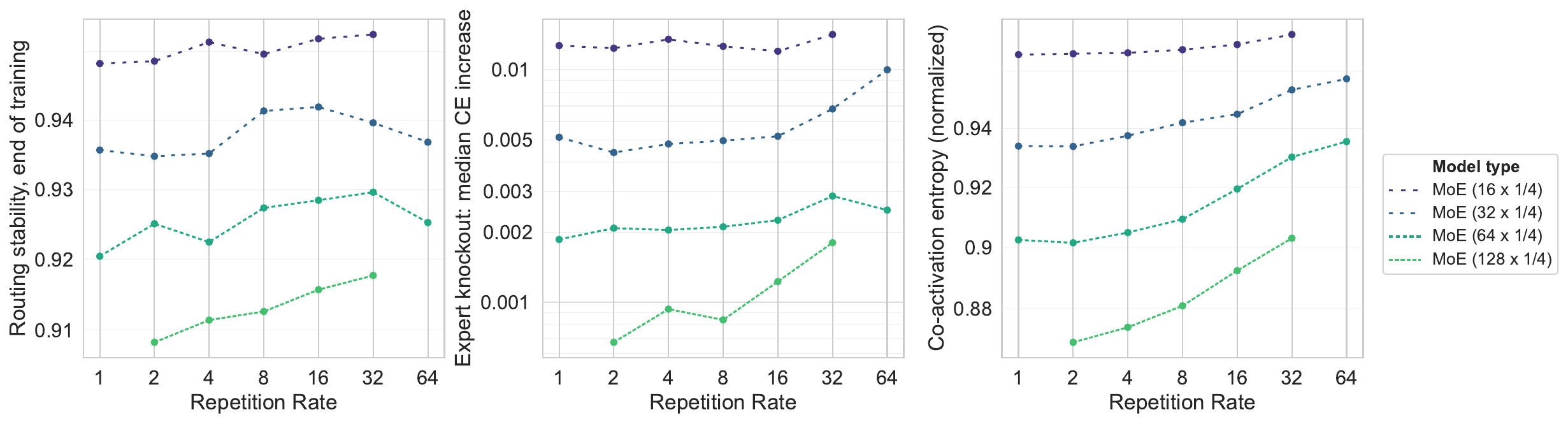}
        \end{subfigure}
        
        \begin{subfigure}[t]{\textwidth}
            \centering
            \includegraphics[width=\linewidth]{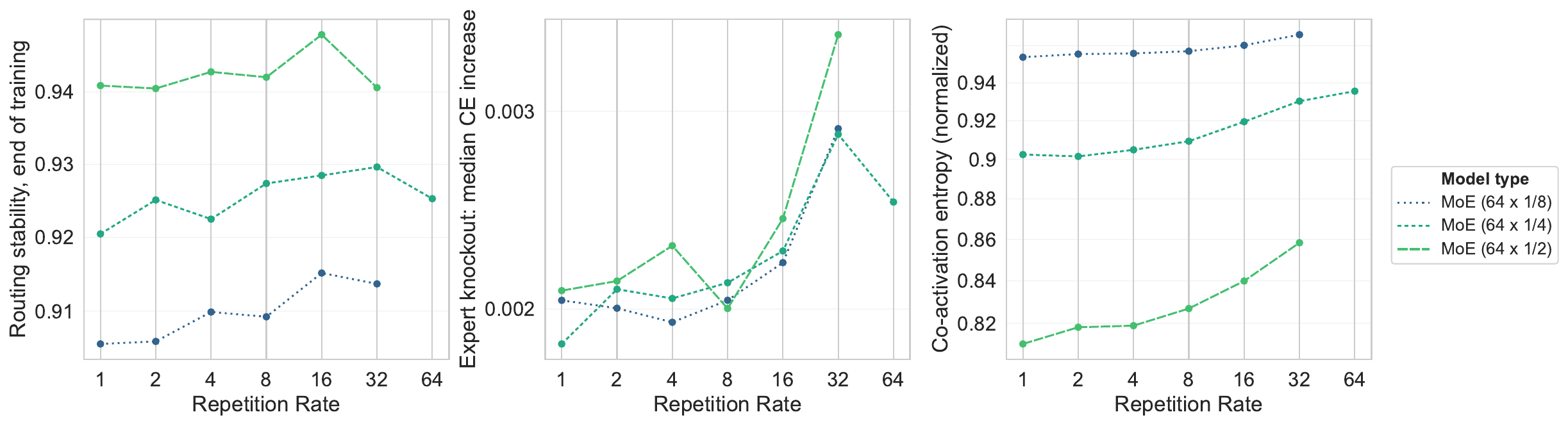}
        \end{subfigure}
        
    \end{subfigure}
    \caption{\textbf{Routing stability, expert specialization, and expert-coactivation entropy rise slightly with repetition rate (\S\ref{sec:analysis}).} We show more MoE configurations for (a) end-of-training routing stability, (b) expert knockout effect, and (c) expert co-activation entropy.
    (a) End-of-training routing stability increases slightly with repetition rate $R$. Higher expert count naturally results in lower stability, as there are more experts to choose from. Fewer active, but larger experts slightly decreases stability.
    (b) Expert specialization increases slightly with $R$. Higher expert count results in lower specialization. Varying active expert count along with expert size has no clear effect.
    (c) Co-activation entropy of expert pairs (normalized by maximum) rises steadily with $R$ towards uniformly distributed pairings. Entropy is higher with fewer total experts, and with a larger active number of smaller experts.}
    \label{fig:routing_extra_80m}
\end{figure*}

\begin{figure*}[!ht]
    \centering
    \begin{subfigure}[t]{\textwidth}
    \centering
        
            \centering
            \includegraphics[width=\linewidth]{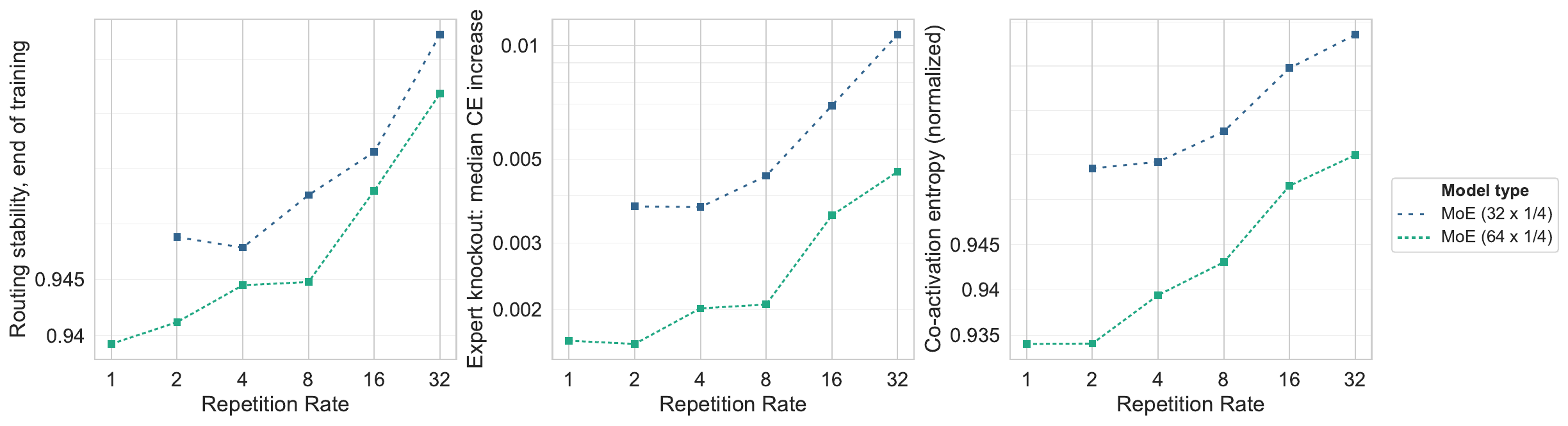}
    \end{subfigure}
    \caption{\textbf{Routing stability, expert specialization, and expert-coactivation entropy rise slightly with repetition rate at 200M active parameters (\S\ref{sec:analysis}).} We show 200M MoE (32 x 1/4) and MoE (64 x 1/4) configurations for (a) end-of-training routing stability, (b) expert knockout effect, and (c) expert co-activation entropy.
    (a) End-of-training routing stability increases slightly with repetition rate $R$. Higher expert count naturally results in lower stability, as there are more experts to choose from.
    (b) Expert specialization increases slightly with $R$. Higher expert count results in lower specialization. 
    (c) Co-activation entropy of expert pairs (normalized by maximum) rises steadily with $R$ towards uniformly distributed pairings. Entropy is higher with fewer total experts.}
    \label{fig:routing_extra_200m}
\end{figure*}

\begin{figure*}[!ht]
    \centering
    \begin{subfigure}[t]{\textwidth}
            \centering
        \begin{subfigure}[t]{\textwidth}
            \centering
            \includegraphics[width=\linewidth]{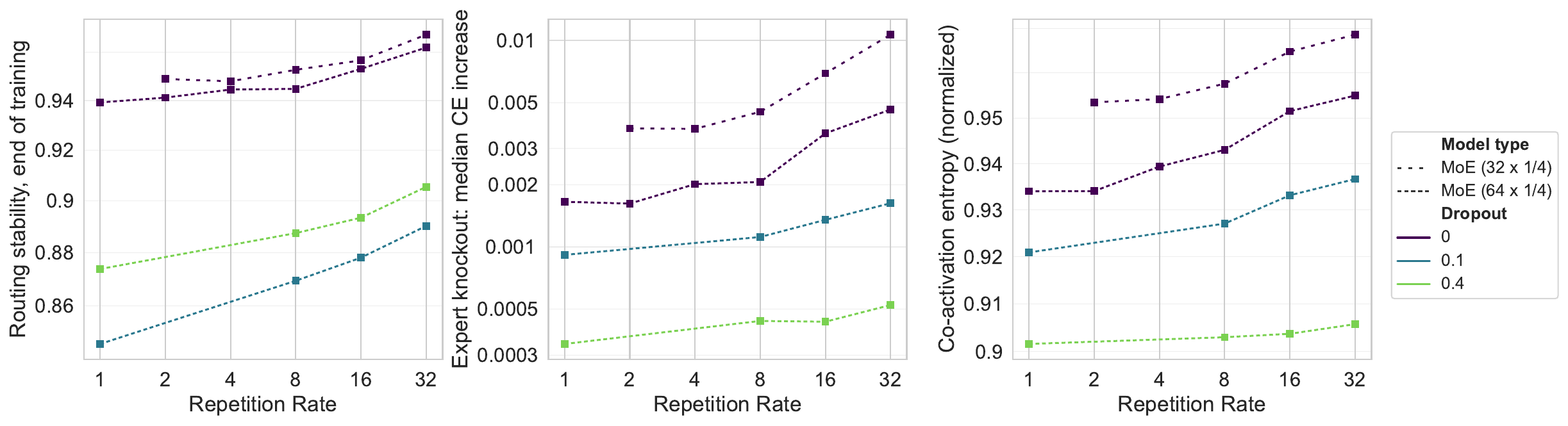}
        \end{subfigure} 
    \end{subfigure}
    \caption{\textbf{Dropout increases routing stability and expert-coactivation entropy, but decreases expert specialization (\S\ref{sec:analysis}).} We show, for various dropout settings on our 200M MoE (64 x 1/4) configurations: (a) end-of-training routing stability, (b) expert knockout effect, and (c) expert co-activation entropy.
    (a) End-of-training routing stability increases with higher dropout probability, and still rises with repetition rate $R$. 
    (b) Expert specialization decreases with higher dropout probability, but still increases slightly with $R$. 
    (c) Co-activation entropy of expert pairs (normalized by maximum) increases with dropout probability, and still rises steadily with $R$ towards uniformly distributed pairings. }
    \label{fig:routing_extra_200m_dropout}
\end{figure*}

\clearpage
\subsection{Additional Language Modeling Tasks}
\label{app:extra_lm}
We show results for the additional held-out language modeling tasks of Appendix~\ref{app:eval_data_sources} on the models from \S\ref{sec:expts}, with the extended settings of Appendix~\ref{app:extended}. 
\begin{figure*}[!ht]
    \centering
    \begin{subfigure}[t]{\textwidth}
        \begin{subfigure}[t]{0.33\textwidth}
            \centering
            \includegraphics[width=\linewidth]{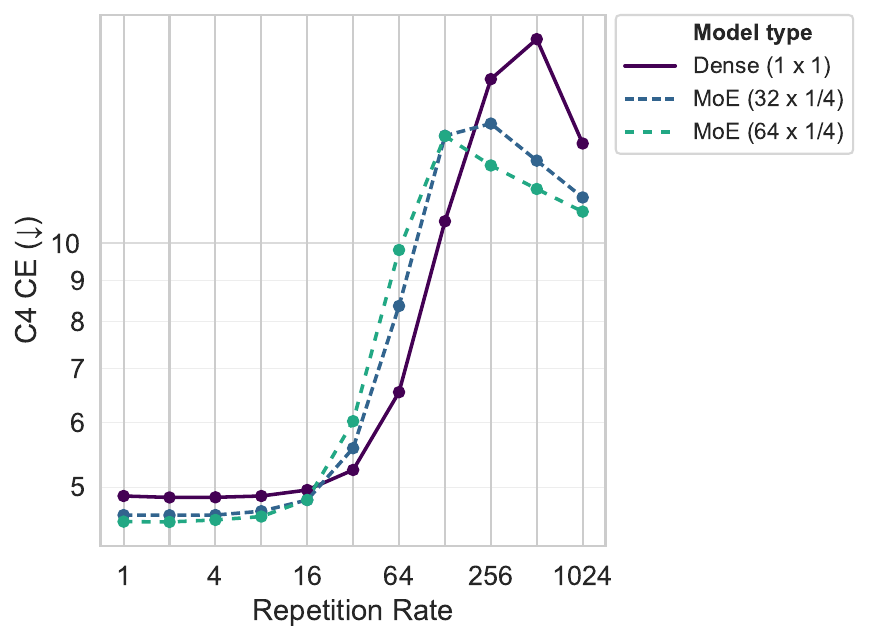}
        \end{subfigure}
        \begin{subfigure}[t]{0.33\textwidth}
            \centering
            \includegraphics[width=\linewidth]{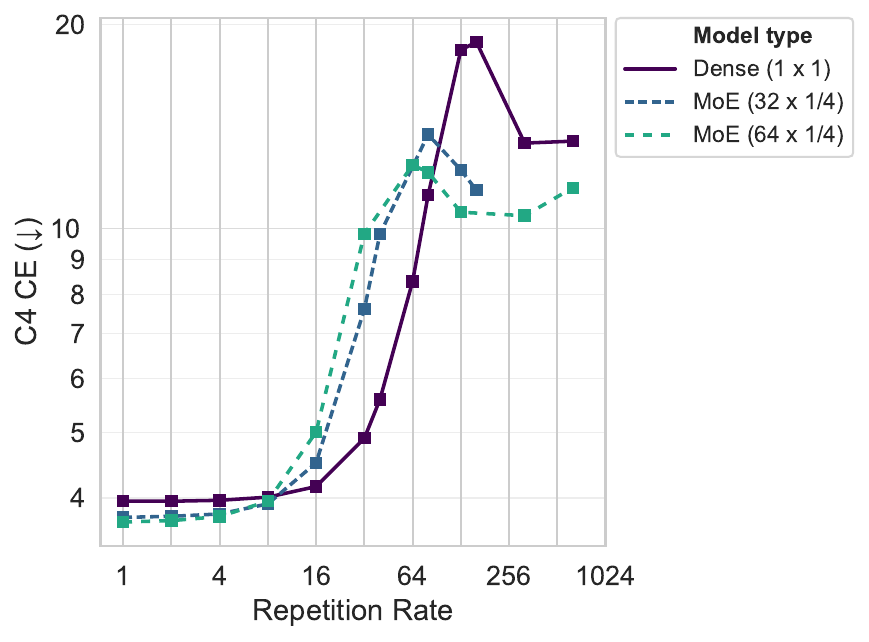}
        \end{subfigure}
        \begin{subfigure}[t]{0.33\textwidth}
            \centering
            \includegraphics[width=\linewidth]{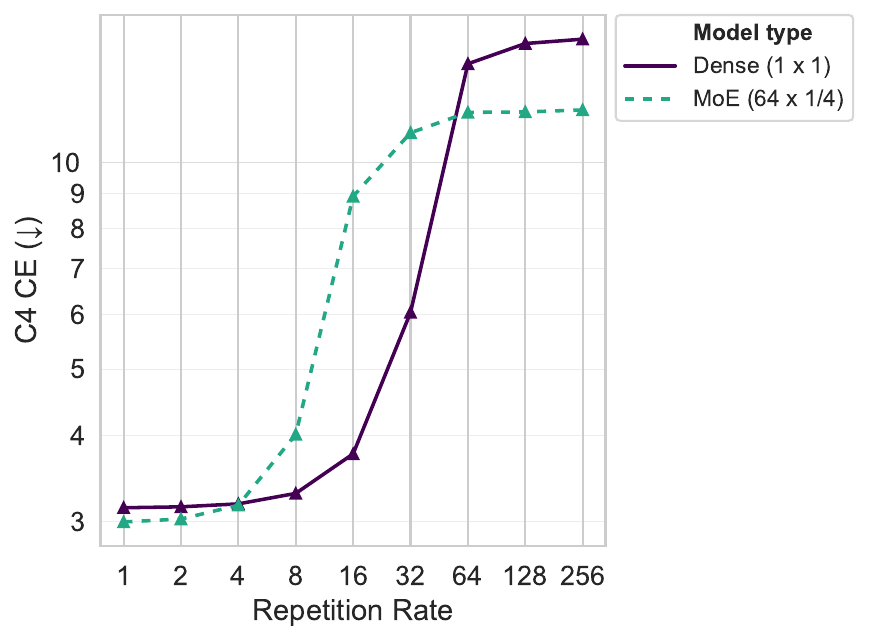}
        \end{subfigure}
    \subcaption{C4 (CE Loss $\downarrow$)}
    \end{subfigure}
    \par\vspace{1em}
    \begin{subfigure}[t]{\textwidth}
        \begin{subfigure}[t]{0.33\textwidth}
            \centering
            \includegraphics[width=\linewidth]{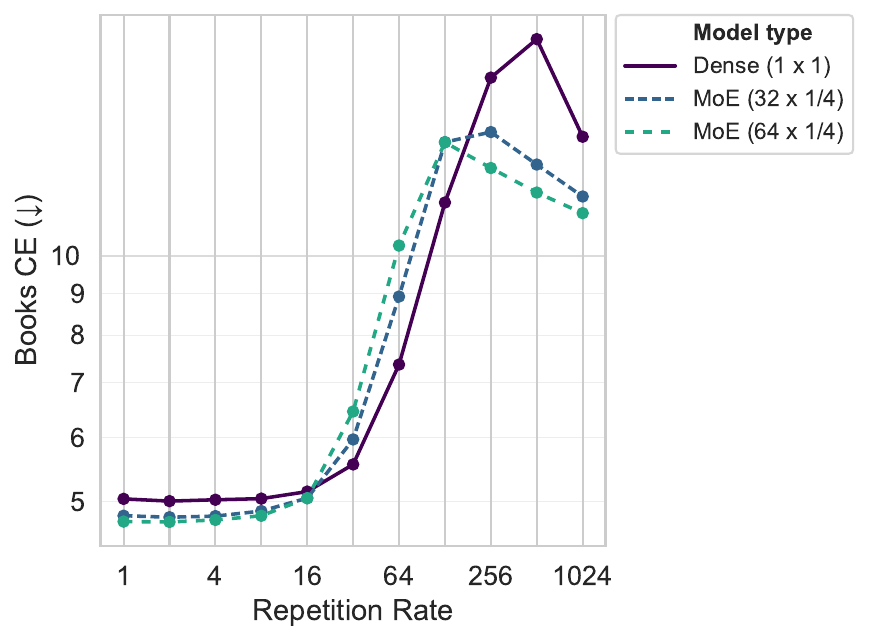}
        \end{subfigure}
        \begin{subfigure}[t]{0.33\textwidth}
            \centering
            \includegraphics[width=\linewidth]{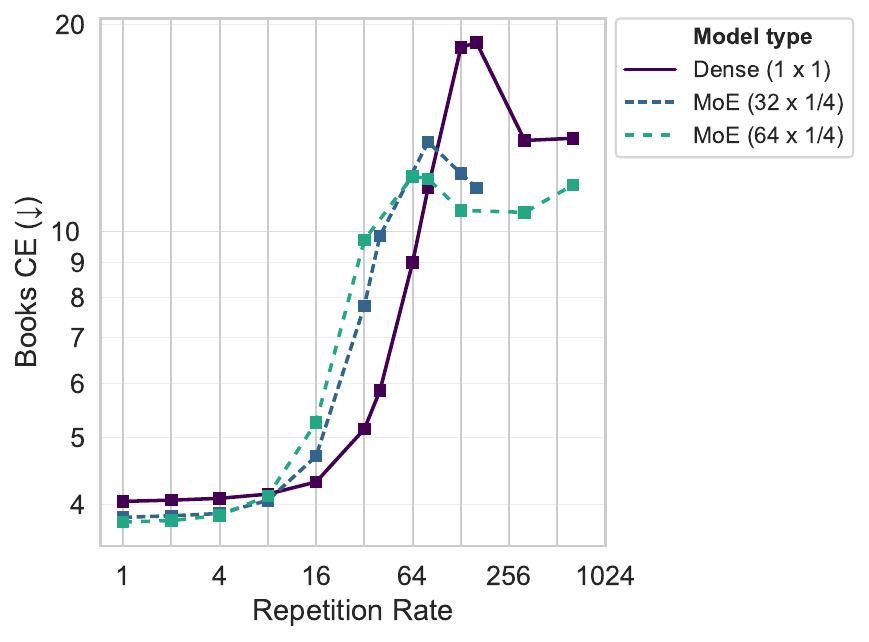}
        \end{subfigure}
        \begin{subfigure}[t]{0.33\textwidth}
            \centering
            \includegraphics[width=\linewidth]{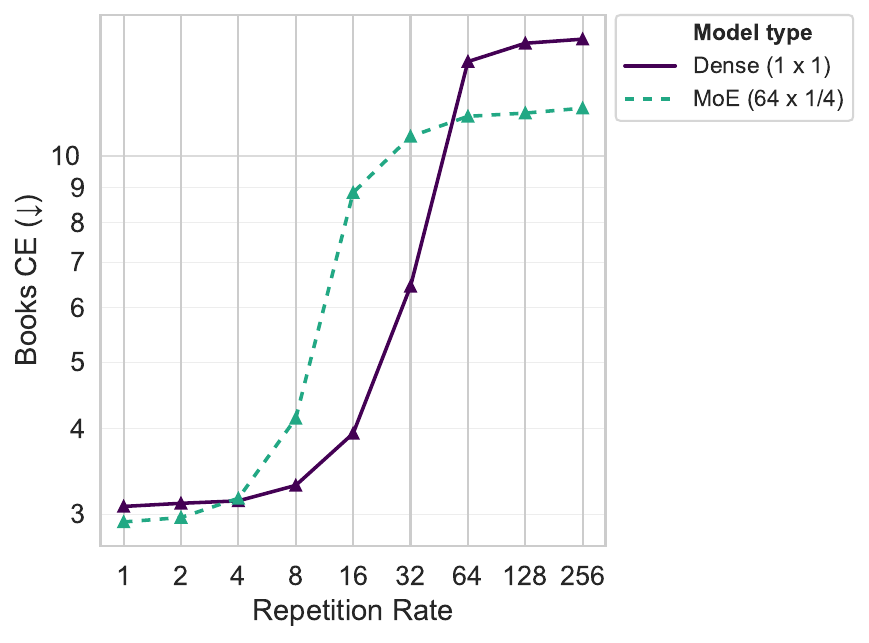}
        \end{subfigure}
    \subcaption{Dolma books (CE Loss $\downarrow$)}
    \end{subfigure}
    \par\vspace{1em}
    \begin{subfigure}[t]{\textwidth}
        \begin{subfigure}[t]{0.33\textwidth}
            \centering
            \includegraphics[width=\linewidth]{fig_pdfs/olmoe_mix/olmo_mix__80M__dense_moe32_moe64__eval_dolma_cc_loss.pdf}
        \end{subfigure}
        \begin{subfigure}[t]{0.33\textwidth}
            \centering
            \includegraphics[width=\linewidth]{fig_pdfs/olmoe_mix/olmo_mix__200M__dense_moe32_moe64__eval_dolma_cc_loss.pdf}
        \end{subfigure}
        \begin{subfigure}[t]{0.33\textwidth}
            \centering
            \includegraphics[width=\linewidth]{fig_pdfs/olmoe_mix/olmo_mix__1B__dense_moe32_moe64__eval_dolma_cc_loss.pdf}
        \end{subfigure}
    \subcaption{Dolma common-crawl (CE Loss $\downarrow$)}
    \end{subfigure}
    \par\vspace{1em}
    \begin{subfigure}[t]{\textwidth}
        \begin{subfigure}[t]{0.33\textwidth}
            \centering
            \includegraphics[width=\linewidth]{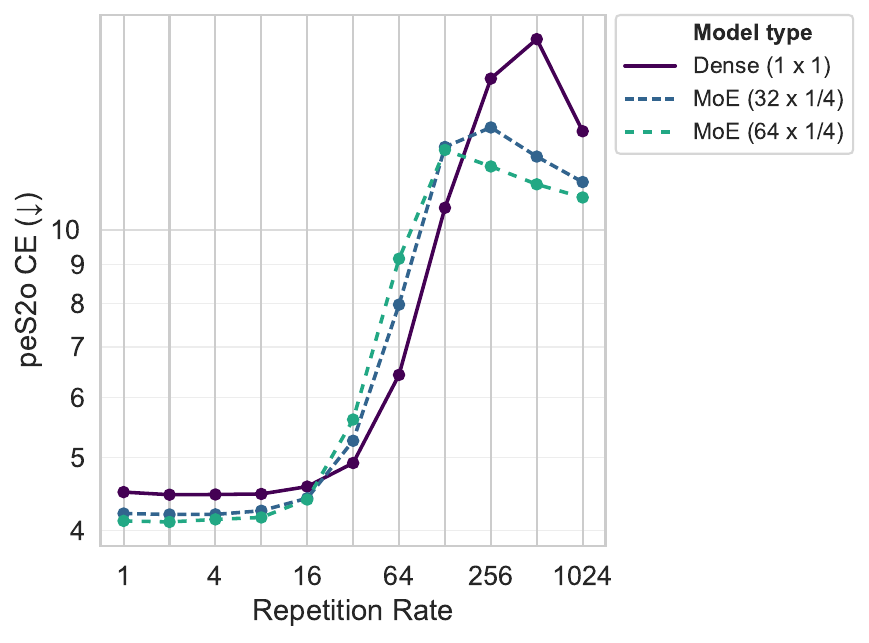}
        \end{subfigure}
        \begin{subfigure}[t]{0.33\textwidth}
            \centering
            \includegraphics[width=\linewidth]{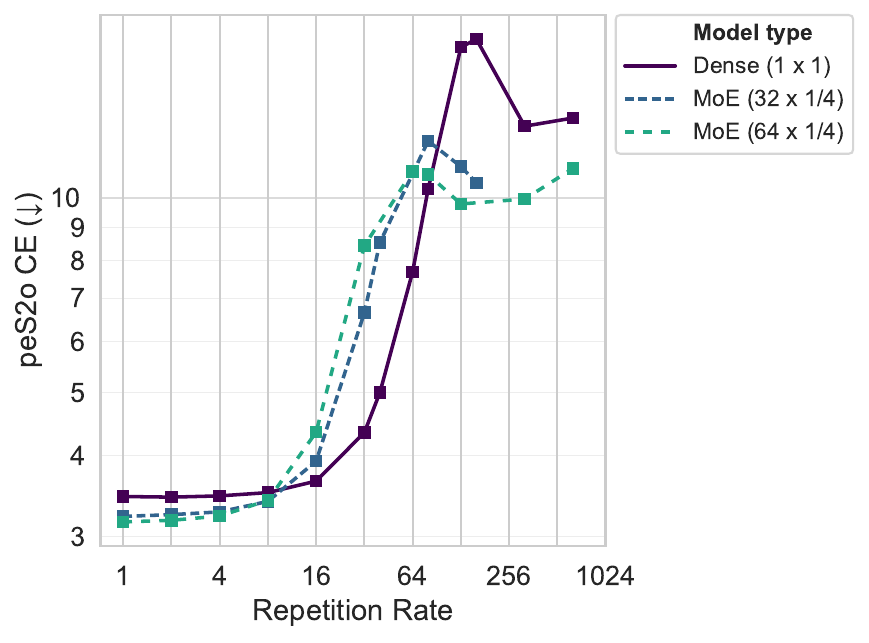}
        \end{subfigure}
        \begin{subfigure}[t]{0.33\textwidth}
            \centering
            \includegraphics[width=\linewidth]{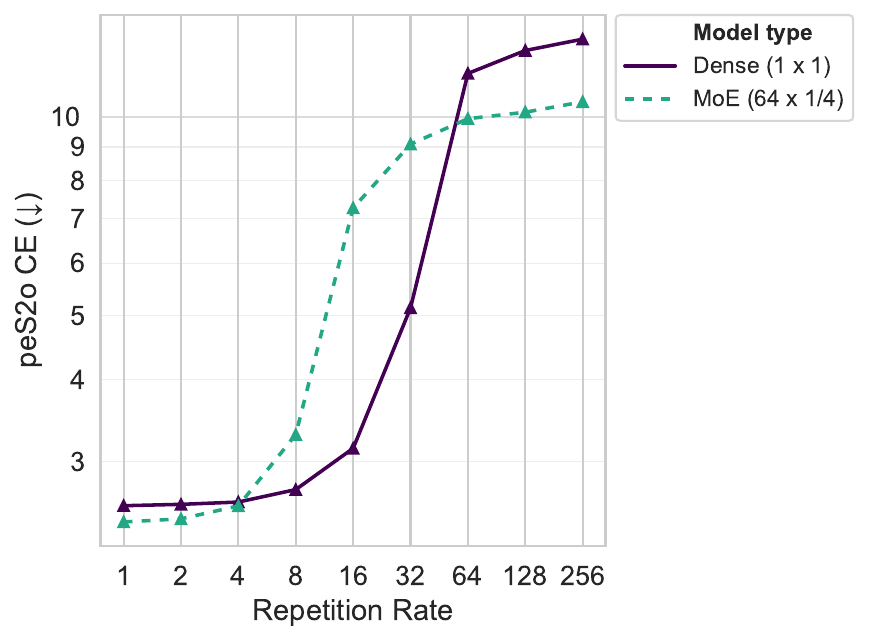}
        \end{subfigure}
    \subcaption{Dolma pes2o (CE Loss $\downarrow$)}
    \end{subfigure}
\end{figure*}

\begin{figure*}[!ht]
    \centering
    \ContinuedFloat
    \begin{subfigure}[t]{\textwidth}
        \begin{subfigure}[t]{0.33\textwidth}
            \centering
            \includegraphics[width=\linewidth]{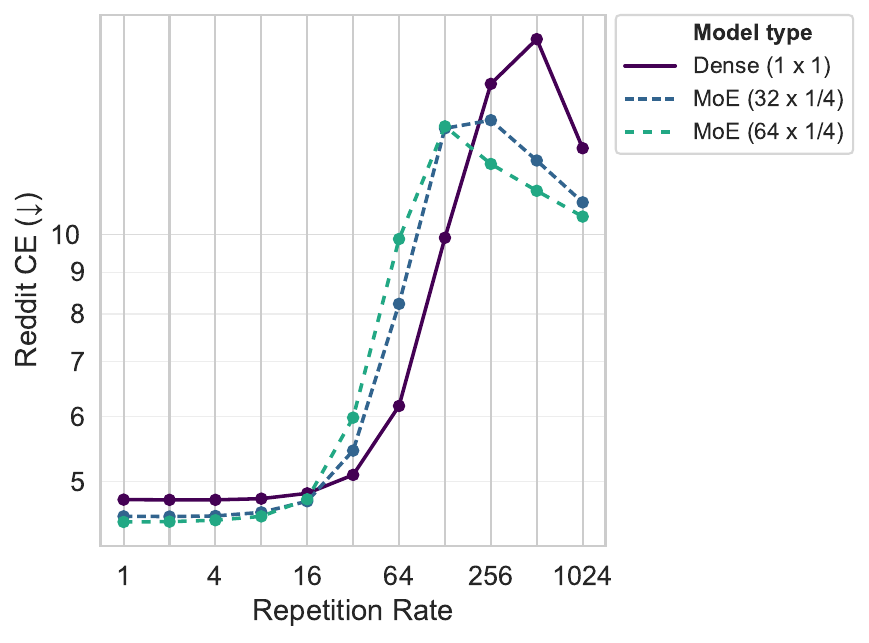}
        \end{subfigure}
        \begin{subfigure}[t]{0.33\textwidth}
            \centering
            \includegraphics[width=\linewidth]{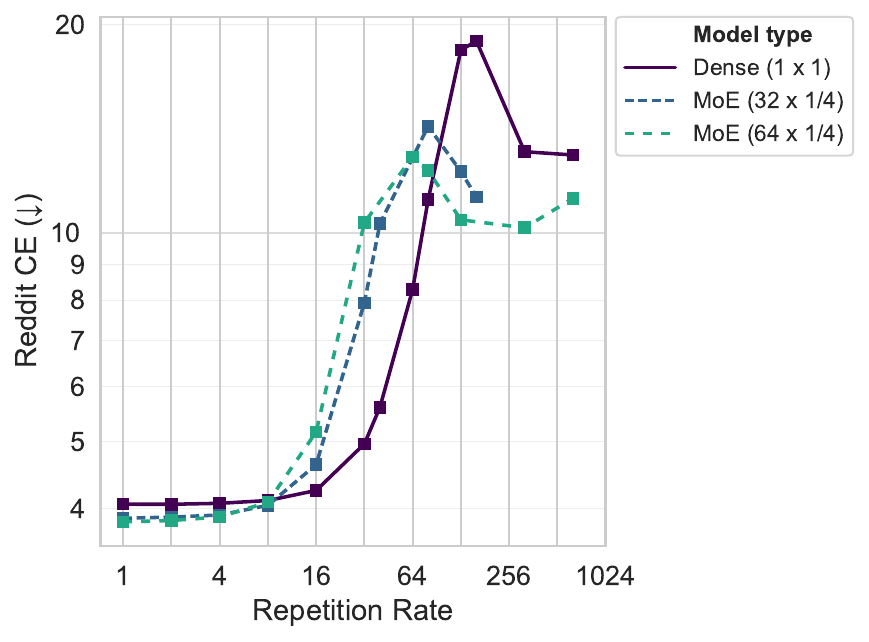}
        \end{subfigure}
        \begin{subfigure}[t]{0.33\textwidth}
            \centering
            \includegraphics[width=\linewidth]{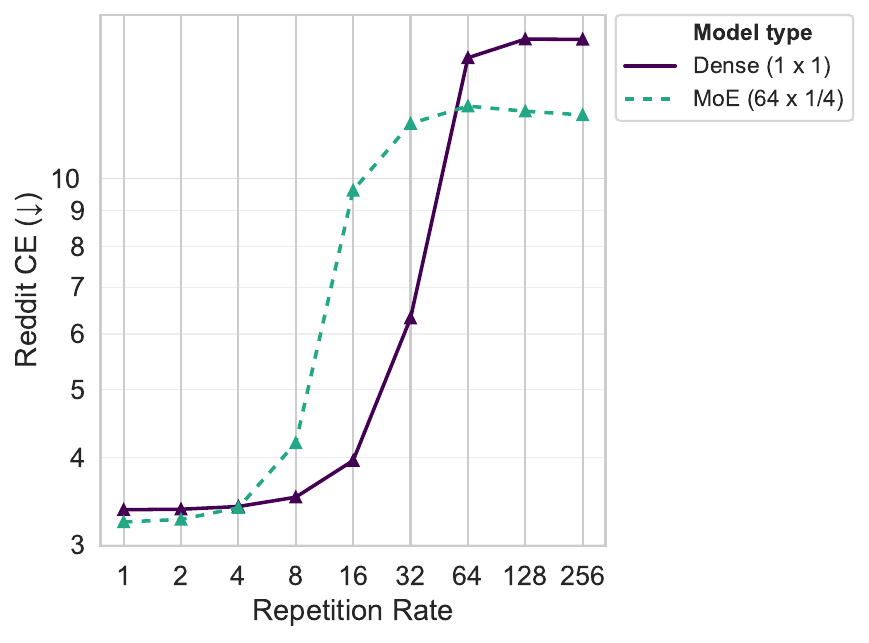}
        \end{subfigure}
    \subcaption{Dolma reddit (CE Loss $\downarrow$)}
    \end{subfigure}
    \par\vspace{1em}
    \begin{subfigure}[t]{\textwidth}
        \begin{subfigure}[t]{0.33\textwidth}
            \centering
            \includegraphics[width=\linewidth]{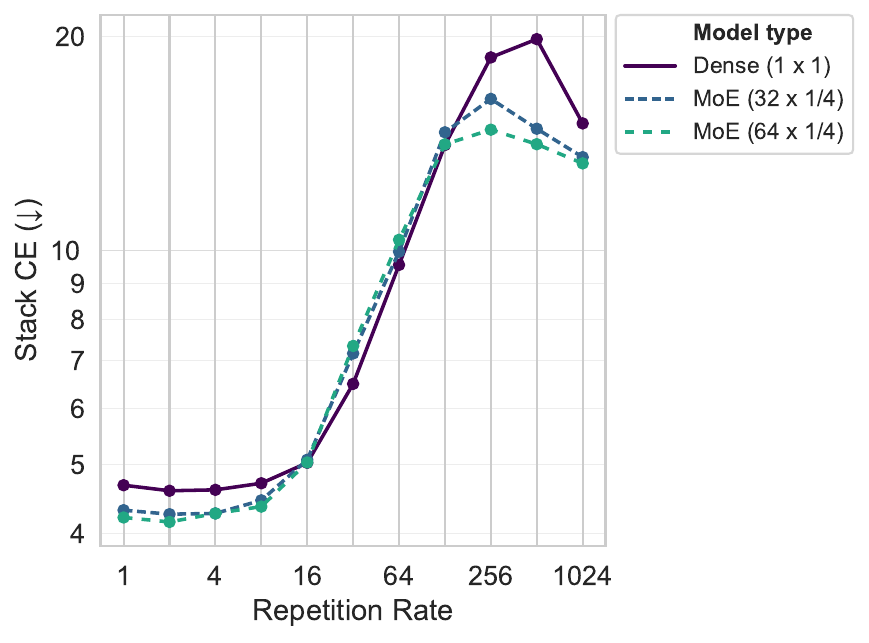}
        \end{subfigure}
        \begin{subfigure}[t]{0.33\textwidth}
            \centering
            \includegraphics[width=\linewidth]{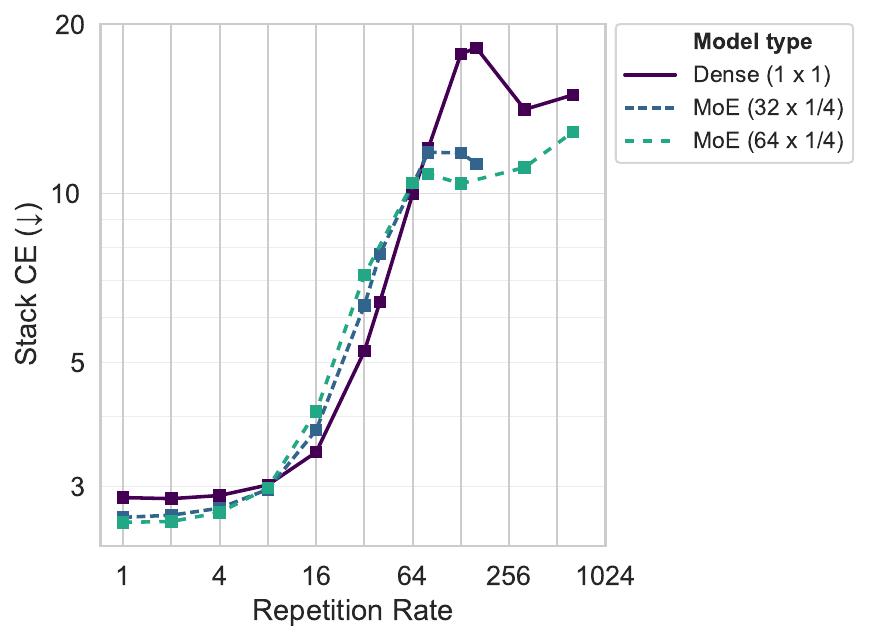}
        \end{subfigure}
        \begin{subfigure}[t]{0.33\textwidth}
            \centering
            \includegraphics[width=\linewidth]{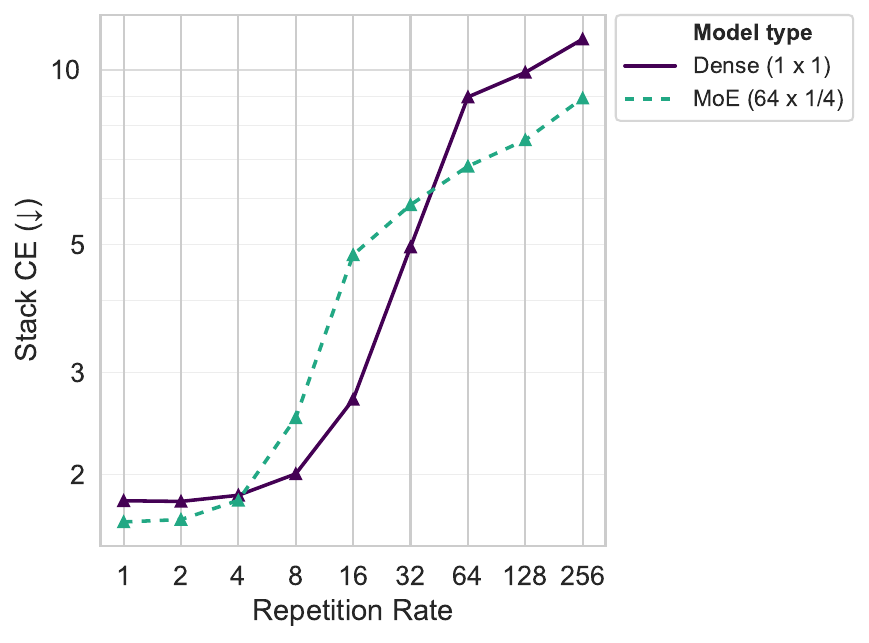}
        \end{subfigure}
    \subcaption{Dolma stack (CE Loss $\downarrow$)}
    \end{subfigure}
    \par\vspace{1em}
    \begin{subfigure}[t]{\textwidth}
        \begin{subfigure}[t]{0.33\textwidth}
            \centering
            \includegraphics[width=\linewidth]{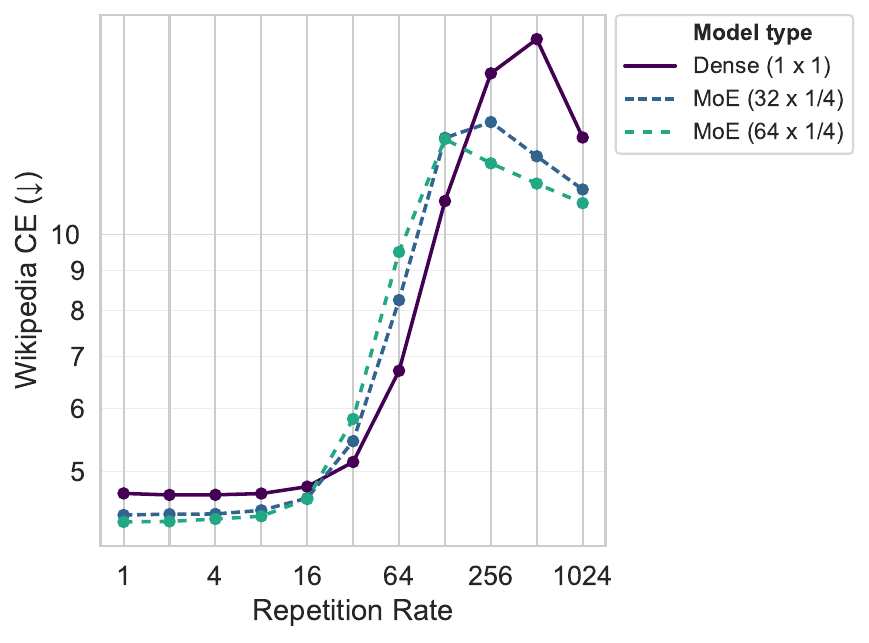}
        \end{subfigure}
        \begin{subfigure}[t]{0.33\textwidth}
            \centering
            \includegraphics[width=\linewidth]{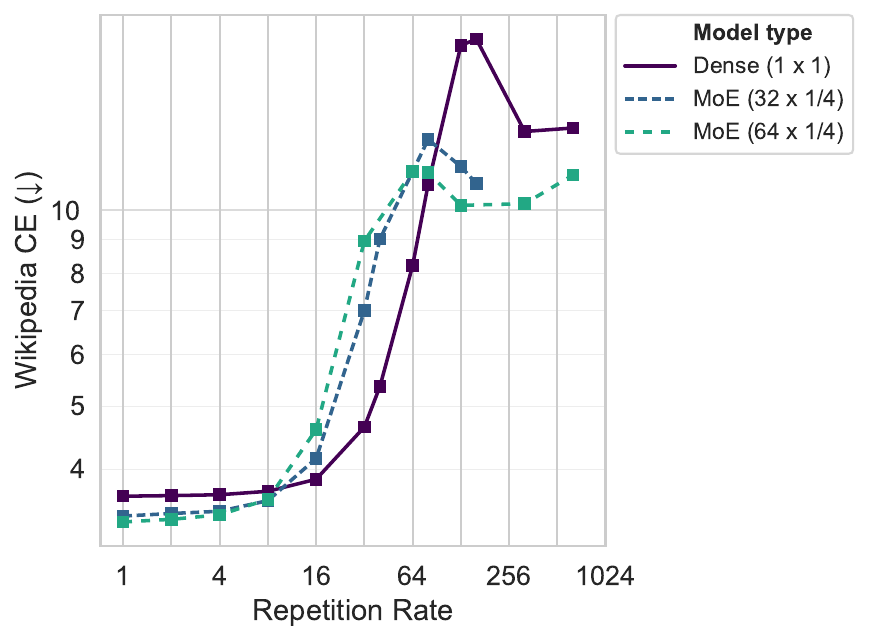}
        \end{subfigure}
        \begin{subfigure}[t]{0.33\textwidth}
            \centering
            \includegraphics[width=\linewidth]{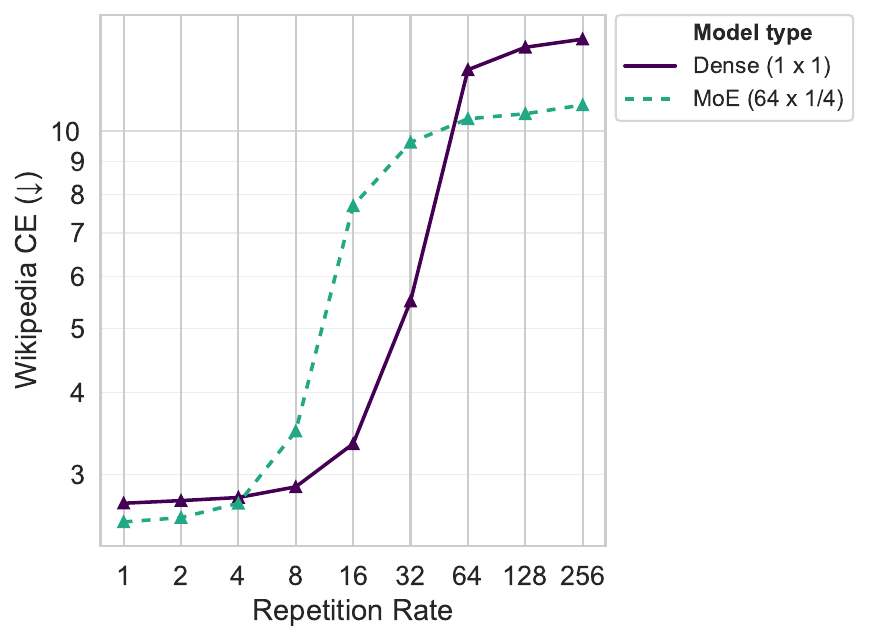}
        \end{subfigure}
    \subcaption{Dolma wiki (CE Loss $\downarrow$)}
    \end{subfigure}
    \par\vspace{1em}
    \begin{subfigure}[t]{\textwidth}
        \begin{subfigure}[t]{0.33\textwidth}
            \centering
            \includegraphics[width=\linewidth]{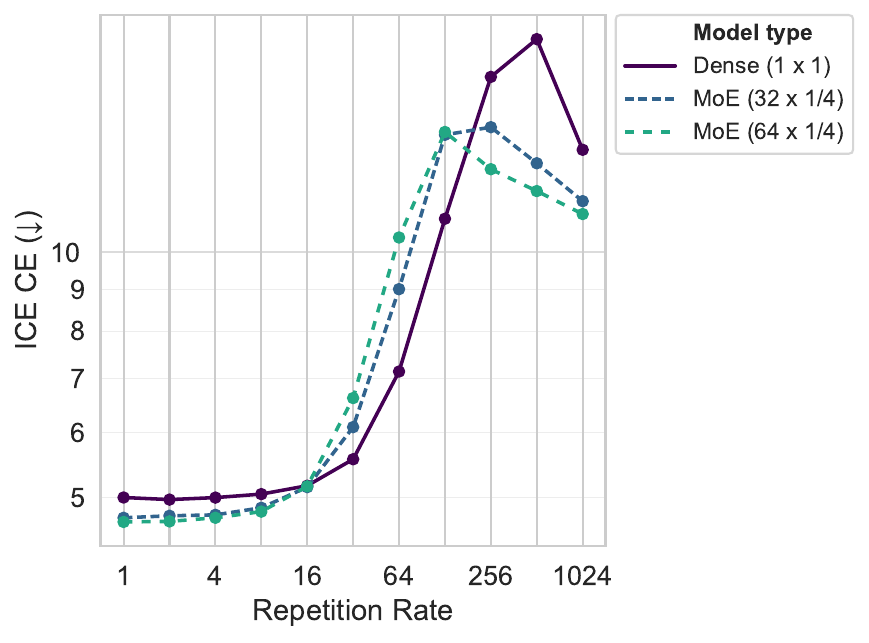}
        \end{subfigure}
        \begin{subfigure}[t]{0.33\textwidth}
            \centering
            \includegraphics[width=\linewidth]{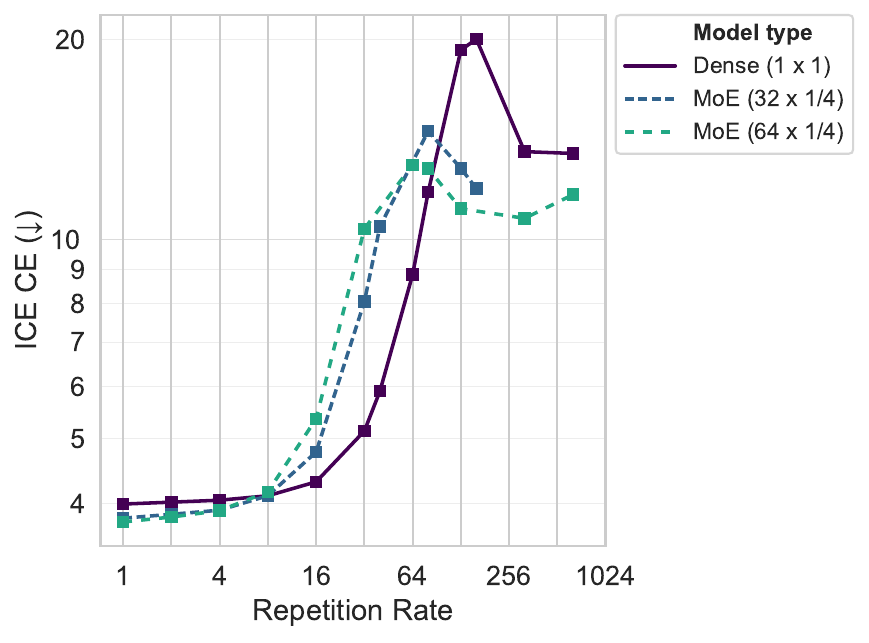}
        \end{subfigure}
        \begin{subfigure}[t]{0.33\textwidth}
            \centering
            \includegraphics[width=\linewidth]{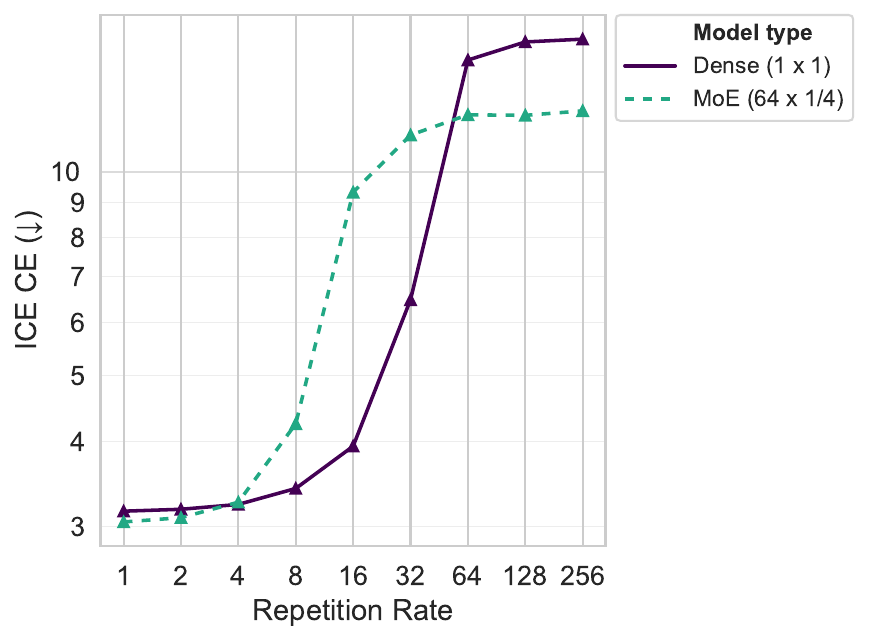}
        \end{subfigure}
    \subcaption{ICE (CE Loss $\downarrow$)}
    \end{subfigure}
\end{figure*}

\begin{figure*}[!ht]
    \centering
    \ContinuedFloat
    \begin{subfigure}[t]{\textwidth}
        \begin{subfigure}[t]{0.33\textwidth}
            \centering
            \includegraphics[width=\linewidth]{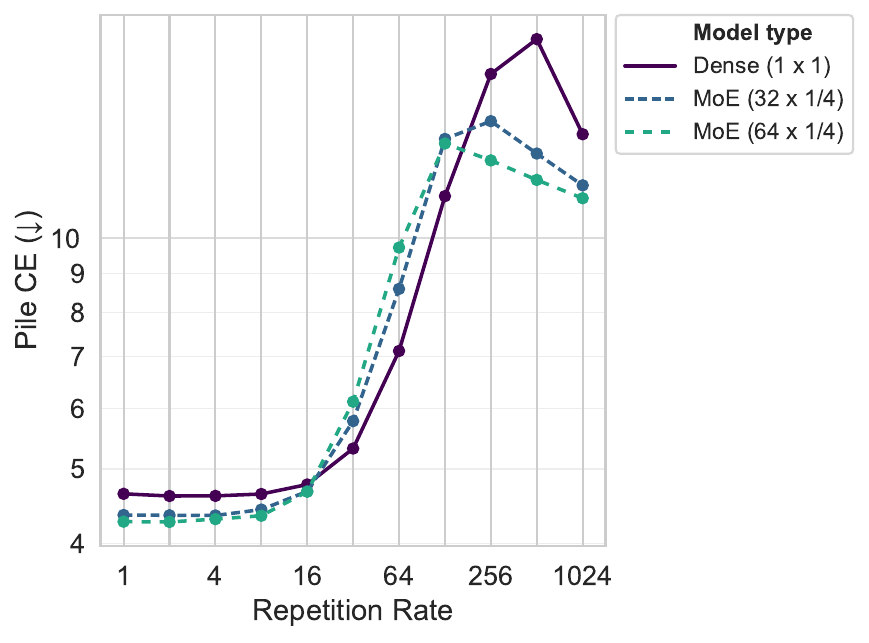}
        \end{subfigure}
        \begin{subfigure}[t]{0.33\textwidth}
            \centering
            \includegraphics[width=\linewidth]{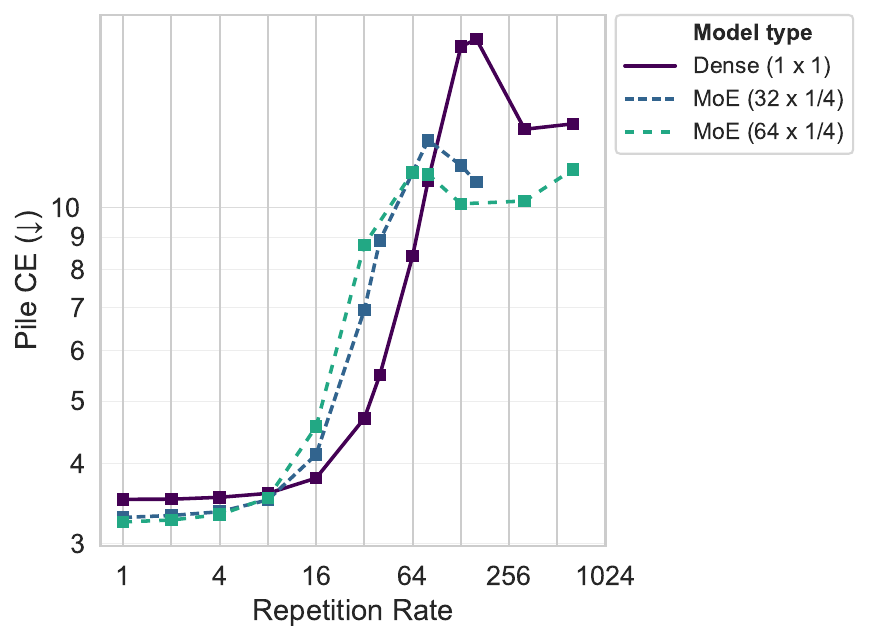}
        \end{subfigure}
        \begin{subfigure}[t]{0.33\textwidth}
            \centering
            \includegraphics[width=\linewidth]{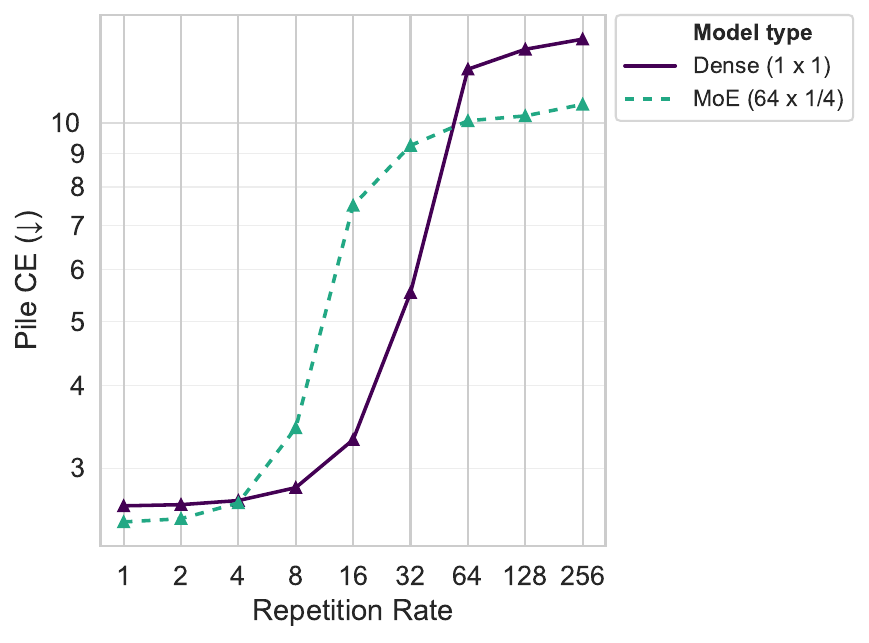}
        \end{subfigure}
    \subcaption{Pile (CE Loss $\downarrow$)}
    \end{subfigure}
    \par\vspace{1em}
    \begin{subfigure}[t]{\textwidth}
        \begin{subfigure}[t]{0.33\textwidth}
            \centering
            \includegraphics[width=\linewidth]{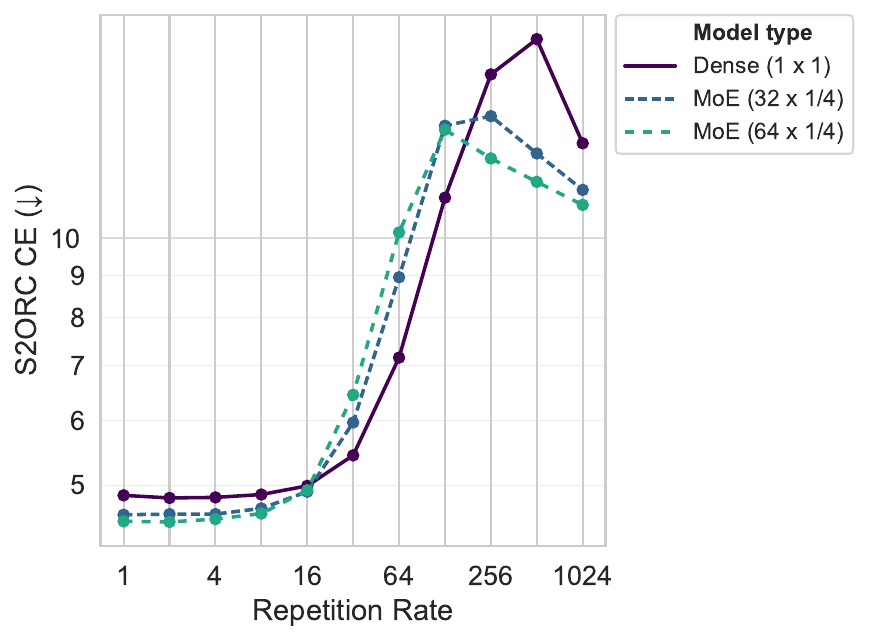}
        \end{subfigure}
        \begin{subfigure}[t]{0.33\textwidth}
            \centering
            \includegraphics[width=\linewidth]{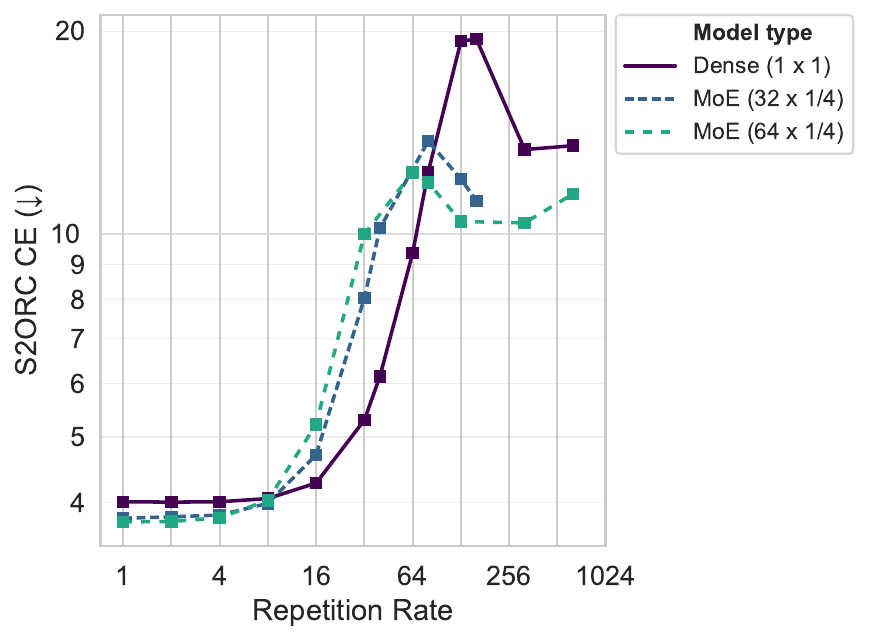}
        \end{subfigure}
        \begin{subfigure}[t]{0.33\textwidth}
            \centering
            \includegraphics[width=\linewidth]{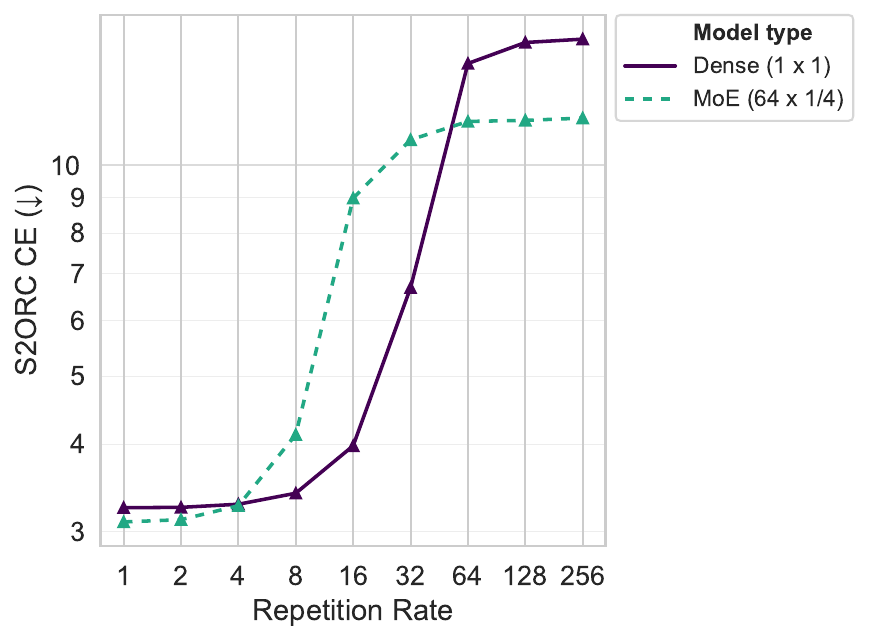}
        \end{subfigure}
    \subcaption{S2ORC (CE Loss $\downarrow$)}
    \end{subfigure}
    \par\vspace{1em}
    \begin{subfigure}[t]{\textwidth}
        \begin{subfigure}[t]{0.33\textwidth}
            \centering
            \includegraphics[width=\linewidth]{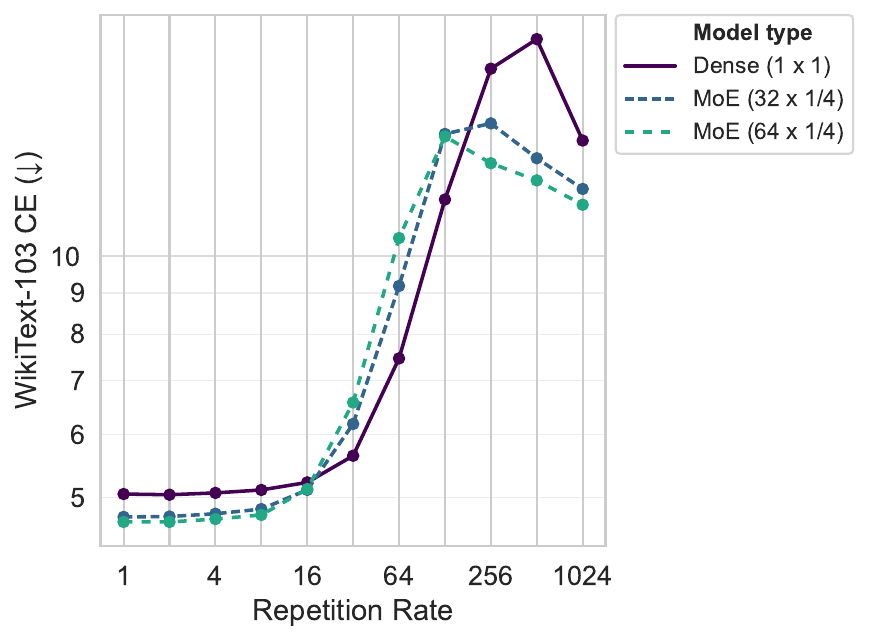}
        \end{subfigure}
        \begin{subfigure}[t]{0.33\textwidth}
            \centering
            \includegraphics[width=\linewidth]{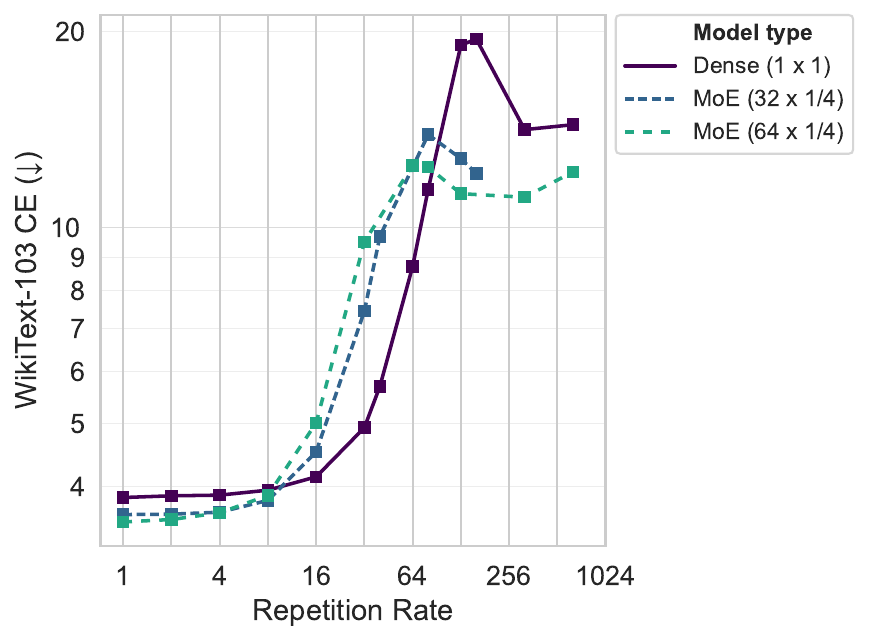}
        \end{subfigure}
        \begin{subfigure}[t]{0.33\textwidth}
            \centering
            \includegraphics[width=\linewidth]{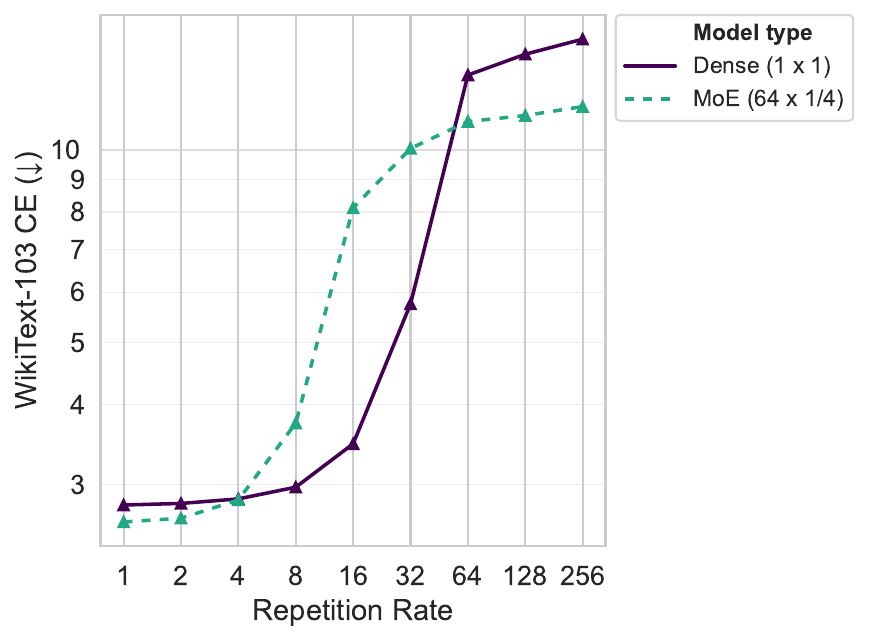}
        \end{subfigure}
    \subcaption{Wikitext (CE Loss $\downarrow$)}
    \end{subfigure}
    \caption{\textbf{Held-out language modeling loss largely follows the same patterns.} For 80M, 200M, and 1B active parameter models trained on OLMoE mix, we show held-out language modeling loss on a wide variety of domains. Performance depends on the exact domain, but largely follows the trends shown in Figure~\ref{fig:olmoe_mix}.}
    \label{fig:lm_app}
\end{figure*}

\clearpage
\subsection{Downstream Tasks}
\label{app:downstream_tasks}
We show results for the downstream tasks of Appendix~\ref{app:eval_data_sources} on the models from \S\ref{sec:expts}, with the extended settings of Appendix~\ref{app:extended}. We include CE Loss on all tasks, as well as accuracy on Hellaswag. Accuracy on all other tasks remains near-chance, even at 1B scale.

\begin{figure*}[!ht]
    \centering
    \begin{subfigure}[t]{\textwidth}
        \begin{subfigure}[t]{0.33\textwidth}
            \centering
            \includegraphics[width=\linewidth]{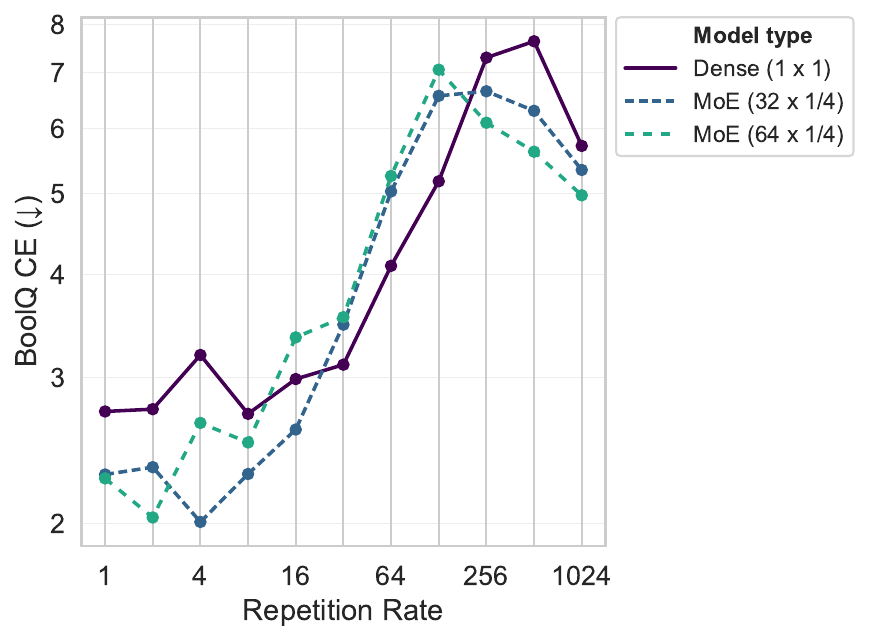}
        \end{subfigure}
        \begin{subfigure}[t]{0.33\textwidth}
            \centering
            \includegraphics[width=\linewidth]{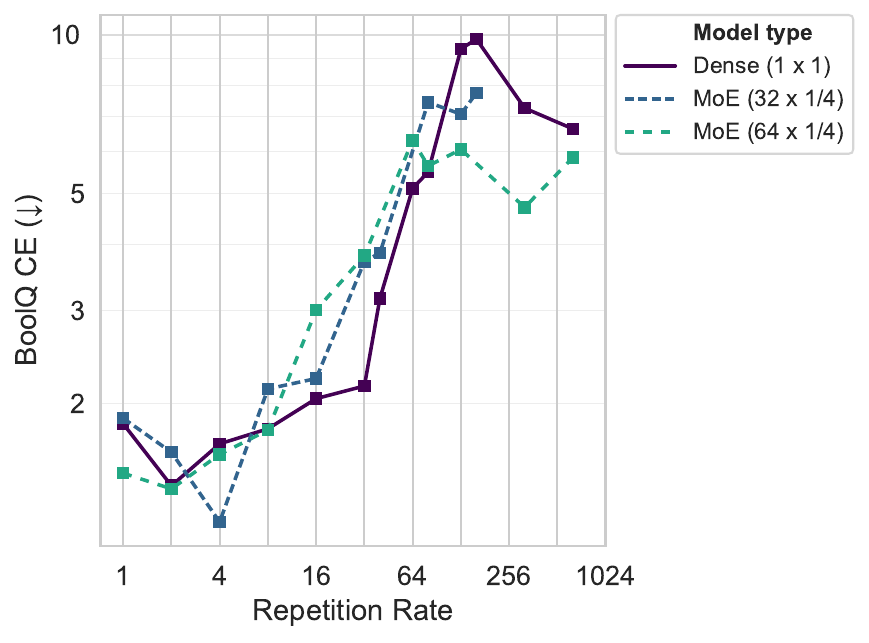}
        \end{subfigure}
        \begin{subfigure}[t]{0.33\textwidth}
            \centering
            \includegraphics[width=\linewidth]{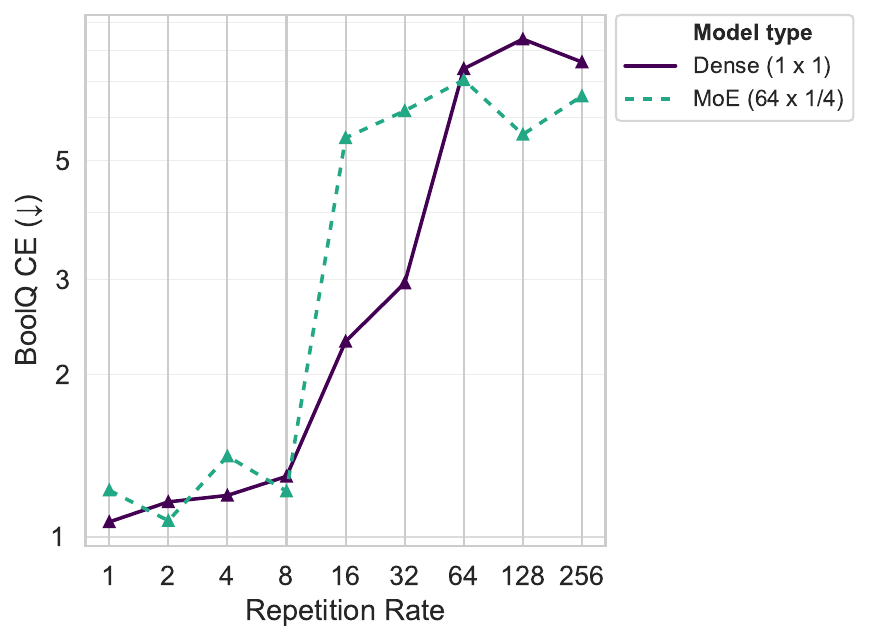}
        \end{subfigure}
    \subcaption{BoolQ (CE Loss $\downarrow$)}
    \end{subfigure}
    \par\vspace{1em}
    \begin{subfigure}[t]{\textwidth}
        \begin{subfigure}[t]{0.33\textwidth}
            \centering
            \includegraphics[width=\linewidth]{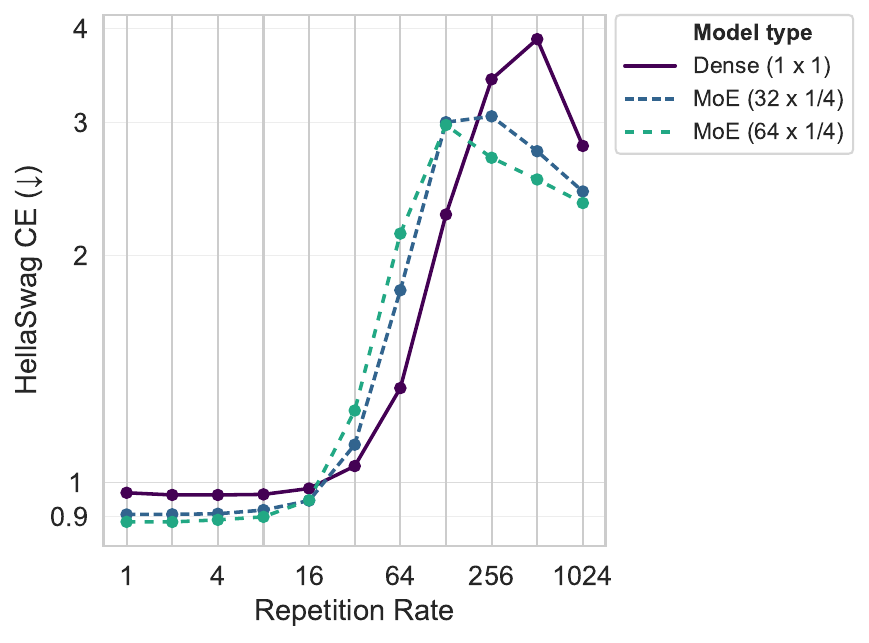}
        \end{subfigure}
        \begin{subfigure}[t]{0.33\textwidth}
            \centering
            \includegraphics[width=\linewidth]{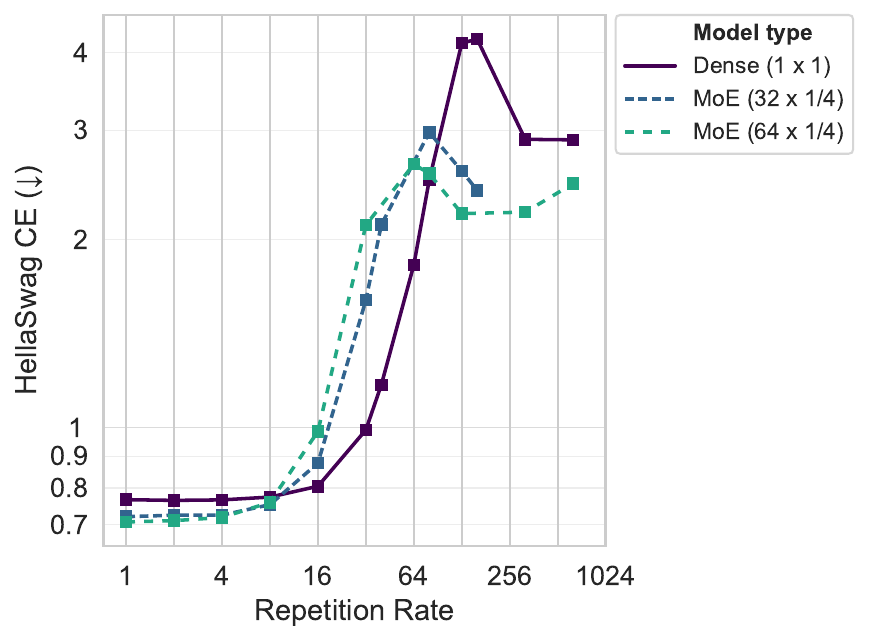}
        \end{subfigure}
        \begin{subfigure}[t]{0.33\textwidth}
            \centering
            \includegraphics[width=\linewidth]{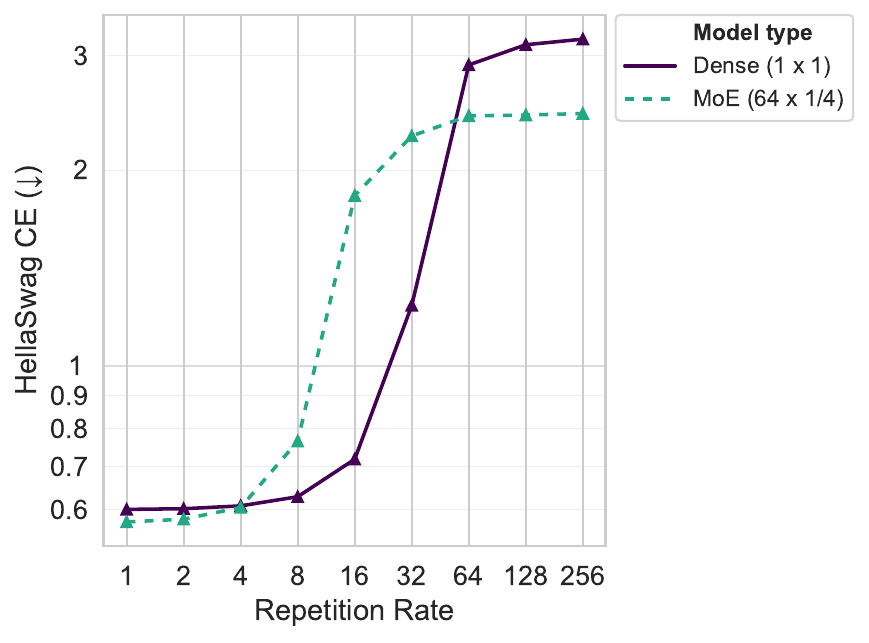}
        \end{subfigure}
    \subcaption{Hellaswag (CE Loss $\downarrow$)}
    \end{subfigure}
    \par\vspace{1em}
    \begin{subfigure}[t]{\textwidth}
        \begin{subfigure}[t]{0.33\textwidth}
            \centering
            \includegraphics[width=\linewidth]{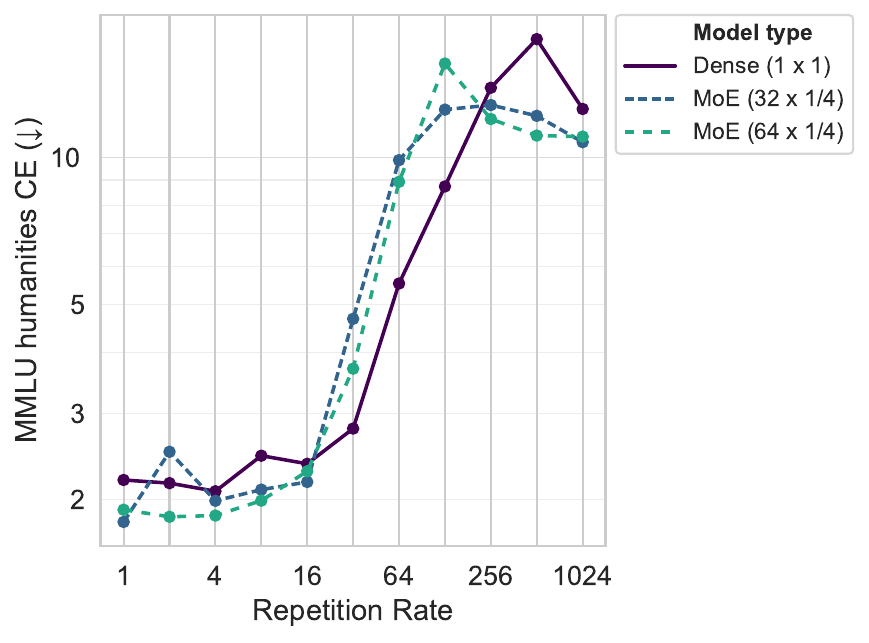}
        \end{subfigure}
        \begin{subfigure}[t]{0.33\textwidth}
            \centering
            \includegraphics[width=\linewidth]{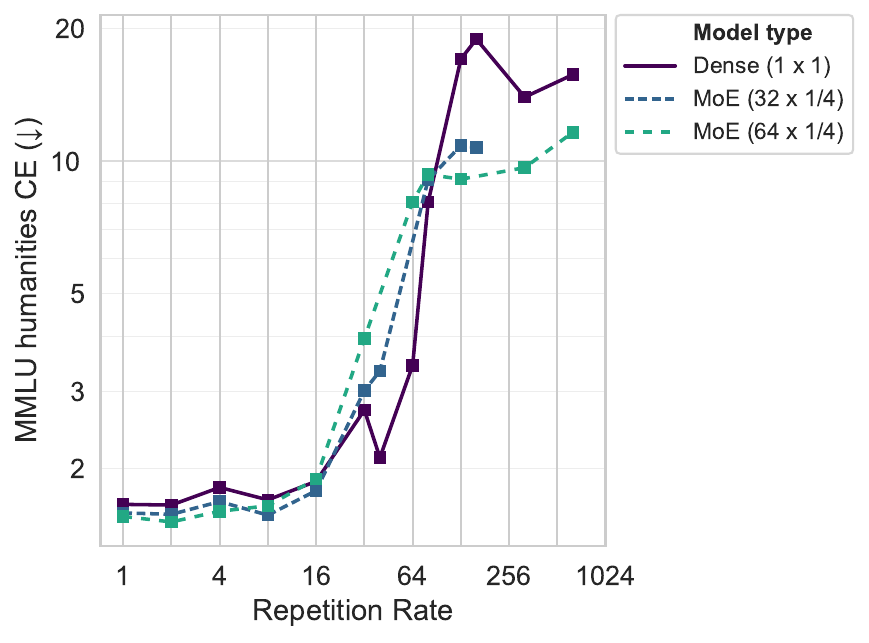}
        \end{subfigure}
        \begin{subfigure}[t]{0.33\textwidth}
            \centering
            \includegraphics[width=\linewidth]{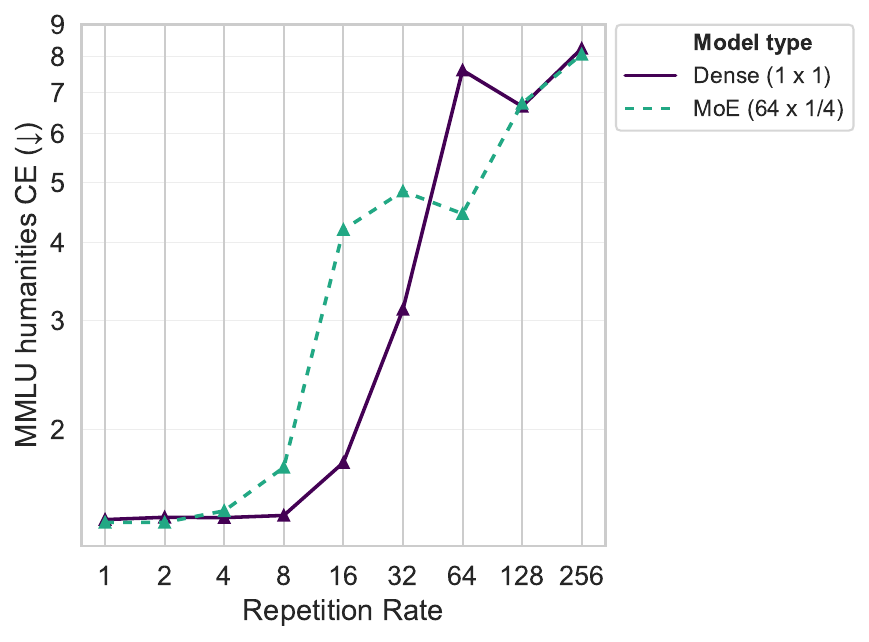}
        \end{subfigure}
    \subcaption{MMLU Humanities (CE Loss $\downarrow$)}
    \end{subfigure}
    \par\vspace{1em}
    \begin{subfigure}[t]{\textwidth}
        \begin{subfigure}[t]{0.33\textwidth}
            \centering
            \includegraphics[width=\linewidth]{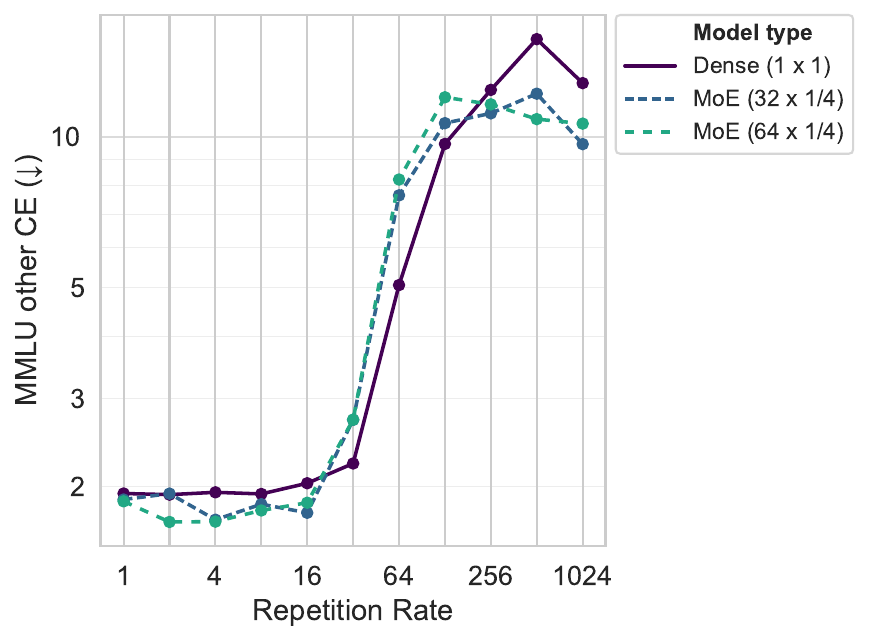}
        \end{subfigure}
        \begin{subfigure}[t]{0.33\textwidth}
            \centering
            \includegraphics[width=\linewidth]{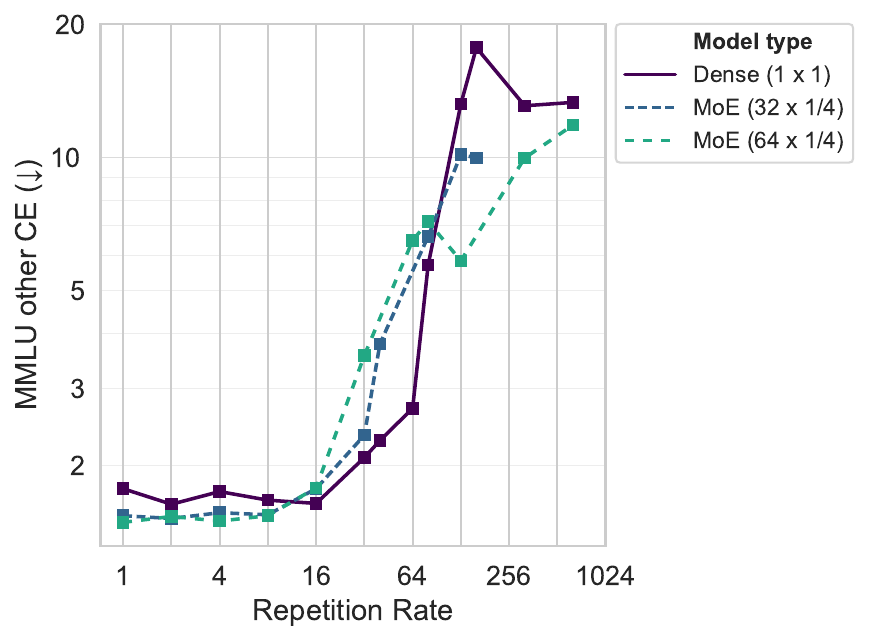}
        \end{subfigure}
        \begin{subfigure}[t]{0.33\textwidth}
            \centering
            \includegraphics[width=\linewidth]{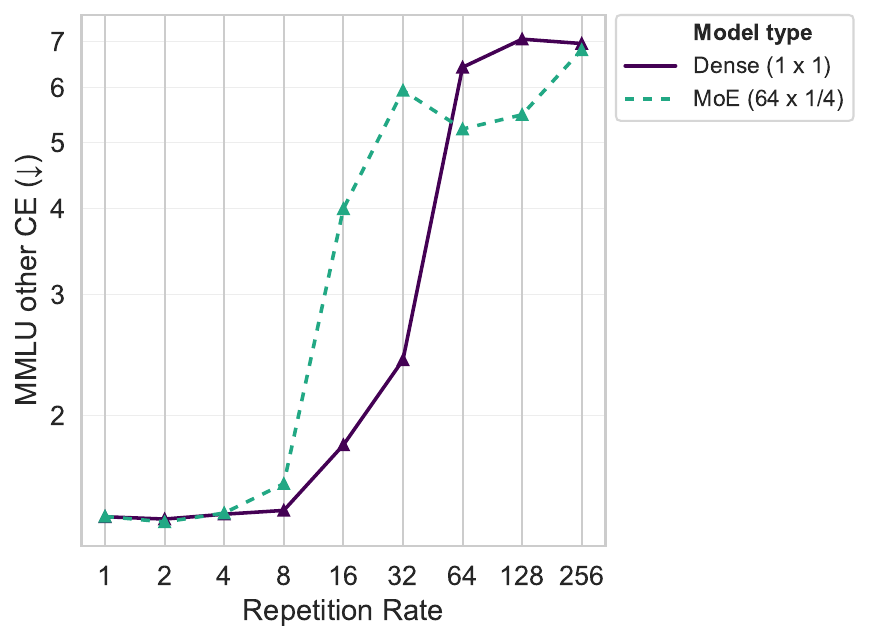}
        \end{subfigure}
    \subcaption{MMLU Other (CE Loss $\downarrow$)}
    \end{subfigure}
\end{figure*}

\begin{figure*}[!ht]
    \centering
    \ContinuedFloat
    \begin{subfigure}[t]{\textwidth}
        \begin{subfigure}[t]{0.33\textwidth}
            \centering
            \includegraphics[width=\linewidth]{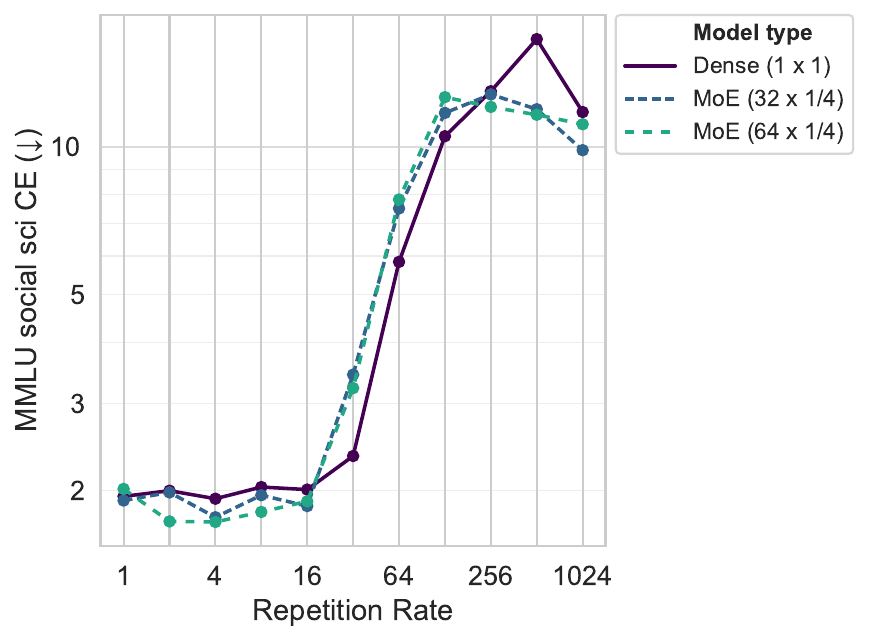}
        \end{subfigure}
        \begin{subfigure}[t]{0.33\textwidth}
            \centering
            \includegraphics[width=\linewidth]{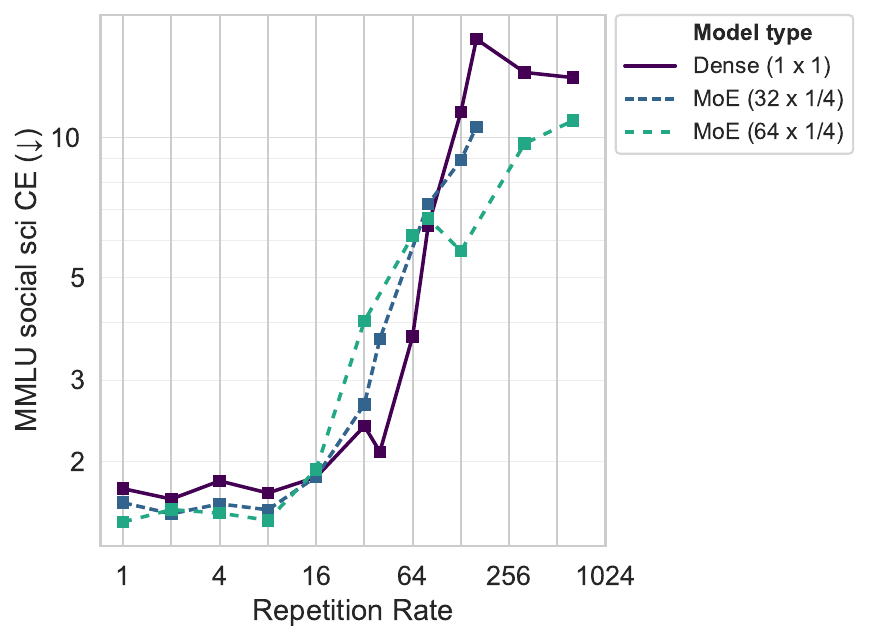}
        \end{subfigure}
        \begin{subfigure}[t]{0.33\textwidth}
            \centering
            \includegraphics[width=\linewidth]{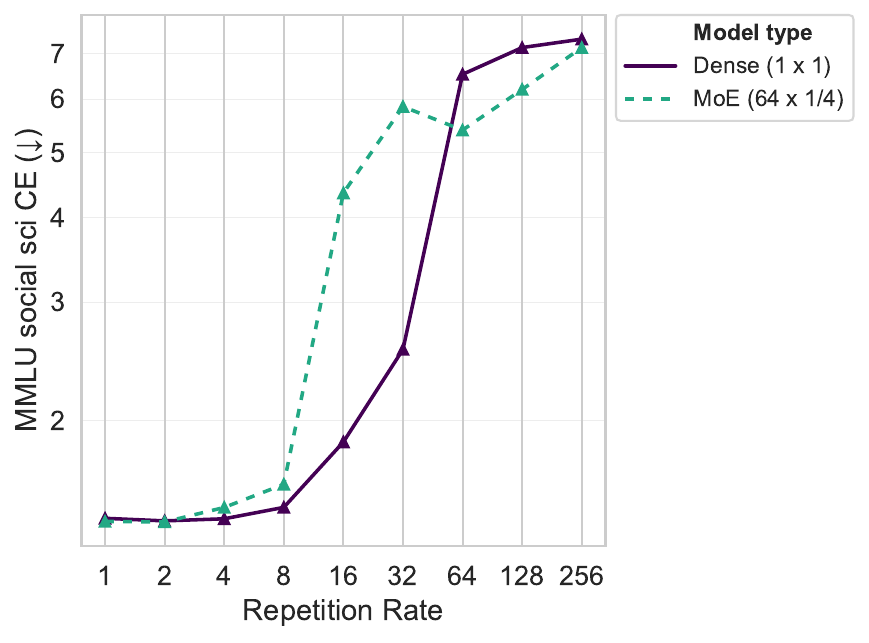}
        \end{subfigure}
    \subcaption{MMLU Social Sciences (CE Loss $\downarrow$)}
    \end{subfigure}
    \par\vspace{1em}
    \begin{subfigure}[t]{\textwidth}
        \begin{subfigure}[t]{0.33\textwidth}
            \centering
            \includegraphics[width=\linewidth]{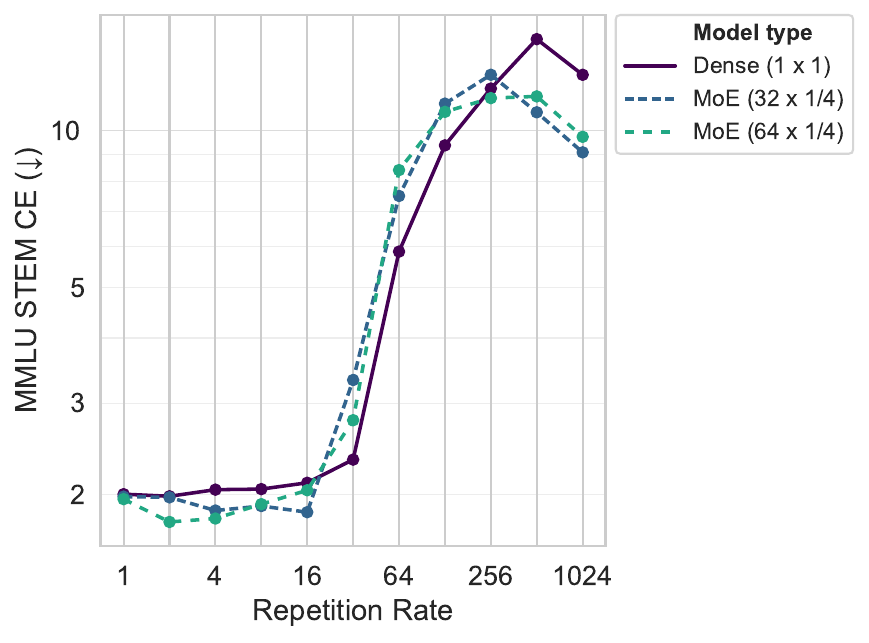}
        \end{subfigure}
        \begin{subfigure}[t]{0.33\textwidth}
            \centering
            \includegraphics[width=\linewidth]{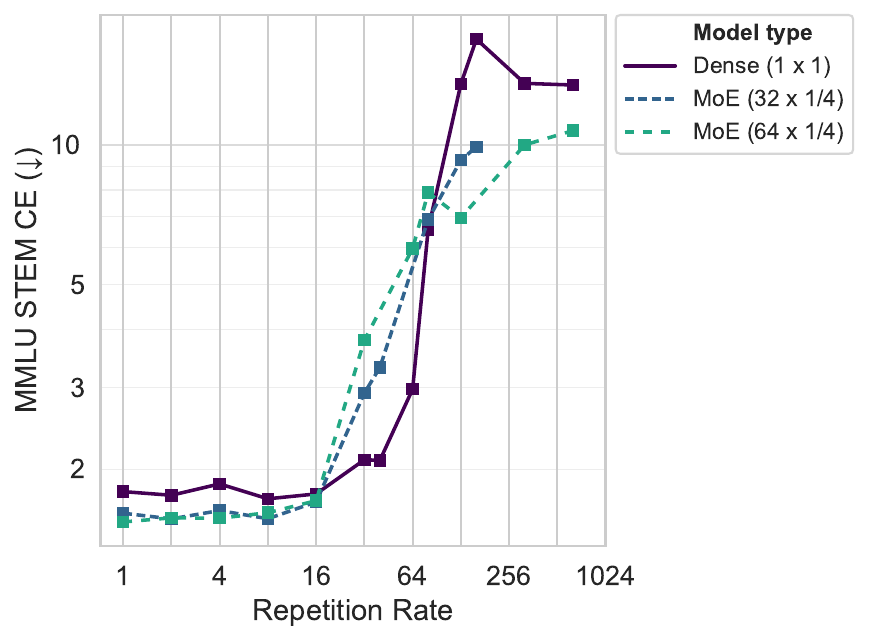}
        \end{subfigure}
        \begin{subfigure}[t]{0.33\textwidth}
            \centering
            \includegraphics[width=\linewidth]{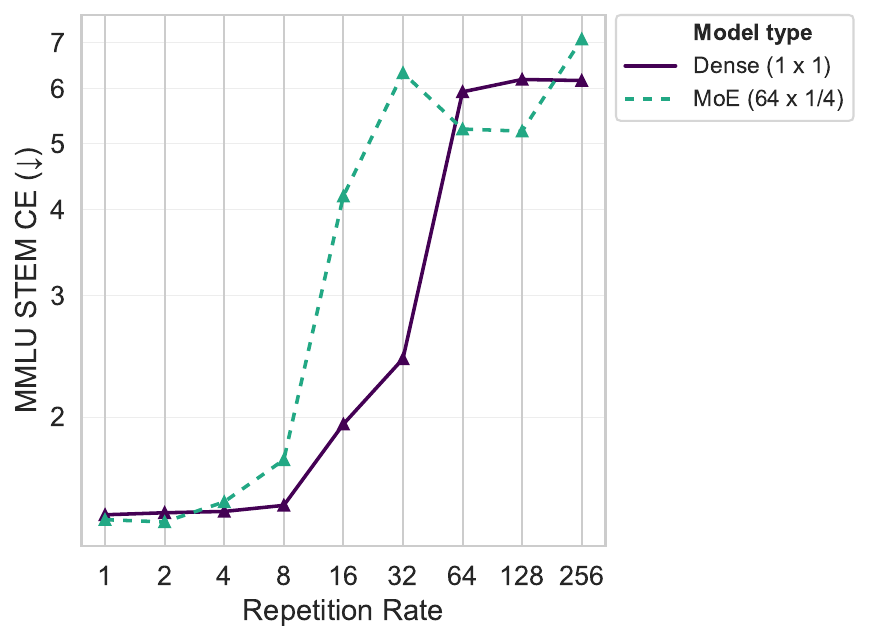}
        \end{subfigure}
    \subcaption{MMLU Stem (CE Loss $\downarrow$)}
    \end{subfigure}
    \par\vspace{1em}
    \begin{subfigure}[t]{\textwidth}
        \begin{subfigure}[t]{0.33\textwidth}
            \centering
            \includegraphics[width=\linewidth]{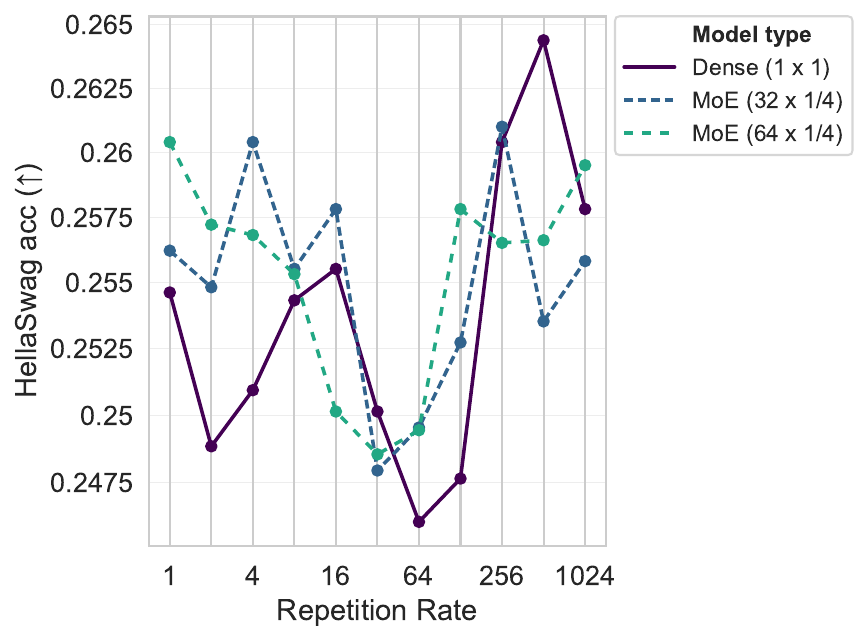}
        \end{subfigure}
        \begin{subfigure}[t]{0.33\textwidth}
            \centering
            \includegraphics[width=\linewidth]{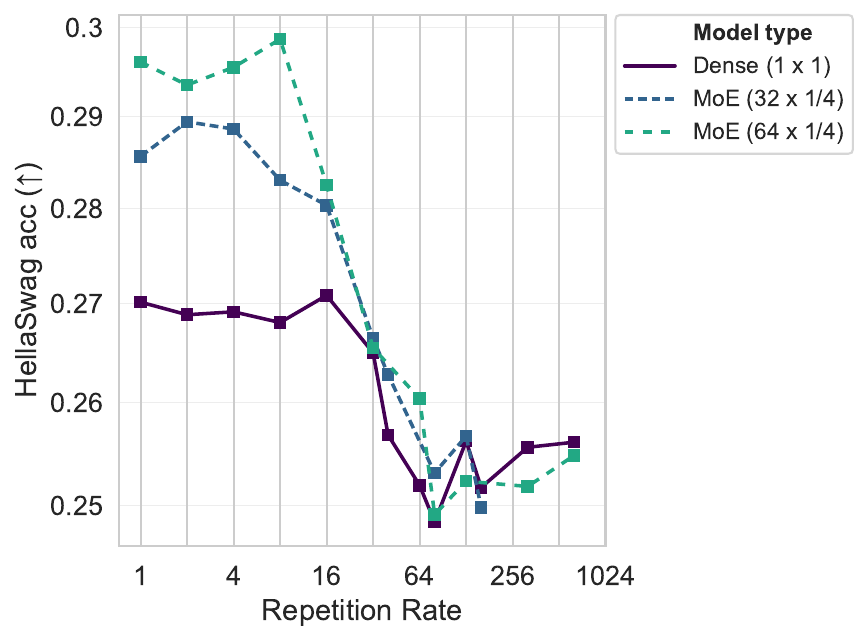}
        \end{subfigure}
        \begin{subfigure}[t]{0.33\textwidth}
            \centering
            \includegraphics[width=\linewidth]{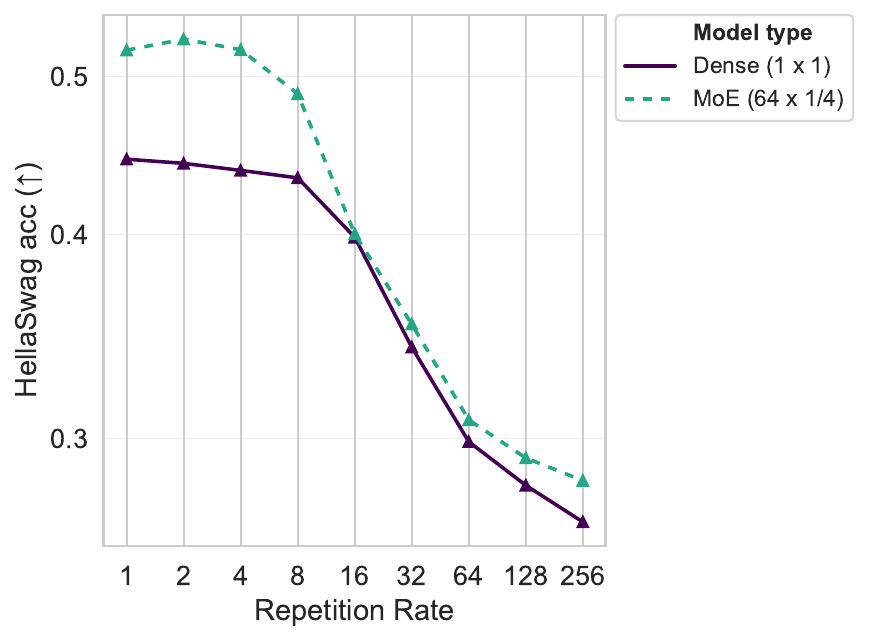}
        \end{subfigure}
    \subcaption{Hellaswag (Accuracy $\uparrow$)}
    \end{subfigure}
    \par\vspace{1em}
    \caption{\textbf{Downstream task performance closely follows validation loss.} For 80M, 200M, and 1B active parameter models, downstream task loss is subject to random noise, but loosely follows trends of validation loss.}
    \label{fig:downstream_tasks}
\end{figure*}

\end{document}